\documentclass[10pt,twocolumn,letterpaper]{article}

\usepackage[pagenumbers]{main/cvpr}
\usepackage{algorithmic}
\usepackage{array}
\usepackage{textcomp}
\usepackage{stfloats}
\usepackage{url}
\usepackage{graphicx}
\usepackage{amsmath,amssymb,amsfonts}
\usepackage{bbm}
\usepackage{multirow,multicol}
\usepackage[T1]{fontenc}
\usepackage[accsupp]{axessibility}
\usepackage{microtype}
\usepackage{balance}
\usepackage{wrapfig,booktabs}
\usepackage{overpic}
\usepackage{pifont}
\usepackage{tikz}
\usetikzlibrary{patterns}
\usepackage{pgfplots}
\usepackage{comment}
\usepackage{dsfont}
\pgfplotsset{compat=1.18}
\pgfplotsset{every axis/.append style={
    title={\small \textbf{Overall Preference}},
    title style={at={(0.5,1)}, anchor=south, yshift=0pt},
    ymin=0, ymax=100,
    ybar stacked, bar width=15pt,
    xtick=data,
    ytick={},
    yticklabels={},
    grid=major,
    major grid style={line width=0.3pt, gray!30},
    axis x line=bottom,
    axis y line=left,
    axis line style = {draw=none},
    x tick label style={rotate=0, anchor=north},
    enlarge x limits=0.35,
    nodes near coords={\pgfmathprintnumber{\pgfplotspointmeta}\%},
    every node near coord/.append style={font=\tiny, text=black},
    width=3cm, height=4cm,
    tick label style={font=\small},
}}

\usepackage{xcolor}
\definecolor{lightgraybg}{RGB}{245,245,245}
\definecolor{deepgrayborder}{RGB}{120,120,120}
\usepackage[table]{xcolor}
\definecolor{verylightpink}{HTML}{FFD6D6}
\definecolor{brightgreen}{RGB}{0,176,80}

\usepackage{listings}
\usepackage{tabularx}
\usepackage{todonotes}
\usepackage{fontawesome5}

\usepackage{cuted}
\usepackage{currfile}
\usepackage{caption}

\definecolor{cvprblue}{rgb}{0.21,0.49,0.74}
\usepackage[pagebackref,breaklinks,colorlinks,allcolors=cvprblue]{hyperref}

\author{
\begin{minipage}{0.19\textwidth}
    \vspace*{-7mm}
    \centering
    Jaime Corsetti$^{1,2}$
\end{minipage}
\begin{minipage}{0.19\textwidth}
    \vspace*{-7mm}
    \centering
    Francesco Giuliari$^1$
\end{minipage}
\begin{minipage}{0.19\textwidth}
    \vspace*{-7mm}
    \centering
    Davide Boscaini$^1$
\end{minipage}
\begin{minipage}{0.16\textwidth}
    \vspace*{-7mm}
    \centering
    Fabio Poiesi$^1$
\end{minipage} \\
\hfill
\begin{minipage}[t]{0.45\textwidth}
\centering
     $^1$Fondazione Bruno Kessler
\end{minipage}
\begin{minipage}[t]{0.45\textwidth}
\centering
    $^2$University of Trento
\end{minipage} \\
\hfill
}
\title{
Training-free steering of video diffusion models\\ for functional object interaction
}

\begin{document}
\newcommand{\higherbetter}[0]{{\color{black!50}{$\,\uparrow$}}}
\newcommand{\oracle}[1]{\textcolor{gray}{#1}}

\newcommand{\impp}[1]{{\textcolor{Green}{+#1}}}
\newcommand{\impn}[1]{{\textcolor{BrickRed}{-#1}}}

\newcommand{\tbd}[1]{\textcolor{red}{#1}}

\newcommand{\cmark}{\ding{51}}
\newcommand{\xmark}{\ding{55}}

\newcommand{\predflow}{\hat{\mathbf{P}}\xspace} 
\newcommand{\gtflow}{\mathbf{P}\xspace} 
\newcommand{\image}{$\mathcal{I}$\xspace}
\newcommand{\depth}{$\mathcal{D}$\xspace} 
\newcommand{\prompt}{$\mathcal{P}$\xspace} 
\newcommand{\mask}{$\mathcal{M}$\xspace} 
\newcommand{\gtmask}{$\mathcal{M}_{\mathrm{GT}}$\xspace} 
\newcommand{\evalmask}{$\mathcal{M}_{\mathrm{EVAL}}$\xspace} 
\newcommand{\video}{$\mathcal{V}$\xspace} 
\newcommand{\vdm}{$\Phi$\xspace} 
\newcommand{\lockmask}{$\mathcal{F}$\xspace} 
\newcommand{\calibset}{$\mathcal{S}$\xspace} 
\newcommand{\motionheadset}{$\mathcal{H}$\xspace}
\newcommand{\objectv}{$\mathbf{v}^{*}$\xspace}

\newcommand{\predvalmask}{\hat{\mathbf{V}}}
\newcommand{\initmask}{\mathbf{M}}
\newcommand{\shapethree}[3]{\in \mathbb{R}^{#1 \times #2 \times #3}}

\definecolor{customgreen}{RGB}{113,170,96}
\definecolor{customgray}{RGB}{180,180,180}
\definecolor{customred}{RGB}{220,90,90}
\definecolor{forestgreen}{RGB}{34,139,34}
\definecolor{myazure}{rgb}{0.8509,0.8980,0.9412}
\definecolor{mygreen}{rgb}{0.85,0.94,0.85}
\definecolor{myred}{rgb}{0.94,0.85,0.85}

\newcommand{\ourmethod}[0]{Steer2Flow\xspace}
\newcommand{\ourdataset}[0]{FunFlow\xspace}
\newcommand{\suppmat}[0]{Supp.~Mat.\xspace}
\captionsetup[subfigure]{font=scriptsize, labelformat=empty, skip=1pt}

\twocolumn[{
    \renewcommand\twocolumn[1][]{#1}%
    \vspace{-1cm}
    \maketitle
    \thispagestyle{empty}
    \vspace{-15mm}
\begin{center}
    \centering
    \begin{overpic}[width=\linewidth,clip,trim={0px 0px 0px 0px}]{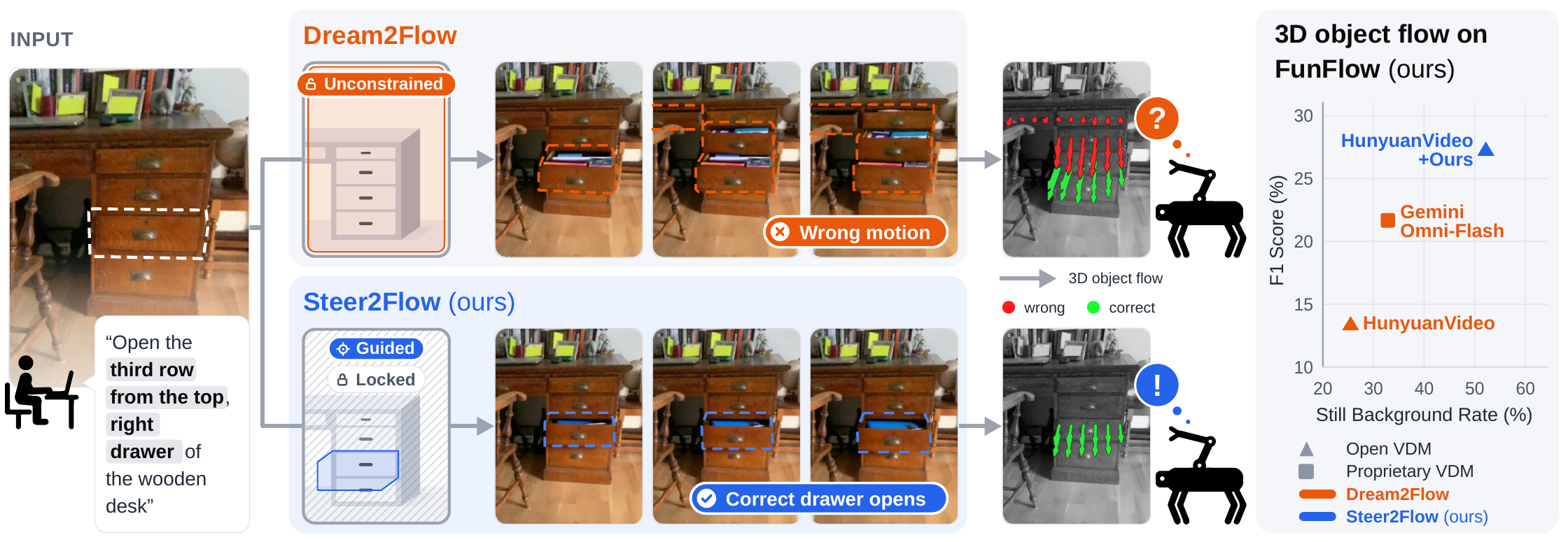}
    \end{overpic}

    \vspace{-3mm}
    \captionof{figure}{
    When a spatial reference identifies the target object among similar distractors (left), a pre-trained video diffusion model (VDM) often moves a different object (\eg a drawer), so the 3D object flow extracted by Dream2Flow \cite{dharmarajan2025dream2flow} is wrong (top).
    \ourmethod steers the same VDM without training by locking the scene to the starting image and guiding the motion within a region that expands from the target object, so only the correct drawer opens and its flow can be converted into robot actions (bottom).
    On \ourdataset (right), \ourmethod improves a VDM in object flow accuracy (F1) and in the share of background correctly left still (Still Background Rate).
    Code and dataset will be released.
    }
    \label{fig:teaser}
\end{center}

}]

\begin{abstract}
Video diffusion models (VDMs) can turn an image and a task description into a video whose 3D object flow can be converted into robot actions for functional interaction.
Yet when a spatial reference identifies the target among distractors, \eg \texttt{open the middle drawer of the cabinet}, they often move the wrong object or several at once.
We observe that a small set of attention heads, which we call motion heads, reveals early in the denoising process whether and in which direction the target will move.
We build on this in \ourmethod, a training-free method that steers a VDM by first segmenting candidate instances and selecting the target with Set-of-Mark prompting.
It then probes the motion heads after 15\% of denoising and restarts from new noise if the target shows no coherent motion.
Lastly, a latent lock freezes the scene to the starting image and lets only the target move along the probed direction.
On \ourdataset, a benchmark we built with 178 spatial interactions in 63 indoor SceneFun3D scenes and ground-truth 3D object flows, \ourmethod raises the whole-image object flow F1 score of HunyuanVideo over Dream2Flow from 13.4 to 27.3, above the 21.7 of the proprietary Omni-Flash, and comes within 2.6 points of it within the target mask.
\end{abstract}

\section{Introduction}\label{sec:intro}

Functionality understanding in 3D scenes interprets task descriptions to locate functional interactive elements, \eg handles and knobs, and infer motion to operate them~\cite{delitzas2024scenefun3d}.
Prior methods ground them and estimate motion type, axis and origin~\cite{delitzas2024scenefun3d,corsetti2025fun3du,zhang2025openfungraph}, but not how interaction unfolds, which a robot needs to act.
Video diffusion models (VDMs) can fill this gap with no robotic training data, unlike world action models~\cite{zhen20243dvla,wu2023unleashing}.
Dream2Flow~\cite{dharmarajan2025dream2flow} feeds a pre-trained image-and-text-to-video (IT2V) model an image and task description, extracts from its video the 3D object flow, \ie the object's 3D point trajectories, and maps it to robot actions.

Flow reliability depends on the generated video, and the task description does not guarantee the intended motion.
In indoor scenarios, several functional elements share an affordance category and only a spatial reference identifies the target, \eg \texttt{open the top left drawer of the dresser}.
State-of-the-art VDMs, open~\cite{kong2024hunyuanvideo} and proprietary~\cite{deepmind2026geminiomniflash}, often fail here (\cref{fig:teaser}).
They lack \emph{spatial awareness} and may move a distractor or several instances, and the pipeline cannot recover as it localises the moving object by a name all drawers share.
They may hallucinate new elements.
They may leave the target still, discovered only after decoding, tracking and lifting the video to 3D.

We observe that a VDM reveals its intended motion before the video is generated.
After 15\% of denoising, a small set of attention heads establishing temporal correspondences~\cite{nam2026emergent}, which we name \emph{motion heads}, show whether and in which direction the target will move.
On \ourdataset, they anticipate a moving target with 61\% precision against a base rate of 52\%, and 79\% recall, and its direction with a median error of 9.6$^\circ$.
While DiffTrack~\cite{nam2026emergent} selects whole layers with pseudo ground-truth tracks to guide sampling, we select single heads to read the motion of a specific region.
\ourmethod steers a VDM without training by reading its attention and overwriting its latents during denoising.
To resolve the spatial reference, it segments all target-category instances and selects the \emph{target object}, \ie the part actuated by the functional element (\eg the drawer, not its handle), with Set-of-Mark prompting~\cite{yang2023setofmark}.
\emph{Motion probing} reads the target motion from the motion heads, identified once per model by supervised calibration on 75 annotated scenes, the only step needing annotations, as no weight is updated.
Motion-aware seed selection (MASS) restarts denoising from new noise if the target shows no coherent motion.
Each probe costs 15\% of a generation and at most five are run, for an average cost of 1.46 generations.
At each step, the \emph{latent lock} overwrites the latents outside a free region around the target with those of the starting image.
Blended latent diffusion~\cite{avrahami2023blended-inpainting} assumes a given region, whereas ours grows along the probed direction.
The lock discourages hallucinations, confines the motion to the target object, and is released for the last 20\% of the denoising steps to remove boundary artefacts.

We evaluate on \ourdataset, a benchmark of 178 samples from 63 scenes that we built from SceneFun3D~\cite{delitzas2024scenefun3d} by selecting task descriptions with spatial references among same-affordance distractors and converting the motion parameters to ground-truth 3D object flows.
We consider two VDMs, the open HunyuanVideo~\cite{kong2024hunyuanvideo}, which can be steered, and the proprietary Omni-Flash~\cite{deepmind2026geminiomniflash}, a stronger model that can only be prompted.
Over the whole image, \ourmethod raises the F1 of Dream2Flow~\cite{dharmarajan2025dream2flow} with HunyuanVideo from 13.4 to 27.3 and the background left still from 25.4\% to 52.2\%, above the 21.7 and 32.9\% of Omni-Flash.
Selecting between two Dream2Flow generations with our gates gives 15.2.
With the same target mask, F1 grows from 34.2 to 42.3, within 2.6 points from Omni-Flash.
In summary, our contributions are:
\begin{itemize}
\item We experimentally show on HunyuanVideo that the motion heads of a VDM reveal early in denoising whether and in which direction the target will move.
\item We introduce \ourmethod, a training-free method for functional object interaction using this motion to select the noise early in generation and confine motion to the target with a latent lock expanding along the probed direction.
\item We introduce \ourdataset, a benchmark of 178 spatially referenced functional interactions in 63 real-world indoor scenes, with ground-truth 3D object flows.
\end{itemize}

\section{Related work}\label{sec:related}

\noindent\textbf{Functionality understanding in 3D scenes.}
SceneFun3D~\cite{delitzas2024scenefun3d} introduces segmentation, grounding and motion estimation of indoor functional elements, later addressed with vision-language models~\cite{corsetti2025fun3du}, 3D scene graphs~\cite{zhang2025openfungraph} and scene-scale articulation annotations~\cite{halacheva2025articulate3d}.
None provides object flow.
We adapt~\cite{corsetti2025fun3du} to segment the target and derive \ourdataset, with ground-truth 3D object flows, from it.

\noindent\textbf{Video generation for acting.}
Generated videos are turned into robot actions by world action models trained on robotic data~\cite{wu2023unleashing}, by learned video-to-action models~\cite{jang2025dreamgen,black2024zero}, or training-free by extracting the 3D object flow \cite{ko2024learning,dharmarajan2025dream2flow}.
Most execute the generated motion as is.
NovaFlow~\cite{li2025novaflow} uses a VLM to reject implausible flows among candidates, each fully generated, tracked and lifted to 3D.
\ourmethod verifies generation from VDM attention after 15\% of denoising and constrains motion to the target, requiring model access.

\noindent\textbf{Motion in the internal representations of VDMs.}
VDMs encode motion in their attention~\cite{jun2026imamap,wen2025analysisattsinks}, where some heads act as sinks~\cite{wu2026attentionsinks}.
DiffTrack~\cite{nam2026emergent} shows that some layers establish temporal correspondences early in denoising, selects them with pseudo ground-truth tracks and uses them to guide sampling away from degenerate motion.
We ask instead whether these correspondences, read inside the mask of a referred object, predict if and in which direction it will move.
We thus select single heads from their correspondence statistics on a small calibration set, discarding heads acting as sink or with a constant positional bias.

\noindent\textbf{Spatially controlled video diffusion.}
Objects can be steered without training along boxes or trajectories given by users~\cite{jain2024peekaboo,ma2024trailblazer,namekata2025sg} or layouts predicted by a language model~\cite{lian2024lvd}, while SpatialAlign~\cite{liu2026spatialalign} fine-tunes the VDM on spatial relationships.
Blended Latent Diffusion~\cite{avrahami2023blended-inpainting} inpaints a region by replacing the latents outside a mask at each denoising step, video inpainting trains dedicated models~\cite{li2025diffueraser,zhang2024avid}, and Time-to-Move~\cite{singer2026timetomove} denoises a user-animated region at a different noise level from the rest.
In all of them the controlled region comes from outside the VDM, so with our mask they would still need the direction of motion, which \ourmethod reads from the VDM.
Our latent lock adopts the replacement of~\cite{avrahami2023blended-inpainting}, but its mask grows from the target along the direction probed from the VDM, so the constraint follows the motion about to be generated.

\noindent\textbf{Noise selection.}
The initial noise affects diffusion outputs~\cite{xu2025good}, motivating noise optimisation~\cite{guo2024initno} and verifier-guided search for images~\cite{ma2025scaling} and videos~\cite{liu2025video}.
To avoid scoring complete generations, LatSearch~\cite{zhao2026latsearch} evaluates intermediate latents with a latent reward model, and ANSE~\cite{kim2026model} selects seeds by attention uncertainty at the first step.
They target overall visual quality and coherence.
Like ANSE, our probe reads VDM attention early in denoising, but it verifies a task-specific outcome, \ie whether the target moves coherently, and returns the direction driving the latent lock.

\begin{figure*}
    \begin{overpic}[width=1.0\linewidth,clip,trim={0px 0px 0px 0px}]{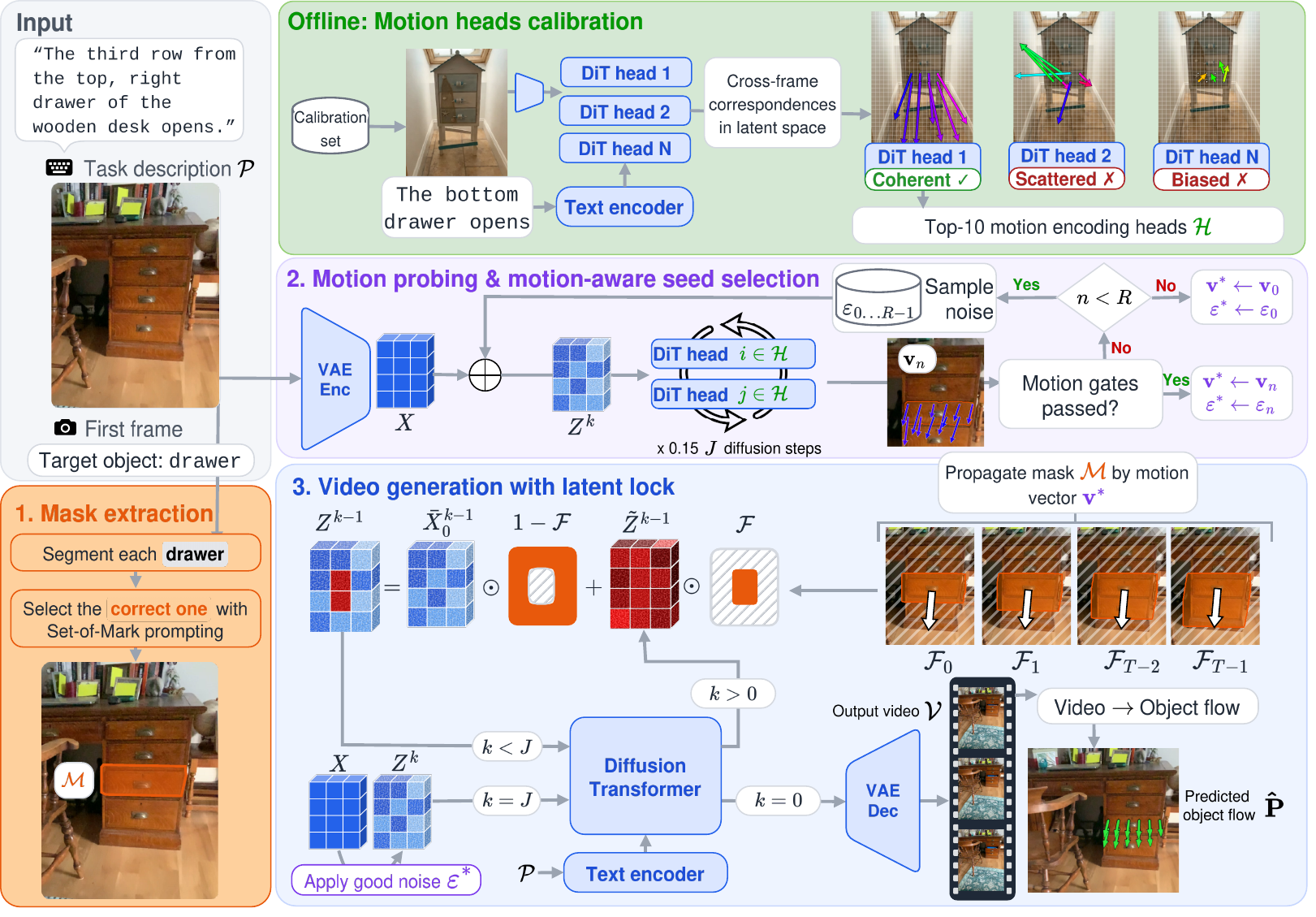}
    \end{overpic}

    \vspace{-2mm}
    \captionof{figure}{
    \ourmethod's framework. It allows a DiT-based video diffusion model to better follow prompts with fine-grained spatial references, without finetuning.
    Its key idea is to read the target motion from a few motion heads after 15\% of denoising, then use it to reject unpromising seeds and to confine generation to a latent lock that grows along the probed direction.
    }
    \label{fig:diagram}
\end{figure*}

\section{Preliminaries: Dream2Flow}\label{sec:preliminary}

Dream2Flow~\cite{dharmarajan2025dream2flow} is a training-free pipeline that uses a VDM as a motion source for robotic manipulation.
Given the starting image $\mathcal{I}$, its depth \depth and a task description $\mathcal{P}$, \eg \texttt{put the bread in the basket}, an IT2V model $\Phi$ generates a video $\mathcal{V} = \Phi(\mathcal{I}, \mathcal{P})$ of a human performing the task.
An open-vocabulary detector~\cite{liu2024groundingdino} and a segmentation model~\cite{ravi2024sam2} localise the object in $\mathcal{I}$, a point tracker~\cite{karaev2025cotracker3} follows the points within its mask, and a depth estimator~\cite{xiao2025spatialtrackerv2} lifts their trajectories to 3D.
The result is the 3D object flow $\predflow \in \mathbb{R}^{N \times T \times 3}$, \ie the trajectories of $N$ object points over $T$ frames, which is converted into robot actions.

Dream2Flow's instructions name the target by category or attribute, never by a spatial reference among same-affordance instances.
The failures described in \cref{sec:intro} corrupt $\predflow$ differently.
If the VDM moves a distractor, the name-based segmentation either selects the moving instance (robot acting on the wrong object), or the referred one (flow close to zero).
If it hallucinates elements, the flow describes content absent from the scene and cannot be executed.
If it leaves the target still, a near-zero flow is returned with the same confidence as a correct one.
Nothing checks $\predflow$ against the prompt, so errors are detected only at execution.

\section{\ourmethod}\label{sec:method}

\ourmethod is a training-free method for functional object interaction with a target among similar objects (\cref{fig:diagram}), and shares some models with Dream2Flow~\cite{dharmarajan2025dream2flow}.
Offline, we identify the \emph{motion heads}, a small subset of attention heads that encode temporal correspondences in their cross-frame attention matrix (\cref{sec:motion_calibration}).
At inference, \ourmethod runs three steps, namely mask extraction, motion probing with seed selection, and the latent lock.
First, we extract a mask of the target object from \image with off-the-shelf vision-language and segmentation models (\cref{sec:mask}).
Second, we probe the motion heads to estimate the motion of the target object during generation, and restart the process from another noise if no motion is detected (\cref{sec:mass}).
Lastly, inspired by inpainting~\cite{avrahami2023blended-inpainting}, we use the estimated motion and the starting mask to expand the area around the target object in which motion is allowed.
Outside this mask, latents are replaced by those of the static starting frame, noised to the current level (\cref{sec:latent_lock}).
The method needs access to the attention and latents, which proprietary models do not expose.

\noindent\textbf{Notation.}
Subscripts index time and superscripts index the diffusion step.
$t \in \left[0, T-1\right]$ denotes output video frames and, at the VAE's coarser temporal resolution, $t \in \left[0, T'-1\right]$ for latent frames.
$k \in \left[0, J\right]$ indexes the noise levels, from full noise ($k=J$) to the clean sample ($k=0$), and denoising step $k \in \left[1, J\right]$ maps level $k$ to level $k-1$.

\subsection{Motion heads calibration}\label{sec:motion_calibration}
Some attention heads of video diffusion transformers establish temporal correspondences~\cite{nam2026emergent}.
Unlike DiffTrack (\cref{sec:related}), which selects layers with an external tracker, we score each head from its match statistics alone, over the first 15\% of denoising steps, as motion emerges early in diffusion~\cite{nam2026emergent,jun2026imamap}.
Calibration is supervised, using ground-truth masks and a manual motion check, but needs no point tracks.
Selection is on a calibration set $\mathcal{C}$ of 75 annotated SceneFun3D scenes disjoint from the test set, on which we also set thresholds and hyperparameters for motion probing gates.

\noindent\textbf{Scoring heads from matches.}
Let $Q_t, K_t \in \mathbb{R}^{P \times D}$ be the queries and keys of the $P$ tokens of latent frame $t \in \left[0, T'-1\right]$ at a given layer and head, with $D$ the head dimension, and let \gtmask be the ground-truth mask of the target object.
The attention of frame $0$ to frame $t$ is
\begin{equation}
    A_t = \operatorname{softmax}\!\Big( \frac{Q_0 K_t^\top}{\sqrt{D}} \Big) \in \mathbb{R}^{P \times P}.
    \label{eq:attention}
\end{equation}
For each patch $p$ of frame $0$, its correspondence $m_t(p)$ in frame $t$ and its displacement $d_t(p)$ are
\begin{equation}\label{eq:matches}
    m_t(p) = \operatorname*{arg\,max}_{x} A_t[p, x],\;
    d_t(p) = m_t(p) - m_0(p),
\end{equation}
where correspondences are 2D coordinates on the latent grid.
We measure displacements from $m_0(p)$ rather than $p$ so that a constant offset of a head when matching frame $0$ to itself is not counted as motion.
We consider only the set $\Omega$ of patches on \gtmask, pooled to the latent grid.

\begin{figure}[t]
    \centering
    \captionsetup[subfigure]{font=scriptsize, labelformat=empty, skip=1pt, justification=centering}
    \definecolor{corrmu}{HTML}{0072B2}
    \definecolor{corrR}{HTML}{E69F00}
    \definecolor{corrnu}{HTML}{CC79A7}
    \newcommand{\corrbars}[3]{%
      \makebox[\linewidth]{\begin{tikzpicture}[x=0.2667\linewidth, y=0.7cm]
        \foreach \v/\c/\l [count=\i from 0] in {#1/corrmu/\mu, #2/corrR/r, #3/corrnu/\nu} {
          \fill[\c] (\i+0.15,0) rectangle (\i+0.85,\v);
          \node[font=\tiny, above, inner sep=1pt] at (\i+0.5,\v) {\v};
          \node[font=\scriptsize, below, inner sep=1pt] at (\i+0.5,0) {$\l$};
        }
        \draw[gray] (0,0) -- (3,0);
      \end{tikzpicture}}}
    \begin{subfigure}[t]{0.245\linewidth}
      \begin{overpic}[width=\linewidth]{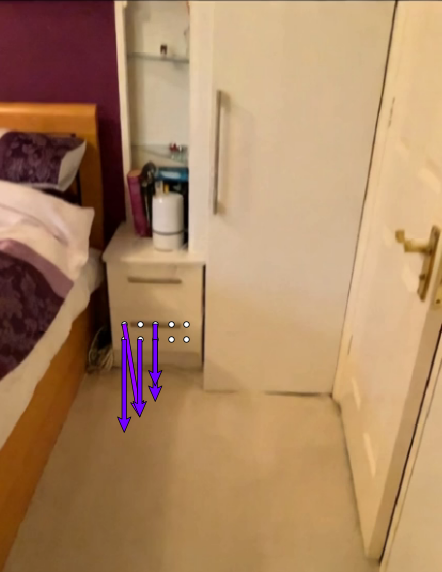}
      \end{overpic}\par\vspace{1pt}
      \corrbars{0.28}{1.00}{0.84}
      \caption[Static]{(a) Static}
      \label{fig:corr_static}
    \end{subfigure}\hfill
    \begin{subfigure}[t]{0.245\linewidth}
      \begin{overpic}[width=\linewidth]{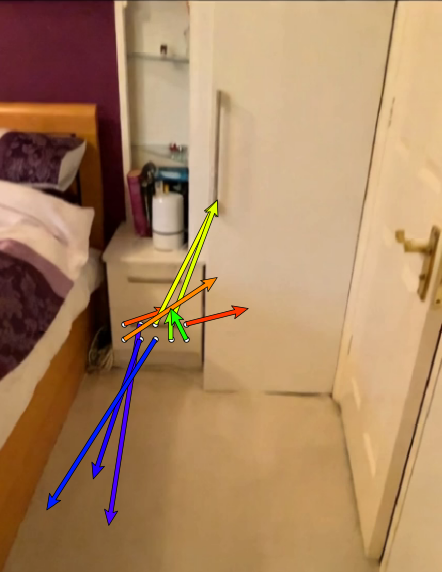}
      \end{overpic}\par\vspace{1pt}
      \corrbars{0.98}{0.10}{0.84}
      \caption[Uniform]{(b) Scattered}
      \label{fig:corr_scattered}
    \end{subfigure}\hfill
    \begin{subfigure}[t]{0.245\linewidth}
      \begin{overpic}[width=\linewidth]{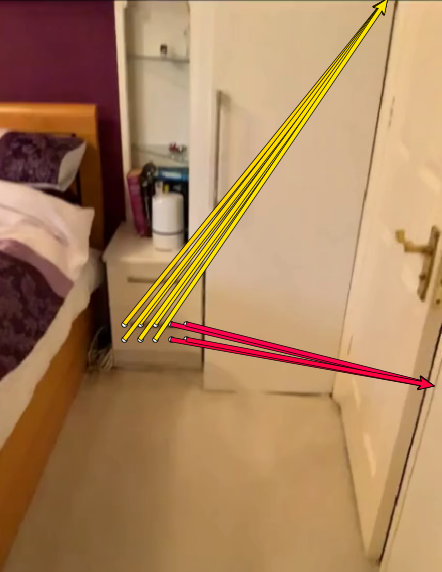}
      \end{overpic}\par\vspace{1pt}
      \corrbars{1.00}{0.80}{0.14}
      \caption[Sink]{(c) Sink}
      \label{fig:corr_sink}
    \end{subfigure}
    \begin{subfigure}[t]{0.245\linewidth}
      \begin{overpic}[width=\linewidth]{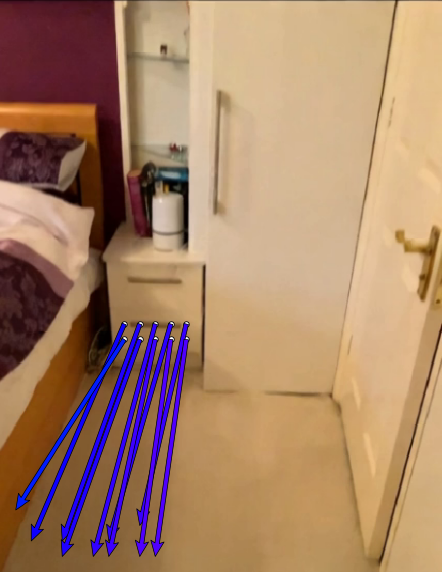}
      \end{overpic}\par\vspace{1pt}
      \corrbars{1.00}{0.99}{0.90}
      \caption[Good]{(d) Good}
      \label{fig:corr_good}
    \end{subfigure}\hfill
  
    \vspace{-2mm}
    \caption{
    Correspondence maps of different attention heads. 
    (a) Static head, low moving fraction $\mu$, 
    (b) Scattered head, low consistency $r$, and 
    (c) Sink head, low unique rate $\nu$. 
    (d) high-quality head, with all scores high. 
    Arrow colour encodes direction.
    }
    \label{fig:correspondence_examples}
  \end{figure}

We score each head on three criteria: the \textit{moving fraction} $\mu$, the fraction of correspondences with significant movement; the \textit{consistency} $r$, their average agreement; and the \textit{unique rate} $\nu$, the fraction of correspondences pointing to a unique cell.
$\mu$ and $r$ capture the presence and agreement of motion, while $\nu$ penalises sink heads, typical of DiTs~\cite{wu2026attentionsinks,wen2025analysisattsinks}.
The head score $S = \mu \, r \, \nu$ is computed for each denoising step $k$, layer $\ell$ and head $h$. 
Examples in \cref{fig:correspondence_examples} show how each component reflects the correspondences.

\noindent\textbf{Selecting the motion heads.}
We keep the scenes of $\mathcal{C}$ whose generated video moves the target correctly, \ie 46 of the 75, and average the score of each head $(\ell, h)$ over them and over the first 15\% of the denoising steps, obtaining $\bar{S}(\ell,h)$.
Some top-scoring heads encode a constant positional bias, pointing the same way in every scene.
For each head, we compute its mean motion direction $e_s(\ell,h)$ in each scene $s$ and their agreement $B(\ell,h)$ across scenes.
We discard heads with $B(\ell,h) \geq 0.9$ and select the 10 heads with highest $\bar{S}(\ell,h)$ as the motion heads \motionheadset, with formulas in the \suppmat

\subsection{Mask extraction}\label{sec:mask}

\ourmethod needs a mask $\mathcal{M} \in \{0,1\}^{H \times W}$ of the target object in \image, with $H \times W$ the resolution of \image.
We obtain it zero-shot with off-the-shelf models, inspired by Fun3DU~\cite{corsetti2025fun3du} (\cref{sec:related}), but instead of its point-then-segment pipeline we segment all candidates and then select one by Set-of-Mark prompting~\cite{yang2023setofmark}.
Claude Haiku 5.5~\cite{anthropic2026haiku55} reads \prompt and returns the category of the moving part (\eg \texttt{drawer}) and a referring phrase for the intended instance (\eg \texttt{top drawer of the cabinet}).
SAM~3~\cite{carion2025sam3segmentconcepts}, prompted with the category as a concept, segments all its instances in \image, and Claude Haiku selects among the masks outlined on \image the one matching the task description \prompt, which we take as \mask.
On \ourdataset, \mask reaches 82.7 mIoU against the ground-truth masks.

\subsection{Motion-aware seed selection}\label{sec:mass}
At inference, \motionheadset estimates the motion that \vdm intends to generate.
We run the first 15\% of the $J$ denoising steps ($k > 0.85\,J$) without intervention and compute the correspondences of \cref{eq:matches} in \motionheadset, with $\Omega$ the patches of the predicted mask \mask pooled to the latent grid.
Each motion head and denoising step votes for the correspondence of each patch $p \in \Omega$ in latent frames $t \geq 1$.
Motion gates on coverage, magnitude and coherence check that the correspondences describe a coherent motion: coverage requires enough moving cells, magnitude a significant displacement, and coherence consistent directions across frames and steps (\suppmat).
If all gates pass, we take the median displacement $\bar{d}_t$ of the moving patches $p \in \Omega$ in each frame $t$, assume a constant velocity $\mathbf{v}$ and fit $\bar{d}_t \approx \tau_t \mathbf{v}$ by least squares,
\begin{equation}
    \mathbf{v} = \frac{\sum_{t=1}^{T'-1} \tau_t \, \bar{d}_t}{\sum_{t=1}^{T'-1} \tau_t^2},
    \label{eq:velocity}
\end{equation}
where $\tau_t = 4t - \tfrac{3}{2}$ is the central video frame of latent frame $t$, given the VAE's temporal compression factor of 4.
Converted to pixels of \image, $\mathbf{v}$ is the target motion per video frame and sets where it may move in the latent lock.
If any gate fails, we restart from new noise, up to $R=5$ attempts $\varepsilon_0,\dots,\varepsilon_{R-1}$.
We set $\varepsilon^{*} \leftarrow \varepsilon_n$ and $\mathbf{v}^{*} \leftarrow \mathbf{v}_n$ for the first passing attempt $n$, or fall back to $\varepsilon_0$ and $\mathbf{v}_0$ if none passes, and run the full generation from $\varepsilon^{*}$ with the latent lock.
Each attempt costs 15\% of a full generation, so that the total cost ranges from 1.15 generations (best case) to $1+0.15R=1.75$ (worst case), with an average of 1.46 on \ourdataset.

\subsection{Latent lock}\label{sec:latent_lock}
Motion probing does not control which object moves, as other instances, \eg other drawers, can also move.
Following image inpainting~\cite{avrahami2023blended-inpainting}, the latent lock confines video changes to a region around the target object: at each denoising step, the latents outside it are replaced by the noised latents of a static video, \ie the replicated first frame \image, while those inside are denoised normally.
The free mask $\mathcal{F} \in \{0,1\}^{T \times H \times W}$ defines this region, with 1 where the video can change and 0 where it is locked, and should cover an area around \mask that expands with time $t$, so we build it from \mask and the probed motion direction \objectv.
As the lock keeps most of the scene static, the prompt should describe the object motion without an agent, \eg \texttt{the drawer opens}, and the baseline receives the same prompt\textcolor{red}.
The prompt is rephrased from the original with Claude Haiku.

Let $\Delta(\mathcal{M}, \mathbf{u})$ translate $\mathcal{M}$ by the 2D vector $\mathbf{u}$, and $\mathrm{Dil}(\mathcal{M}, \delta)$ dilate it with a square kernel of radius $\delta$.
At frame $t$, \lockmask is \mask swept along \objectv up to $t$ and dilated by $\delta_0$, plus a slower isotropic expansion:
\begin{equation}
    \mathcal{F}_t = \mathrm{Dil}\Big( \bigcup_{s=0}^{t} \Delta(\mathcal{M}, s\mathbf{v}^*), \delta_0 \Big) \cup \mathrm{Dil}(\mathcal{M}, \delta_{\text{iso}}\, t),
    \label{eq:directional_free_mask}
\end{equation}
where $\delta_0 = 16$ pixels and $\delta_{\text{iso}} = 4$ pixels per frame.
The first dilation fills sweep gaps, and the isotropic one adds tolerance to diffusion.
Mask translation suits translating parts such as drawers, but only approximates rotating ones such as doors.
Max pooling brings \lockmask to latent time and space.

Let $Z^k$ be the latent video at noise level $\sigma^k$, from $\sigma^{J} = 1$ to $\sigma^{0} = 0$, $X$ the clean latents of the static video and, assuming \vdm to be a rectified linear flow model, $\bar{X}^{k-1} = (1-\sigma^{k-1})\, X + \sigma^{k-1}\, \varepsilon^{*}$ its version noised at level $\sigma^{k-1}$.
Given the output $\tilde{Z}^{k-1}$ of denoising step $k$, we set
\begin{equation}
\small
    Z^{k-1} = (1 - \mathcal{F}) \odot \bar{X}^{k-1} + \mathcal{F} \odot \tilde{Z}^{k-1}.
    \label{eq:latent_lock}
\end{equation}
We apply the lock only in the first 80\% of denoising ($k > 0.2\,J$), which determine motion and structure, and then free every cell. \cref{fig:diagram} shows an example of \lockmask over the frames.

\section{FunFlow benchmark}\label{sec:funflow_benchmark}

Evaluating spatial awareness requires prompts that reference the target through spatial relations, images with distractors of the same kind, and ground-truth object flows, which no existing dataset provides.
We therefore build \ourdataset from SceneFun3D~\cite{delitzas2024scenefun3d}, whose RGB-D videos of indoor scenes annotate the functional elements, \eg handles and knobs of doors, windows and kitchen appliances, with 3D masks, task descriptions and motion parameters.
As 3D masks cover only the functional elements, we annotate the target objects with automatic filtering, SAM~3 and human supervision.

\noindent\textbf{Task description filtering.}
We retain only task descriptions with spatial references, as these imply the presence of distractors, \eg \texttt{open the top left drawer of the dresser} implies several drawers, and discard unambiguous ones such as \texttt{close the kitchen door}.
We filter descriptions with a list of spatial terms, \eg \textit{top}, \textit{bottom} or \textit{to the left of}, and verify each selection manually.
SceneFun3D provides labels which classify functional elements by the movement needed to act on them, and we use them to discard those producing no significant object flow, such as light switches, buttons and rotating knobs.

\noindent\textbf{Visible frame filtering.}
As target visibility varies greatly across frames, we select frames where the target is visible by projecting the functional element's 3D mask onto each frame and ranking frames by Fun3DU's mask score~\cite{corsetti2025fun3du}, which favours large and centred masks.
Frames are extracted from videos in practice, and are often blurry.
Therefore, we combine the score by geometric mean with the Laplacian variance~\cite{pertuz2013laplacian}, which measures sharpness.
The top 10 frames of each sample are shown to the annotator.

\noindent\textbf{Mask annotation.}
The annotator selects the best of these frames and clicks on the target object in a web interface, and SAM~3~\cite{carion2025sam3segmentconcepts} turns clicks into a mask, which can be refined with positive and negative points or rejected.
We discard samples where no frame shows the target clearly.
As we verify every resulting mask by hand, the evaluation mask \gtmask (\cref{sec:motion_calibration}) does not inherit uncorrected SAM~3 errors.

\noindent\textbf{Object flow derivation.}
SceneFun3D annotates each functional element with its motion type, \ie translation or rotation, and the motion direction.
We obtain the object flow by translating the element's 3D mask points along the motion direction or rotating them about the motion axis and pivot.
The target object moves rigidly with the functional element, so we apply the same translation or rotation to the target's 3D points, obtained by lifting \gtmask to 3D.
In \ourdataset, each sample has 5 same-category distractors on average.

\section{Experiments}\label{sec:exps}

\begin{table*}[t]
    \centering
    \caption{Comparison between Steer2Flow and Dream2Flow~\cite{dharmarajan2025dream2flow} on FunFlow.
    We report F1, recall (R), precision (Pr) and Still Background Rate (SBR) of the 3D object flow over the whole image, within our predicted mask, within the ground-truth mask, which also steers Steer2Flow, or within each method's own mask (whole pipeline).
    Row 1 reports Dream2Flow with the proprietary Omni-Flash~\cite{deepmind2026geminiomniflash}, whose API hides the internal features Steer2Flow needs.
    \textbf{Bold} denotes the best result for open VDMs. Our results and gains are highlighted.}
    \label{tab:funflow_main}
    \vspace{-3mm}
    \tabcolsep 5pt
    \resizebox{\textwidth}{!}{%
    \begin{tabular}{rl cccc c ccc c ccc c cccc}
    \toprule
    & & \multicolumn{4}{c}{Whole image} & & \multicolumn{3}{c}{Mask filter (our mask)} & & \multicolumn{3}{c}{Mask filter (GT mask)} & & \multicolumn{4}{c}{Whole pipeline} \\
    \cmidrule{3-6} \cmidrule{8-10} \cmidrule{12-14} \cmidrule{16-19}
    \color{gray} \# & Method & F1 $\uparrow$ & R $\uparrow$ & Pr $\uparrow$ & SBR $\uparrow$ & & F1 $\uparrow$ & R $\uparrow$ & Pr $\uparrow$ & & F1 $\uparrow$ & R $\uparrow$ & Pr $\uparrow$ & & F1 $\uparrow$ & R $\uparrow$ & Pr $\uparrow$ & SBR $\uparrow$ \\
    \midrule

    \multicolumn{19}{l}{\textit{Gemini Omni-Flash~\cite{deepmind2026geminiomniflash} \color{gray}(API)}} \\
    \color{gray} 1 & Dream2Flow~\cite{dharmarajan2025dream2flow} & 21.7 & 40.1 & 18.6 & 32.9 & & 44.9 & 38.0 & 60.3 & & 48.1 & 40.1 & 65.7 & & 19.5 & 21.8 & 23.1 & 32.5 \\
    \specialrule{0.1pt}{\aboverulesep}{\belowrulesep}
    \multicolumn{19}{l}{\textit{HunyuanVideo~\cite{kong2024hunyuanvideo}}} \\
    \color{gray} 2 & Dream2Flow~\cite{dharmarajan2025dream2flow} & 13.4 & 30.6 & 10.5 & 25.4 & & 34.2 & 28.8 & 49.2 & & 37.0 & 30.6 & 54.5 & & 13.2 & 17.7 & 15.8 & 29.5 \\
    \rowcolor{myazure}\color{gray} 3 & Steer2Flow (ours) & \textbf{27.3} & \textbf{35.9} & \textbf{27.0} & \textbf{52.2} & & \textbf{42.3} & \textbf{35.1} & \textbf{60.5} & & \textbf{44.8} & \textbf{36.9} & \textbf{64.6} & & \textbf{42.3} & \textbf{35.1} & \textbf{60.5} & \textbf{88.1} \\
    \rowcolor{myazure}\color{gray} 4 & Improvement & \color{forestgreen}{+13.9} & \color{forestgreen}{+5.3} & \color{forestgreen}{+16.5} & \color{forestgreen}{+26.8} & & \color{forestgreen}{+8.1} & \color{forestgreen}{+6.3} & \color{forestgreen}{+11.3} & & \color{forestgreen}{+7.8} & \color{forestgreen}{+6.3} & \color{forestgreen}{+10.1} & & \color{forestgreen}{+29.1} & \color{forestgreen}{+17.4} & \color{forestgreen}{+44.7} & \color{forestgreen}{+58.6} \\
    \bottomrule
    \end{tabular}
    }
\end{table*}

\subsection{Experimental setup}\label{sec:exp_setup}

We evaluate on \ourdataset, with 178 samples from 63 SceneFun3D scenes, 76\% with translation (\eg drawers) and 24\% with rotation (\eg doors and windows).
We consider one open and one proprietary VDM.
HunyuanVideo1.5~\cite{kong2024hunyuanvideo} (further named HunyuanVideo) runs with its original parameters, except for frame count, reduced to $T{=}21$ for efficiency.
Omni-Flash~\cite{deepmind2026geminiomniflash}, accessed via Gemini API, can only be prompted, so it is evaluated only as a Dream2Flow baseline.
Methods receive the same prompt, and their object flows are compared with ground-truth one of \ourdataset.
Object flow is typically evaluated with end-point 3D error (EPE3D)~\cite{liu2019flownet3d}, and we also evaluate where flow should be zero, \ie on wrong moving objects, by computing flow over whole images.
Each predicted frame is matched to the temporally nearest ground-truth frame, and as in Dream2Flow, predicted depth is scaled to initial depth \depth across frames by least-squares minimisation.
Precision, recall and F1 scores are based on true-positive ratios across EPE3D thresholds of 2, 5, 10, 15 and 20\,cm, with all metrics averaged per sample.
Still Background Rate (SBR) is the share of background that does not move, so a high SBR means little spurious motion.
Gains are tested with 95\% paired bootstrap intervals over samples, and metric details are reported in \suppmat

We evaluate the object flow in four settings.
\emph{Whole image} applies no filtering.
\emph{Our mask} and \emph{GT mask} crop every method's flow with our predicted or ground-truth part mask, and in the latter \ourmethod is also steered with it.
\emph{Whole pipeline} crops each flow with its own method's mask.
SBR is not reported under a shared mask, which removes background motion equally for all methods.
Unless stated otherwise, \ourmethod steers with our predicted mask.

\subsection{Quantitative results}\label{sec:quant}

\cref{tab:funflow_main} compares \ourmethod with Dream2Flow on HunyuanVideo, and Dream2Flow on proprietary Omni-Flash.
Whole image is the main setting, as no mask hides distractor motion.
A static video (with zero flow) reaches 0 F1 and 100\% SBR, so freezing the scene cannot yield high whole-image F1.
\ourmethod raises F1 from 13.4 to 27.3 and SBR from 25.4\% to 52.2\%, exceeding Omni-Flash on both (21.7 and 32.9\%).
Most of this gain comes from the latent lock guided by the probed direction (see \cref{sec:ablation}).
Compared with Omni-Flash, \ourmethod has higher precision (27.0 against 18.6) and SBR (52.2\% agains 32.9\%) but 4.2 points lower recall (35.9 against 40.1), so Omni-Flash moves the target more completely but also moves distractors.
Our main claim is the improvement of the same VDM, which holds in every setting, with Omni-Flash, a stronger model, as reference.

With our mask, \ourmethod gains 8.1 F1 points over Dream2Flow and reaches 42.3 against 44.9 for Omni-Flash.
Ground-truth mask changes these results little, raising F1 of Dream2Flow and \ourmethod by 2.8 and 2.5 points.
In whole pipeline, segmentation explains much of the gap, as our mask reaches 82.7 mIoU against 22.6 for Dream2Flow's, 41\% of which are empty.
In \suppmat, we report results split by probe outcome, motion type and sample difficulty, as well as confidence intervals for all results, showing that all gains over Dream2Flow in \cref{tab:funflow_main} are statistically significant.

\cref{tab:funflow_gate} compares \ourmethod with Dream2Flow given more compute: Dream2Flow generates two videos per sample and keeps the one scoring best on a version of our motion gates in the pixel space, computed on CoTracker, for a total budget of 2.0 generations per sample.
With an average per-sample budget of 1.46, \ourmethod still raises the whole-image F1 by 12.1, precision by 15.2 and SBR by 26.5 points, and with our mask precision gains 8.0 points.
In practice, we found that \ourmethod runs 2.9 attempts on average: the probe passes at the first seed in 40.4\% of samples, at a later one in 26.4\%, and never in the remaining 33.2\%, which fall back to initial noise and its unverified direction.

\begin{figure}[t]
    \centering
    \begin{overpic}[width=0.6\linewidth]{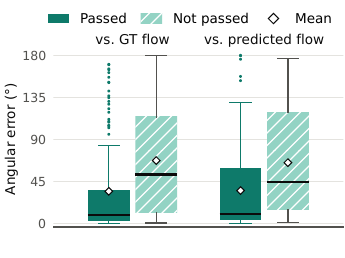}
        \put(19,18.5){\tiny 34.3}  \put(20,12){\tiny 8.8}
        \put(53,27){\tiny 67.3}  \put(53,22.8){\tiny 52.3}
        \put(57,18.5){\tiny 35.1}  \put(58,12){\tiny 9.6}
        \put(91,27){\tiny 65.0}  \put(91,20.8){\tiny 44.7}
    \end{overpic}\hfill
    \includegraphics[width=0.38\linewidth]{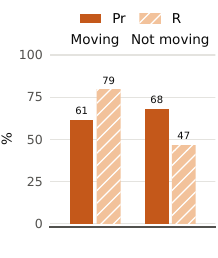}
    \vspace{-4mm}
    \caption{Motion probe accuracy on unconstrained Dream2Flow videos of FunFlow.
            Left, angular error of probed motion $\mathbf{v}^*$ against ground-truth or predicted flow, for samples passing or failing motion gates.
            Diamonds are means, and samples with $\mathbf{v}^*=0$ are excluded.
            Right, precision and recall of passes on moving samples and of failures on non-moving ones.}
    \label{fig:probe_analysis}
    \vspace{-3mm}
\end{figure}

\cref{fig:probe_analysis} evaluates the probe on unconstrained videos of baseline Dream2Flow.
On the left, 2D angular error between $\mathbf{v}^{*}$ and ground-truth or predicted flow has medians of 8.8$^\circ$ and 9.6$^\circ$ when gates pass, and 52.3$^\circ$ and 44.7$^\circ$ when they fail.
Means of 34.3$^\circ$ and 35.1$^\circ$ for passing probes reveal a tail of large errors, but generally low error against predicted flow shows that unconstrained videos follow the probed motion.
On the right, passing gates predict target motion with 61\% precision, against a base rate of 52\%, and recover 79\% of moving samples, but reject only 47\% of still ones, so false passes, rather than missed motion, are the main error source.

\subsection{Qualitative results}\label{sec:qual}

\begin{figure*}[t]
    \centering
    \newcommand{\figrow}[2]{%
        \noindent
        \begin{minipage}[c]{3mm}
            \centering
            \rotatebox{90}{\footnotesize\begin{tabular}{@{}c@{}}#1\end{tabular}}
        \end{minipage}%
        \begin{minipage}[c]{\dimexpr\linewidth-3mm\relax}
            \includegraphics[width=\linewidth]{#2}
        \end{minipage}\par
    }

    \begin{minipage}{\linewidth}
        \footnotesize
        \noindent (a) \texttt{Open the top right drawer of the white closet next to the fireplace}\\[-1pt]
        \figrow{Dream2Flow}{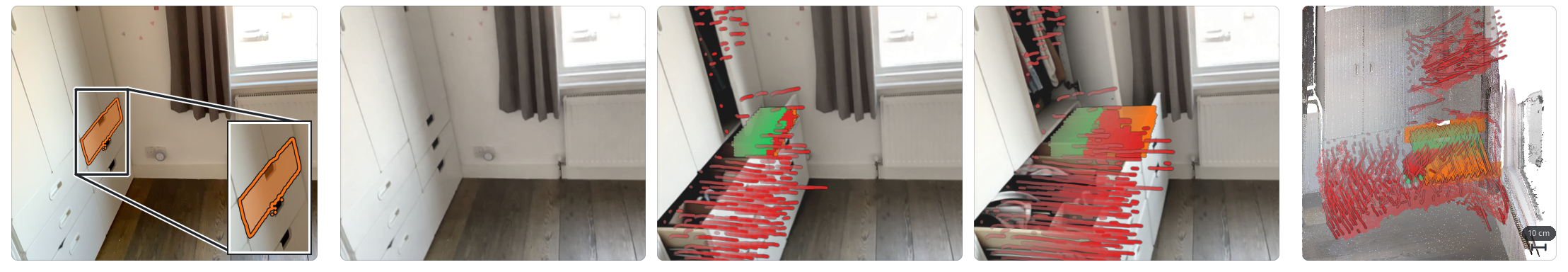}
        \vspace{-4pt}
        \figrow{Steer2Flow (ours)}{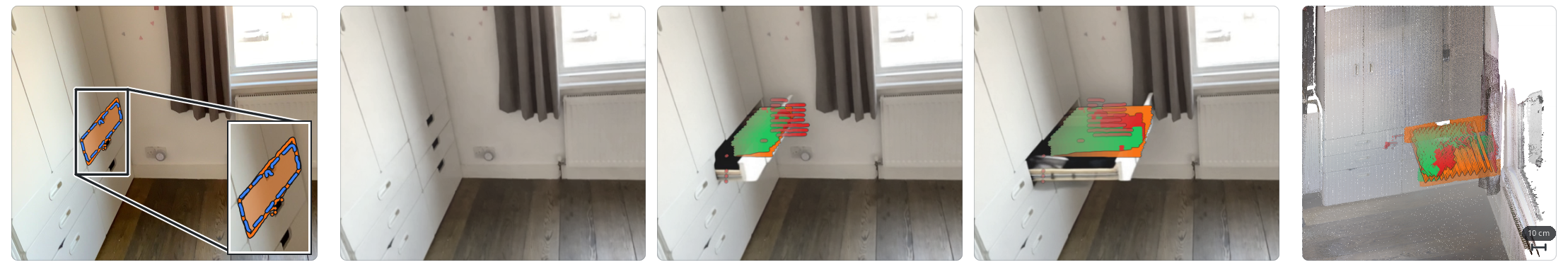}

        \vspace{3pt}
        \noindent (b) \texttt{Open the left cabinet drawer next to the radiator}\\[-1pt]
        \figrow{Dream2Flow}{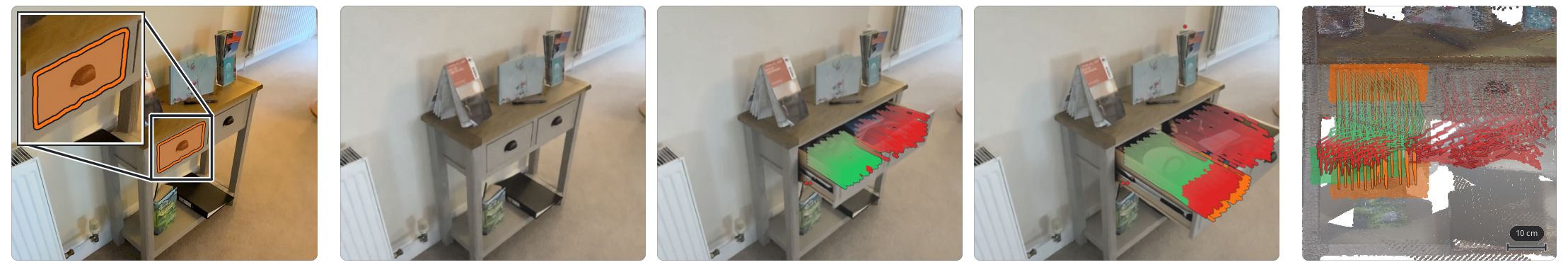}
        \vspace{-4pt}
        \figrow{Steer2Flow (ours)}{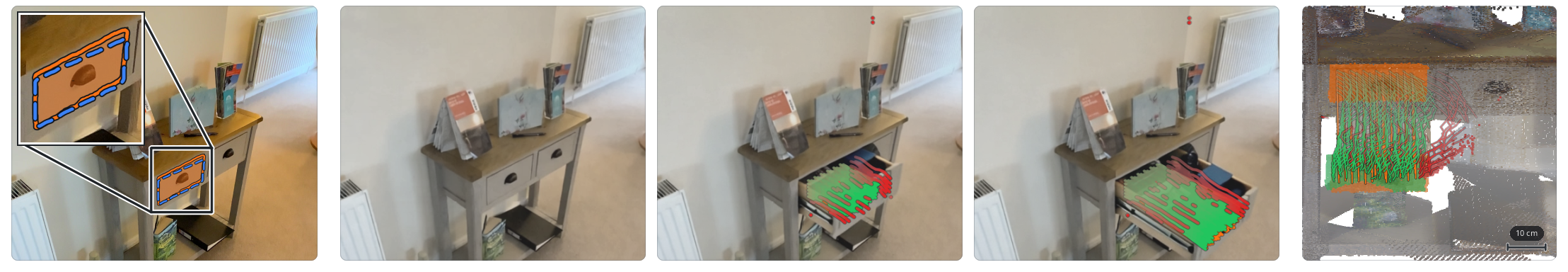}

        \vspace{3pt}
        \noindent (c) \texttt{Open the counter door with the jars on top}\\[-1pt]
        \figrow{Dream2Flow}{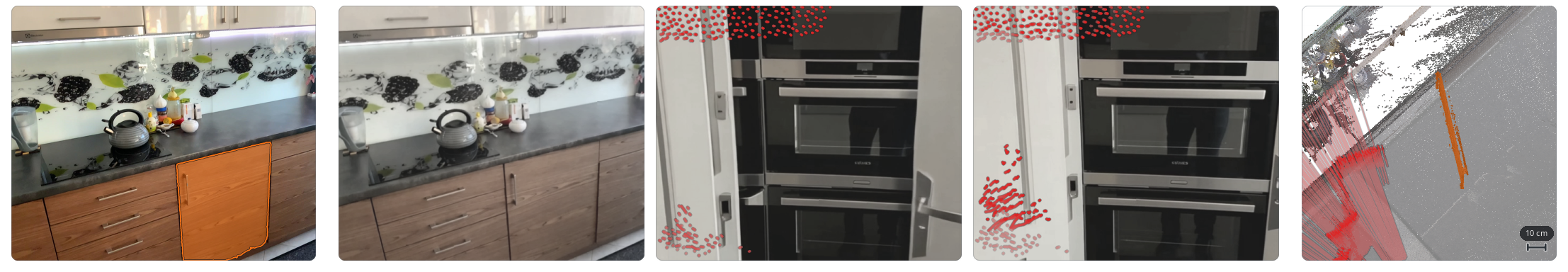}
        \vspace{-4pt}
        \figrow{Steer2Flow (ours)}{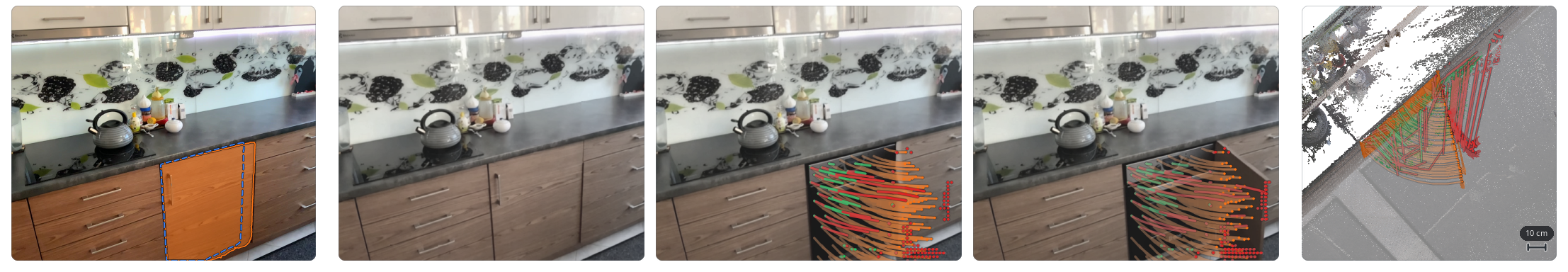}

        \vspace{-1mm}
        \noindent\hspace*{3mm}%
        \begin{minipage}[t]{\dimexpr\linewidth-3mm\relax}
            \setlength{\unitlength}{\linewidth}
            \begin{picture}(1,0.025)
                \put(0.1,0.012){\makebox(0,0){\footnotesize Initial image \image}}
                \put(0.315,0.012){\makebox(0,0){\footnotesize Generated frame $t=0$}}
                \put(0.52,0.012){\makebox(0,0){\footnotesize Generated frame $t=10$}}
                \put(0.72,0.012){\makebox(0,0){\footnotesize Generated frame $t=20$}}
                \put(0.9,0.012){\makebox(0,0){\footnotesize 3D object flow}}
            \end{picture}
        \end{minipage}
    \end{minipage}

    \vspace{-2mm}
    \caption{
    Qualitative results on three scenes of \ourdataset for task descriptions (a), (b), and (c), comparing 3D object flow from Dream2Flow and \ourmethod using HunyuanVideo.
    The leftmost column show the initial frame with our steering mask (blue) and ground-truth mask (orange); central columns show generated video frames with overlaid object flow; the rightmost column shows the 3D object flow.
    }
    \label{fig:qual}
    \vspace{-3mm}
\end{figure*}

\cref{fig:qual} compares \ourmethod with Dream2Flow on HunyuanVideo.
Columns show, from left, the first frame with ground-truth (orange) and our predicted steering (blue) masks, generated frames, and 3D object flow.
Ground-truth flow is orange, and predicted flow is green when its error is below 5\,cm and red otherwise.
In \cref{fig:qual}(a), a closet has several drawers and doors.
Dream2Flow opens all drawers as one, whereas \ourmethod opens only the top right drawer, avoiding a frequent error of VDMs, which tend to act on every instance of the named category object.
\cref{fig:qual}(b) shows a simpler case with two drawers: Dream2Flow opens both, while \ourmethod opens only the left one, although partly morphed on the right side.
We attribute this to the limited resolution of the latent lock, which cannot be pixel-perfect under latent compression.
\cref{fig:qual}(c) shows a common hallucination: as requested, Dream2Flow generates a counter door that opens, but this changes the scene completely, producing wrong object flow.
\ourmethod generates a plausible motion without hallucination, showing the importance of anchoring generation to the first frame through the latent lock.

\begin{table}[t]
    \centering
    \caption{
    Comparison between Steer2Flow and Dream2Flow~\cite{dharmarajan2025dream2flow} on FunFlow.
    Dream2Flow$^\dagger$ uses two seeds per sample and keeps the run whose CoTracker object tracks score higher on our motion probing gates, without using the ground truth.
    We report F1, recall (R), precision (Pr) and Still Background Rate (SBR) of the 3D object flow over the whole image or within our predicted mask.
    \textbf{Bold} denotes the best result. Our results and gains are highlighted.
    }
    \label{tab:funflow_gate}

    \vspace{-3mm}
    \tabcolsep 3pt
    \resizebox{\linewidth}{!}{%
    \begin{tabular}{rl c c cccc c ccc}
    \toprule
    & & Diffusion & & \multicolumn{4}{c}{Whole image} & & \multicolumn{3}{c}{Our mask} \\
    \cmidrule{5-8} \cmidrule{10-12}
    \color{gray} \# & Method & budget $\downarrow$ & & F1 $\uparrow$ & R $\uparrow$ & Pr $\uparrow$ & SBR $\uparrow$ & & F1 $\uparrow$ & R $\uparrow$ & Pr $\uparrow$ \\
    \midrule

    \color{gray} 1 & Dream2Flow$^{\dagger}$~\cite{dharmarajan2025dream2flow} & 2.0 & & 15.2 & 35.6 & 11.8 & 25.7 & & 39.9 & 34.2 & 52.5 \\
    \rowcolor{myazure}\color{gray} 2 & Steer2Flow (ours) & \textbf{1.46} & & \textbf{27.3} & \textbf{35.9} & \textbf{27.0} & \textbf{52.2} & & \textbf{42.3} & \textbf{35.1} & \textbf{60.5} \\
    \rowcolor{myazure}\color{gray} 3 & Improvement & \color{forestgreen}{-0.54} & & \color{forestgreen}{+12.1} & \color{gray}{+0.3} & \color{forestgreen}{+15.2} & \color{forestgreen}{+26.5} & & \color{gray}{+2.4} & \color{gray}{+0.9} & \color{forestgreen}{+8.0} \\
    \bottomrule
    \end{tabular}
    }
    \vspace{-5mm}
\end{table}

\subsection{Ablation studies}\label{sec:ablation}

\newcommand{\no}{\ding{55}}

\begin{table}[t]
\caption{Steer2Flow ablation on HunyuanVideo~\cite{kong2024hunyuanvideo} with our predicted mask and the whole image.
\emph{L.~lock} is the latent lock, expanded along the motion $\mathbf{v}^*$ probed from the heads (probe), or isotropically.
$\mathcal{H}$ is the motion head set used for motion probing: our top-10 heads, their whole layers, or all heads.
Seeds is the number of seeds available to MASS, which is disabled with one.
A dash marks a disabled module. 
Our standard setting is highlighted.}
\label{tab:ablation}
\vspace{-3mm}
\tabcolsep 5pt
\resizebox{\linewidth}{!}{%
\begin{tabular}{r c c c cc ccc}
\toprule
 & & & & \multicolumn{2}{c}{Whole image} & \multicolumn{3}{c}{Our mask} \\
\cmidrule(lr){5-6} \cmidrule(lr){7-9}
\color{gray} \# & L. lock & $\mathcal{H}$ & Seeds & F1 $\uparrow$ & SBR $\uparrow$ & F1 $\uparrow$ & R $\uparrow$ & Pr $\uparrow$\\
\midrule
\color{gray} 1 &  -- & --    & 1 & 13.4 & 25.4 & 34.2 & 28.8 & 49.2 \\
\color{gray} 2 & probe & heads & 1 & 25.4 & \textbf{55.5} & 38.1 & 31.4 & 55.4 \\
\color{gray} 3 & --  & heads & 5 & 14.5 & 24.4 & 38.7 & 32.9 & 52.0 \\
\rowcolor{myazure}
\color{gray} 4 & probe & heads & 5 & \textbf{27.3} & 52.2 & \textbf{42.3} & \textbf{35.1} & \textbf{60.5} \\
\midrule
\multicolumn{9}{l}{\textit{Variants of row 2}} \\
\color{gray} 5 & isotropic & --     & 1 & 21.0 & 51.7 & 32.4 & 26.0 & 51.9 \\
\color{gray} 6 & probe     & layers & 1 & 21.0 & 52.9 & 31.8 & 25.5 & 49.1 \\
\color{gray} 7 & probe     & all    & 1 & 21.4 & 52.4 & 32.7 & 26.3 & 49.7 \\
\bottomrule
\end{tabular}
}
\vspace{-5mm}
\end{table}

\cref{tab:ablation} isolates each module on HunyuanVideo, with rows 1 and 4 as standard Dream2Flow and \ourmethod.
The probed latent lock on a single seed (row 2) contributes most, raising whole-image F1 by 12.0 and SBR by 30.1 over row 1, and F1 by 3.9 with our mask.
MASS alone (row 3) has no significant effect on the whole image but raises F1 by 4.5 with our mask.
Adding the lock to MASS (row 4 against 3) gives the largest gain in precision (8.5) and raises whole-image SBR by 27.8, so anchoring generation to the first frame is key against distractor motion.
Adding MASS to the lock (row 4 against 2) raises F1 by 4.2 with our mask but lowers SBR by 3.3, likely because MASS favours moving samples, which can cause spurious motion when the mask is inaccurate.

Row 5 replaces probed expansion with isotropic, \ie $\mathbf{v}^{*}{=}~\mathbf{0}$ in \cref{eq:directional_free_mask}, lowering F1 by 4.4 and 5.7 points on the whole image and with our mask relative to row 2.
The isotropic lock causes ``clipping'' artefacts, where the object vanishes at the free-mask edge.
Varying \motionheadset, probing all layers containing a top-10 head (row 6) lowers F1 by 4.4 and 6.3 points relative to row 2, like probing every head of every layer (row 7).
Bias or sink heads thus give a wrong direction, along which the lock cuts the target.
We report the confidence intervals for all metrics in the \suppmat

\section{Conclusions}\label{sec:conclusions}

We presented \ourmethod, a training-free method that makes a VDM follow spatially referenced prompts for functional interaction.
Motion heads, identified once per model, reveal after 15\% of denoising whether and in which direction the target object will move, and \ourmethod uses this signal to discard unpromising seeds and grow a latent lock along the probed direction of motion, anchoring the video to the starting image.
On \ourdataset, it doubles HunyuanVideo's whole-image object flow F1 and exceeds proprietary Omni-Flash, which cannot be steered, while remaining 2.6 points below it within the target mask.
\ourmethod requires access to VDM attention and latents, and is validated on one open model.
Its free region translates the target mask, only approximating rotating parts such as doors.
It also fails more often on small or distant targets (see \suppmat), and depth estimates can make flow imprecise even when generated motion is plausible.
Future work will address these limitations.

{
    \small
    \bibliographystyle{main/ieeenat_fullname}
    \bibliography{main}

\begin{thebibliography}{41}
\providecommand{\natexlab}[1]{#1}
\providecommand{\url}[1]{\texttt{#1}}
\expandafter\ifx\csname urlstyle\endcsname\relax
  \providecommand{\doi}[1]{doi: #1}\else
  \providecommand{\doi}{doi: \begingroup \urlstyle{rm}\Url}\fi

\bibitem[{Anthropic}(2026)]{anthropic2026haiku55}
{Anthropic}.
\newblock {Claude Haiku 5.5}.
\newblock Anthropic Model Card, 2026.

\bibitem[Avrahami et~al.(2023)Avrahami, Fried, and Lischinski]{avrahami2023blended-inpainting}
Omri Avrahami, Ohad Fried, and Dani Lischinski.
\newblock Blended latent diffusion.
\newblock \emph{ACM TOG}, 2023.

\bibitem[Black et~al.(2024)Black, Nakamoto, Atreya, Walke, Finn, Kumar, and Levine]{black2024zero}
Kevin Black, Mitsuhiko Nakamoto, Pranav Atreya, Homer Walke, Chelsea Finn, Aviral Kumar, and Sergey Levine.
\newblock Zero-shot robotic manipulation with pre-trained image-editing diffusion models.
\newblock In \emph{ICLR}, 2024.

\bibitem[Carion et~al.(2026)Carion, Gustafson, Hu, Debnath, Hu, Suris Coll-Vinent, Ryali, Alwala, Khedr, Huang, et~al.]{carion2025sam3segmentconcepts}
Nicolas Carion, Laura Gustafson, Yuan-Ting Hu, Shoubhik Debnath, Ronghang Hu, Didac Suris Coll-Vinent, Chaitanya Ryali, Kalyan~Vasudev Alwala, Haitham Khedr, Andrew Huang, et~al.
\newblock {SAM 3}: Segment anything with concepts.
\newblock In \emph{ICLR}, 2026.

\bibitem[Corsetti et~al.(2025)Corsetti, Giuliari, Fasoli, Boscaini, and Poiesi]{corsetti2025fun3du}
Jaime Corsetti, Francesco Giuliari, Alice Fasoli, Davide Boscaini, and Fabio Poiesi.
\newblock Functionality understanding and segmentation in {3D} scenes.
\newblock In \emph{CVPR}, 2025.

\bibitem[Delitzas et~al.(2024)Delitzas, Takmaz, Tombari, Sumner, Pollefeys, and Engelmann]{delitzas2024scenefun3d}
Alexandros Delitzas, Ayca Takmaz, Federico Tombari, Robert Sumner, Marc Pollefeys, and Francis Engelmann.
\newblock {SceneFun3D}: Fine-grained functionality and affordance understanding in {3D} scenes.
\newblock In \emph{CVPR}, 2024.

\bibitem[Dharmarajan et~al.(2026)Dharmarajan, Huang, Wu, Fei-Fei, and Zhang]{dharmarajan2025dream2flow}
Karthik Dharmarajan, Wenlong Huang, Jiajun Wu, Li Fei-Fei, and Ruohan Zhang.
\newblock {Dream2Flow}: Bridging video generation and open-world manipulation with {3D} object flow.
\newblock In \emph{ICRA}, 2026.

\bibitem[{Google DeepMind}(2026)]{deepmind2026geminiomniflash}
{Google DeepMind}.
\newblock {Gemini Omni Flash: Model Card}.
\newblock Google DeepMind Model Card, 2026.

\bibitem[Guo et~al.(2024)Guo, Liu, Cui, Li, Yang, and Huang]{guo2024initno}
Xiefan Guo, Jinlin Liu, Miaomiao Cui, Jiankai Li, Hongyu Yang, and Di Huang.
\newblock {InitNO}: Boosting text-to-image diffusion models via initial noise optimization.
\newblock In \emph{CVPR}, 2024.

\bibitem[Halacheva et~al.(2025)Halacheva, Miao, Zaech, Wang, {Van Gool}, and Paudel]{halacheva2025articulate3d}
Anna-Maria Halacheva, Yang Miao, Jan-Nico Zaech, Xi Wang, Luc {Van Gool}, and Danda~Pani Paudel.
\newblock {Articulate3D}: Holistic understanding of {3D} scenes as universal scene description.
\newblock In \emph{ICCV}, 2025.

\bibitem[Jain et~al.(2024)Jain, Nasery, Vineet, and Behl]{jain2024peekaboo}
Yash Jain, Anshul Nasery, Vibhav Vineet, and Harkirat Behl.
\newblock Peekaboo: Interactive video generation via masked-diffusion.
\newblock In \emph{CVPR}, 2024.

\bibitem[Jang et~al.(2025)Jang, Ye, Lin, Xiang, Bjorck, Fang, Hu, Huang, Kundalia, Lin, et~al.]{jang2025dreamgen}
Joel Jang, Seonghyeon Ye, Zongyu Lin, Jiannan Xiang, Johan Bjorck, Yu Fang, Fengyuan Hu, Spencer Huang, Kaushil Kundalia, Yen-Chen Lin, et~al.
\newblock Dreamgen: Unlocking generalization in robot learning through video world models.
\newblock \emph{arXiv preprint arXiv:2505.12705}, 2025.

\bibitem[Jun et~al.(2026)Jun, Kang, Han, and Hwang]{jun2026imamap}
Youngjun Jun, Seil Kang, Woojung Han, and Seong~Jae Hwang.
\newblock {I'm a map! Interpretable} motion-attentive maps: Spatio-temporally localizing concepts in video diffusion transformers.
\newblock In \emph{CVPR}, 2026.

\bibitem[Karaev et~al.(2025)Karaev, Makarov, Wang, Neverova, Vedaldi, and Rupprecht]{karaev2025cotracker3}
Nikita Karaev, Yuri Makarov, Jianyuan Wang, Natalia Neverova, Andrea Vedaldi, and Christian Rupprecht.
\newblock {CoTracker3}: Simpler and better point tracking by pseudo-labelling real videos.
\newblock In \emph{ICCV}, 2025.

\bibitem[Kim and Kim(2026)]{kim2026model}
Kwanyoung Kim and Sanghyun Kim.
\newblock Model already knows the best noise: Bayesian active noise selection via attention in video diffusion model.
\newblock In \emph{ICLR}, 2026.

\bibitem[Ko et~al.(2024)Ko, Mao, Du, Sun, and Tenenbaum]{ko2024learning}
Po-Chen Ko, Jiayuan Mao, Yilun Du, Shao-Hua Sun, and Joshua~B. Tenenbaum.
\newblock Learning to act from actionless videos through dense correspondences.
\newblock In \emph{ICLR}, 2024.

\bibitem[Li et~al.(2025{\natexlab{a}})Li, Sun, Hu, Ta, Barry, Konidaris, and Fu]{li2025novaflow}
Hongyu Li, Lingfeng Sun, Yafei Hu, Duy Ta, Jennifer Barry, George Konidaris, and Jiahui Fu.
\newblock {NovaFlow}: Zero-shot manipulation via actionable flow from generated videos.
\newblock \emph{arXiv preprint arXiv:2510.08568}, 2025{\natexlab{a}}.

\bibitem[Li et~al.(2025{\natexlab{b}})Li, Xue, Ren, and Bo]{li2025diffueraser}
Xiaowen Li, Haolan Xue, Peiran Ren, and Liefeng Bo.
\newblock {DiffuEraser}: A diffusion model for video inpainting.
\newblock \emph{arXiv preprint arXiv:2501.10018}, 2025{\natexlab{b}}.

\bibitem[Lian et~al.(2024)Lian, Shi, Yala, Darrell, and Li]{lian2024lvd}
Long Lian, Baifeng Shi, Adam Yala, Trevor Darrell, and Boyi Li.
\newblock {LLM}-grounded video diffusion models.
\newblock In \emph{ICLR}, 2024.

\bibitem[Liu et~al.(2025)Liu, Wang, Cai, Zhang, Zhan, and Duan]{liu2025video}
Fangfu Liu, Hanyang Wang, Yimo Cai, Kaiyan Zhang, Xiaohang Zhan, and Yueqi Duan.
\newblock {Video-T1}: Test-time scaling for video generation.
\newblock In \emph{ICCV}, 2025.

\bibitem[Liu et~al.(2026)Liu, Cham, and Zheng]{liu2026spatialalign}
Fengming Liu, Tat-Jen Cham, and Chuanxia Zheng.
\newblock {SpatialAlign}: Aligning dynamic spatial relationships in video generation.
\newblock \emph{arXiv preprint arXiv:2602.22745}, 2026.

\bibitem[Liu et~al.(2024)Liu, Zeng, Ren, Li, Zhang, Yang, Jiang, Li, Yang, Su, et~al.]{liu2024groundingdino}
Shilong Liu, Zhaoyang Zeng, Tianhe Ren, Feng Li, Hao Zhang, Jie Yang, Qing Jiang, Chunyuan Li, Jianwei Yang, Hang Su, et~al.
\newblock {Grounding DINO}: Marrying {DINO} with grounded pre-training for open-set object detection.
\newblock In \emph{ECCV}, 2024.

\bibitem[Liu et~al.(2019)Liu, Qi, and Guibas]{liu2019flownet3d}
Xingyu Liu, Charles~R. Qi, and Leonidas~J. Guibas.
\newblock {FlowNet3D}: Learning scene flow in {3D} point clouds.
\newblock In \emph{CVPR}, 2019.

\bibitem[Ma et~al.(2025)Ma, Tong, Jia, Hu, Su, Zhang, Yang, Li, Jaakkola, Jia, et~al.]{ma2025scaling}
Nanye Ma, Shangyuan Tong, Haolin Jia, Hexiang Hu, Yu-Chuan Su, Mingda Zhang, Xuan Yang, Yandong Li, Tommi Jaakkola, Xuhui Jia, et~al.
\newblock Scaling inference time compute for diffusion models.
\newblock In \emph{CVPR}, 2025.

\bibitem[Ma et~al.(2024)Ma, Lewis, and Kleijn]{ma2024trailblazer}
Wan-Duo~Kurt Ma, John~P. Lewis, and W.~Bastiaan Kleijn.
\newblock Trailblazer: Trajectory control for diffusion-based video generation.
\newblock In \emph{SIGGRAPH Asia}, 2024.

\bibitem[Nam et~al.(2025)Nam, Son, Chung, Kim, Jin, Hur, and Kim]{nam2026emergent}
Jisu Nam, Soowon Son, Dahyun Chung, Jiyoung Kim, Siyoon Jin, Junhwa Hur, and Seungryong Kim.
\newblock Emergent temporal correspondences from video diffusion transformers.
\newblock In \emph{NeurIPS}, 2025.

\bibitem[Namekata et~al.(2025)Namekata, Bahmani, Wu, Kant, Gilitschenski, and Lindell]{namekata2025sg}
Koichi Namekata, Sherwin Bahmani, Ziyi Wu, Yash Kant, Igor Gilitschenski, and David Lindell.
\newblock {SG-I2V}: Self-guided trajectory control in image-to-video generation.
\newblock In \emph{ICLR}, 2025.

\bibitem[Pertuz et~al.(2013)Pertuz, Puig, and Garcia]{pertuz2013laplacian}
Said Pertuz, Domenec Puig, and Miguel~Angel Garcia.
\newblock Analysis of focus measure operators for shape-from-focus.
\newblock \emph{Pattern Recognition}, 2013.

\bibitem[Ravi et~al.(2025)Ravi, Gabeur, Hu, Hu, Ryali, Ma, Khedr, R{\"a}dle, Rolland, Gustafson, et~al.]{ravi2024sam2}
Nikhila Ravi, Valentin Gabeur, Yuan-Ting Hu, Ronghang Hu, Chaitanya Ryali, Tengyu Ma, Haitham Khedr, Roman R{\"a}dle, Chloe Rolland, Laura Gustafson, et~al.
\newblock {SAM 2}: Segment anything in images and videos.
\newblock In \emph{ICLR}, 2025.

\bibitem[Singer et~al.(2026)Singer, Rotstein, Mann, Kimmel, and Litany]{singer2026timetomove}
Assaf Singer, Noam Rotstein, Amir Mann, Ron Kimmel, and Or Litany.
\newblock {Time-to-Move}: Training-free motion controlled video generation via dual-clock denoising.
\newblock In \emph{ICLR}, 2026.

\bibitem[Wen et~al.(2025)Wen, Wu, Jain, Goldstein, and Panda]{wen2025analysisattsinks}
Yuxin Wen, Jim Wu, Ajay Jain, Tom Goldstein, and Ashwinee Panda.
\newblock Analysis of attention in video diffusion transformers.
\newblock \emph{arXiv preprint arXiv:2504.10317}, 2025.

\bibitem[Wu et~al.(2025)Wu, Zou, Li, Huang, Yang, Tan, Peng, Wu, Xiong, Jiang, et~al.]{kong2024hunyuanvideo}
Bing Wu, Chang Zou, Changlin Li, Duojun Huang, Fang Yang, Hao Tan, Jack Peng, Jianbing Wu, Jiangfeng Xiong, Jie Jiang, et~al.
\newblock {HunyuanVideo} 1.5 technical report.
\newblock \emph{arXiv preprint arXiv:2511.18870}, 2025.

\bibitem[Wu and Summa(2026)]{wu2026attentionsinks}
Fangzheng Wu and Brian Summa.
\newblock Attention sinks in diffusion transformers: A causal analysis.
\newblock In \emph{ICML}, 2026.

\bibitem[Wu et~al.(2024)Wu, Jing, Cheang, Chen, Xu, Li, Liu, Li, and Kong]{wu2023unleashing}
Hongtao Wu, Ya Jing, Chilam Cheang, Guangzeng Chen, Jiafeng Xu, Xinghang Li, Minghuan Liu, Hang Li, and Tao Kong.
\newblock Unleashing large-scale video generative pre-training for visual robot manipulation.
\newblock In \emph{ICLR}, 2024.

\bibitem[Xiao et~al.(2025)Xiao, Wang, Xue, Karaev, Makarov, Kang, Zhu, Bao, Shen, and Zhou]{xiao2025spatialtrackerv2}
Yuxi Xiao, Jianyuan Wang, Nan Xue, Nikita Karaev, Yuri Makarov, Bingyi Kang, Xing Zhu, Hujun Bao, Yujun Shen, and Xiaowei Zhou.
\newblock {SpatialTrackerV2}: Advancing {3D} point tracking with explicit camera motion.
\newblock In \emph{ICCV}, 2025.

\bibitem[Xu et~al.(2025)Xu, Zhang, and Shi]{xu2025good}
Katherine Xu, Lingzhi Zhang, and Jianbo Shi.
\newblock Good seed makes a good crop: Discovering secret seeds in text-to-image diffusion models.
\newblock In \emph{WACV}, 2025.

\bibitem[Yang et~al.(2023)Yang, Zhang, Li, Zou, Li, and Gao]{yang2023setofmark}
Jianwei Yang, Hao Zhang, Feng Li, Xueyan Zou, Chunyuan Li, and Jianfeng Gao.
\newblock {Set-of-Mark} prompting unleashes extraordinary visual grounding in {GPT-4V}.
\newblock \emph{arXiv preprint arXiv:2310.11441}, 2023.

\bibitem[Zhang et~al.(2025)Zhang, Delitzas, Wang, Zhang, Ji, Pollefeys, and Engelmann]{zhang2025openfungraph}
Chenyangguang Zhang, Alexandros Delitzas, Fangjinhua Wang, Ruida Zhang, Xiangyang Ji, Marc Pollefeys, and Francis Engelmann.
\newblock Open-vocabulary functional {3D} scene graphs for real-world indoor spaces.
\newblock In \emph{CVPR}, 2025.

\bibitem[Zhang et~al.(2024)Zhang, Wu, Wang, Luo, Zhang, Zhao, Vajda, Metaxas, and Yu]{zhang2024avid}
Zhixing Zhang, Bichen Wu, Xiaoyan Wang, Yaqiao Luo, Luxin Zhang, Yinan Zhao, Peter Vajda, Dimitris Metaxas, and Licheng Yu.
\newblock {AVID}: Any-length video inpainting with diffusion model.
\newblock In \emph{CVPR}, 2024.

\bibitem[Zhao et~al.(2026)Zhao, Liu, Cao, Gong, Zhang, Song, Deng, and Patras]{zhao2026latsearch}
Zengqun Zhao, Ziquan Liu, Yu Cao, Shaogang Gong, Zhensong Zhang, Jifei Song, Jiankang Deng, and Ioannis Patras.
\newblock {LatSearch}: Latent reward-guided search for faster inference-time scaling in video diffusion.
\newblock In \emph{ECCV}, 2026.

\bibitem[Zhen et~al.(2024)Zhen, Qiu, Chen, Yang, Yan, Du, Hong, and Gan]{zhen20243dvla}
Haoyu Zhen, Xiaowen Qiu, Peihao Chen, Jincheng Yang, Xin Yan, Yilun Du, Yining Hong, and Chuang Gan.
\newblock {3D-VLA}: {3D} vision-language-action generative world model.
\newblock In \emph{ICML}, 2024.

\end{thebibliography}


\begin{thebibliography}{14}
\providecommand{\natexlab}[1]{#1}
\providecommand{\url}[1]{\texttt{#1}}
\expandafter\ifx\csname urlstyle\endcsname\relax
  \providecommand{\doi}[1]{doi: #1}\else
  \providecommand{\doi}{doi: \begingroup \urlstyle{rm}\Url}\fi

\bibitem[Agarwal et~al.(2026)Agarwal, Ali, Allen, Antolini, Aubame, Azzolini, Bai, Bala, Balaji, Bapst, et~al.]{agarwal2026cosmos}
Niket Agarwal, Arslan Ali, Jon Allen, Martin Antolini, Adeline Aubame, Alisson Azzolini, Junjie Bai, Maciej Bala, Yogesh Balaji, Josh Bapst, et~al.
\newblock Cosmos 3: Omnimodal world models for physical ai.
\newblock \emph{arXiv preprint arXiv:2606.02800}, 2026.

\bibitem[{Anthropic}(2026)]{anthropic2026haiku55}
{Anthropic}.
\newblock {Claude Haiku 5.5}.
\newblock Anthropic Model Card, 2026.

\bibitem[Carion et~al.(2026)Carion, Gustafson, Hu, Debnath, Hu, Suris Coll-Vinent, Ryali, Alwala, Khedr, Huang, et~al.]{carion2025sam3segmentconcepts}
Nicolas Carion, Laura Gustafson, Yuan-Ting Hu, Shoubhik Debnath, Ronghang Hu, Didac Suris Coll-Vinent, Chaitanya Ryali, Kalyan~Vasudev Alwala, Haitham Khedr, Andrew Huang, et~al.
\newblock {SAM 3}: Segment anything with concepts.
\newblock In \emph{ICLR}, 2026.

\bibitem[Corsetti et~al.(2025)Corsetti, Giuliari, Fasoli, Boscaini, and Poiesi]{corsetti2025fun3du}
Jaime Corsetti, Francesco Giuliari, Alice Fasoli, Davide Boscaini, and Fabio Poiesi.
\newblock Functionality understanding and segmentation in {3D} scenes.
\newblock In \emph{CVPR}, 2025.

\bibitem[Deitke et~al.(2025)Deitke, Clark, Lee, Tripathi, Yang, Park, Salehi, Muennighoff, Lo, Soldaini, et~al.]{deitke2025molmo}
Matt Deitke, Christopher Clark, Sangho Lee, Rohun Tripathi, Yue Yang, Jae~Sung Park, Mohammadreza Salehi, Niklas Muennighoff, Kyle Lo, Luca Soldaini, et~al.
\newblock Molmo and pixmo: Open weights and open data for state-of-the-art vision-language models.
\newblock In \emph{CVPR}, 2025.

\bibitem[Delitzas et~al.(2024)Delitzas, Takmaz, Tombari, Sumner, Pollefeys, and Engelmann]{delitzas2024scenefun3d}
Alexandros Delitzas, Ayca Takmaz, Federico Tombari, Robert Sumner, Marc Pollefeys, and Francis Engelmann.
\newblock {SceneFun3D}: Fine-grained functionality and affordance understanding in {3D} scenes.
\newblock In \emph{CVPR}, 2024.

\bibitem[Dharmarajan et~al.(2026)Dharmarajan, Huang, Wu, Fei-Fei, and Zhang]{dharmarajan2025dream2flow}
Karthik Dharmarajan, Wenlong Huang, Jiajun Wu, Li Fei-Fei, and Ruohan Zhang.
\newblock {Dream2Flow}: Bridging video generation and open-world manipulation with {3D} object flow.
\newblock In \emph{ICRA}, 2026.

\bibitem[Efron(1979)]{efron1979bootstrap}
Bradley Efron.
\newblock Bootstrap methods: Another look at the jackknife.
\newblock \emph{The Annals of Statistics}, 1979.

\bibitem[{Google DeepMind}(2026)]{deepmind2026geminiomniflash}
{Google DeepMind}.
\newblock {Gemini Omni Flash: Model Card}.
\newblock Google DeepMind Model Card, 2026.

\bibitem[Kirillov et~al.(2023)Kirillov, Mintun, Ravi, Mao, Rolland, Gustafson, Xiao, Whitehead, Berg, Lo, et~al.]{kirillov2023segmentany}
Alexander Kirillov, Eric Mintun, Nikhila Ravi, Hanzi Mao, Chloe Rolland, Laura Gustafson, Tete Xiao, Spencer Whitehead, Alexander~C Berg, Wan-Yen Lo, et~al.
\newblock {Segment Anything}.
\newblock In \emph{ICCV}, 2023.

\bibitem[Liu et~al.(2024)Liu, Zeng, Ren, Li, Zhang, Yang, Jiang, Li, Yang, Su, et~al.]{liu2024groundingdino}
Shilong Liu, Zhaoyang Zeng, Tianhe Ren, Feng Li, Hao Zhang, Jie Yang, Qing Jiang, Chunyuan Li, Jianwei Yang, Hang Su, et~al.
\newblock {Grounding DINO}: Marrying {DINO} with grounded pre-training for open-set object detection.
\newblock In \emph{ECCV}, 2024.

\bibitem[Ravi et~al.(2025)Ravi, Gabeur, Hu, Hu, Ryali, Ma, Khedr, R{\"a}dle, Rolland, Gustafson, et~al.]{ravi2024sam2}
Nikhila Ravi, Valentin Gabeur, Yuan-Ting Hu, Ronghang Hu, Chaitanya Ryali, Tengyu Ma, Haitham Khedr, Roman R{\"a}dle, Chloe Rolland, Laura Gustafson, et~al.
\newblock {SAM 2}: Segment anything in images and videos.
\newblock In \emph{ICLR}, 2025.

\bibitem[Wu et~al.(2025)Wu, Zou, Li, Huang, Yang, Tan, Peng, Wu, Xiong, Jiang, et~al.]{kong2024hunyuanvideo}
Bing Wu, Chang Zou, Changlin Li, Duojun Huang, Fang Yang, Hao Tan, Jack Peng, Jianbing Wu, Jiangfeng Xiong, Jie Jiang, et~al.
\newblock {HunyuanVideo} 1.5 technical report.
\newblock \emph{arXiv preprint arXiv:2511.18870}, 2025.

\bibitem[Yang et~al.(2023)Yang, Zhang, Li, Zou, Li, and Gao]{yang2023som}
Jianwei Yang, Hao Zhang, Feng Li, Xueyan Zou, Chunyuan Li, and Jianfeng Gao.
\newblock {Set-of-Mark} prompting unleashes extraordinary visual grounding in {GPT-4V}.
\newblock \emph{arXiv preprint arXiv:2310.11441}, 2023.

\end{thebibliography}
}
\end{document}


\newcommand{\fabio}[1]{\todo[color=green!20, inline, author=Fabio]{#1}}
\newcommand{\davide}[1]{\todo[color=yellow!20, inline, author=Davide]{#1}}
\newcommand{\jaime}[1]{\todo[color=red!20, inline, author=Jaime]{#1}}
\newcommand{\francesco}[1]{\todo[color=orange!20, inline, author=Francesco]{#1}}

\newcommand{\warning}[1]{\textbf{\color{red!90}{#1}}}
\newcommand{\higherbetter}[0]{{\color{black!50}{$\,\uparrow$}}}
\newcommand{\oracle}[1]{\textcolor{gray}{#1}}

\newcommand{\impp}[1]{{\textcolor{Green}{+#1}}}
\newcommand{\impn}[1]{{\textcolor{BrickRed}{-#1}}}

\newcommand{\tbd}[1]{\textcolor{red}{#1}}

\newcommand{\cmark}{\ding{51}}%
\newcommand{\xmark}{\ding{55}}%

\newcommand{\predflow}{\hat{\mathbf{P}}\xspace}
\newcommand{\gtflow}{\mathbf{P}\xspace}
\newcommand{\image}{$\mathcal{I}$\xspace}
\newcommand{\depth}{$\mathcal{D}$\xspace}
\newcommand{\prompt}{$\mathcal{P}$\xspace}
\newcommand{\mask}{$\mathcal{M}$\xspace}
\newcommand{\gtmask}{$\mathcal{M}_{\mathrm{GT}}$\xspace}
\newcommand{\evalmask}{$\mathcal{M}_{\mathrm{EVAL}}$\xspace}
\newcommand{\video}{$\mathcal{V}$\xspace}
\newcommand{\vdm}{$\Phi$\xspace}
\newcommand{\lockmask}{$\mathcal{F}$\xspace}
\newcommand{\calibset}{$\mathcal{S}$\xspace}
\newcommand{\motionheadset}{$\mathcal{H}$\xspace}
\newcommand{\objectv}{$\mathbf{v}^{*}$\xspace}

\newcommand{\predvalmask}{\hat{\mathbf{V}}}
\newcommand{\initmask}{\mathbf{M}}
\newcommand{\shapethree}[3]{\in \mathbb{R}^{#1 \times #2 \times #3}}

\definecolor{customgreen}{RGB}{113,170,96}
\definecolor{customgray}{RGB}{180,180,180}
\definecolor{customred}{RGB}{220,90,90}
\definecolor{forestgreen}{RGB}{34,139,34}
\definecolor{myazure}{rgb}{0.8509,0.8980,0.9412}
\definecolor{mygreen}{rgb}{0.85,0.94,0.85}
\definecolor{myred}{rgb}{0.94,0.85,0.85}

\newcommand{\ourmethod}[0]{Steer2Flow\xspace}
\newcommand{\ourdataset}[0]{FunFlow\xspace}
\newcommand{\suppmat}[0]{Supp.~Mat.\xspace}
\captionsetup[subfigure]{font=scriptsize, labelformat=empty, skip=1pt}

\twocolumn[{%
    \renewcommand\twocolumn[1][]{#1}%
    \vspace{-1cm}
    \maketitle
}]

\noindent This Supplementary Material is structured as follows. 
In \cref{sec:method} we describe in detail the mask extraction process, the motion calibration scores, and the motion probing gates.
We also provide qualitative examples of extracted masks and of head correspondences.
In \cref{sec:eval} we specify the evaluation procedure, in particular how we obtain the ground-truth flow, the predicted flow, and the formulas of the metrics. 
We also outline the process to assess the statistical significance of the results.
We then show that \ourmethod generalizes to another VDM, Cosmos3-Nano, and we report additional ablations related to the motion probing.
We also report quantitative segmentation results, by comparing our mask with Dream2Flow's~\cite{dharmarajan2025dream2flow} and Fun3DU's~\cite{corsetti2025fun3du} masks, the latter of which inspired our approach, and the bootstrap confidence intervals of the results of the main paper.
Finally, in \cref{sec:implementation} we report the details of the VDMs we used, such as the exact checkpoints, the generation settings (denoising steps and number of frames), image resolution and prompts.
\section{Method details}
\label{sec:method}
In this section, we detail the steps and prompts used in the mask extraction step~(\cref{sec:mask_extraction}), the formulation of the scores in the motion calibration phase~(\cref{sec:motion_calib}), and of the gates in motion probing~(\cref{sec:motion_probing}).
Finally, we detail the procedure to set the thresholds on the calibration set~(\cref{sec:thresholds}).

\subsection{Mask extraction}
\label{sec:mask_extraction}
\begin{figure*}[t]
  \centering\small
  \setlength{\tabcolsep}{2pt}
  \newcommand{\query}[1]{\setlength{\fboxsep}{3pt}%
    \colorbox{black!7}{\parbox{\dimexpr\linewidth-2\fboxsep}{\raggedright\itshape #1}}\par}
  \newcommand{\parse}[3]{\footnotesize\query{``#1''}%
    \centerline{\scriptsize$\downarrow$\,\textsc{llm}}%
    \renewcommand{\arraystretch}{1.1}%
    \begin{tabular}{@{}>{$}r<{$}@{\hspace{4pt}}>{\raggedright\arraybackslash\sffamily}p{\dimexpr\linewidth-2.2em}@{}}
      d & #2 \\
      t & #3
    \end{tabular}}
  \newcommand{\select}[3]{\scriptsize\query{``Which label marks the \textbf{#1} that has to be manipulated in
    order to \textbf{#2}?''}%
    \centerline{\scriptsize$\downarrow$\,\textsc{vlm}}%
    \centerline{\footnotesize VLM answer: \textsf{\textbf{#3}}}}
  \newcommand{\note}[1]{\par\vspace{1pt}\centering\scriptsize #1}
  \begin{tabular}{@{}m{0.2\linewidth}m{0.18\linewidth}m{0.18\linewidth}m{0.215\linewidth}m{0.18\linewidth}@{}}
    \multicolumn{1}{c}{\textbf{Parsing}} & \multicolumn{1}{c}{\textbf{SAM~3 candidates}} &
    \multicolumn{1}{c}{\textbf{Set-of-Mark}} & \multicolumn{1}{c}{\textbf{Selection}} &
    \multicolumn{1}{c}{\textbf{Result}} \tabularnewline\midrule
    \parse{Open the second from the left door of the white cabinet with the liquor bottles on top}{door}
          {second from the left door of the white cabinet with the liquor bottles on top}
      & \includegraphics[width=\linewidth]{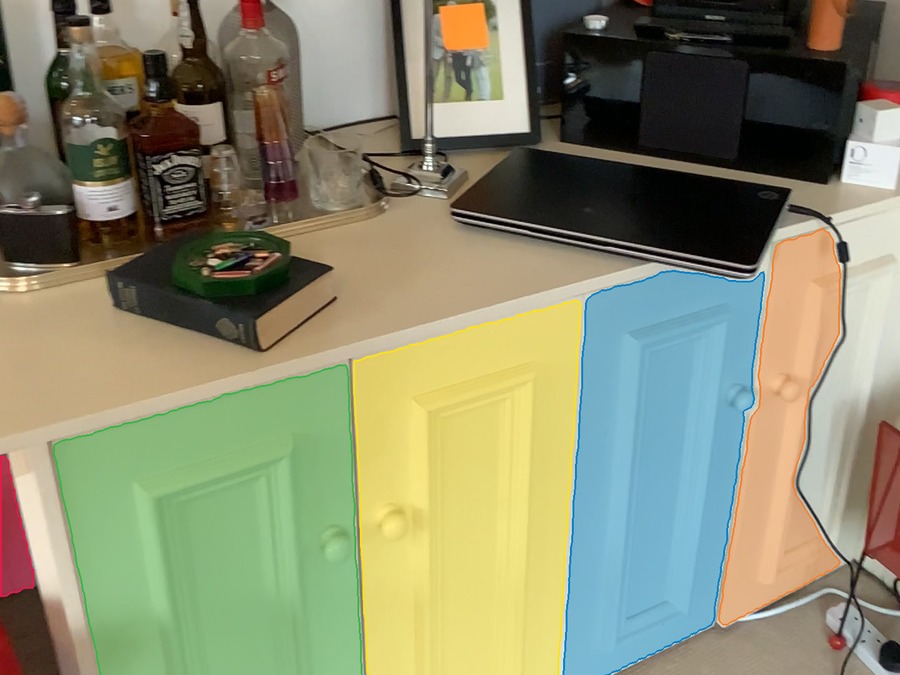}\note{5 candidates} & \includegraphics[width=\linewidth]{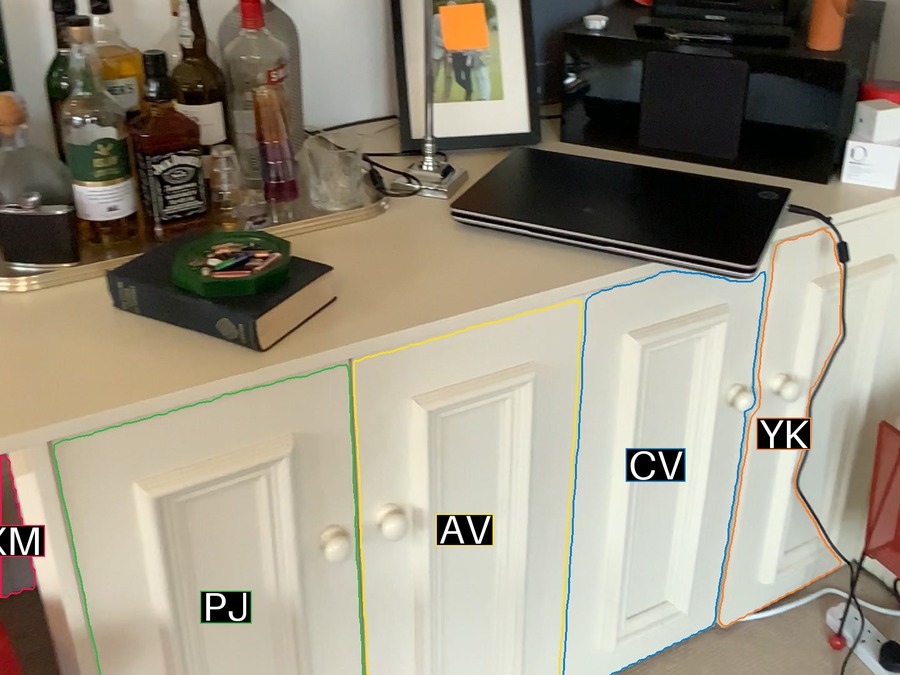}\note{VLM input}
      & \select{second from the left door of the white cabinet with the liquor bottles on top}
              {open the second from the left door of the white cabinet with the liquor bottles on top}{AV}
      & \includegraphics[width=\linewidth]{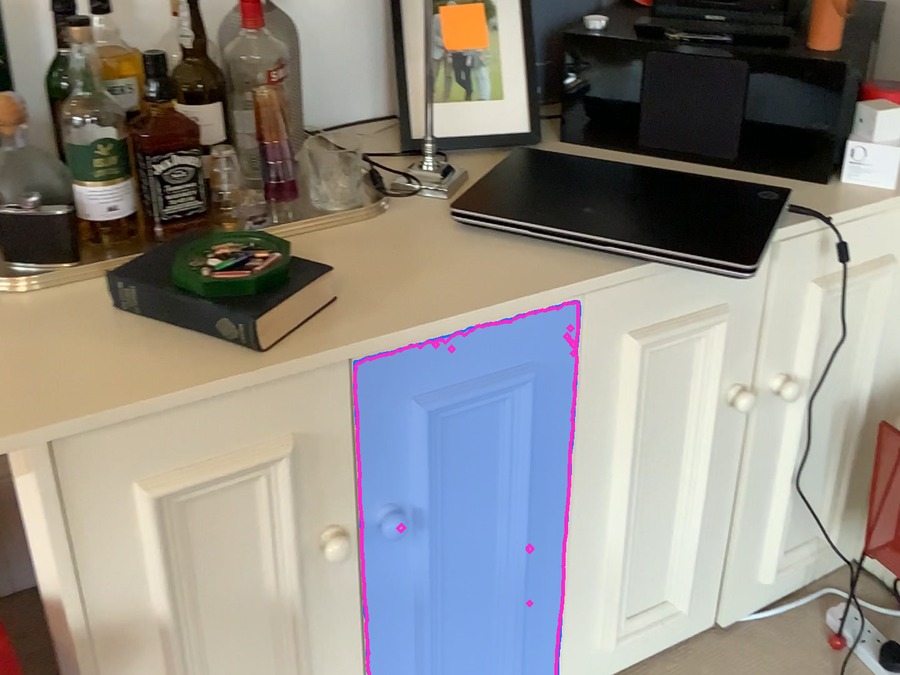}\note{IoU 0.99} \tabularnewline\midrule
    \parse{Open the third from the left cabinet door located directly above the sink}{door}
          {third from the left door of the cabinet}
      & \includegraphics[width=\linewidth]{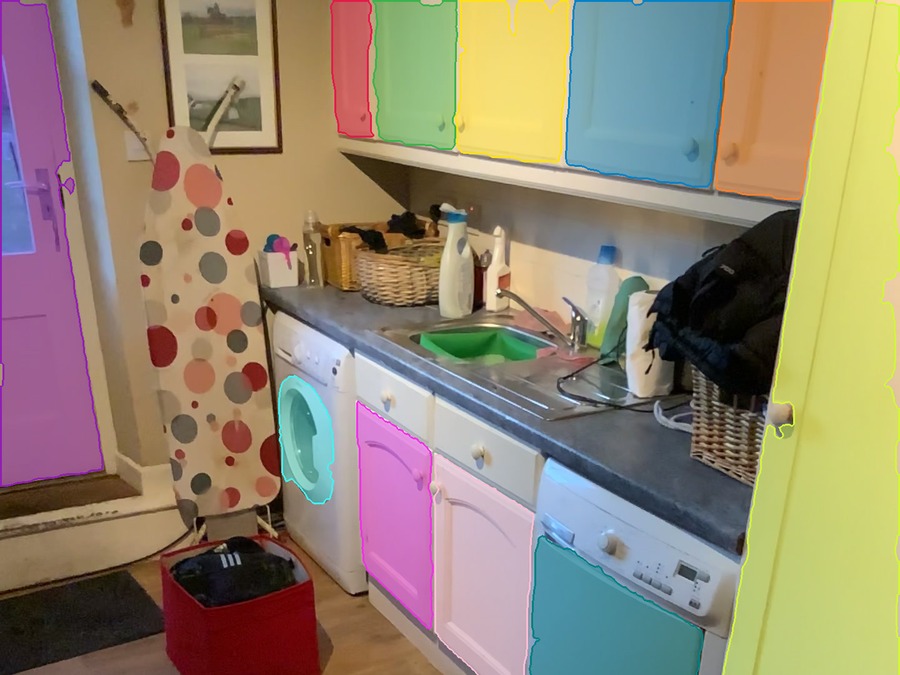}\note{11 candidates} & \includegraphics[width=\linewidth]{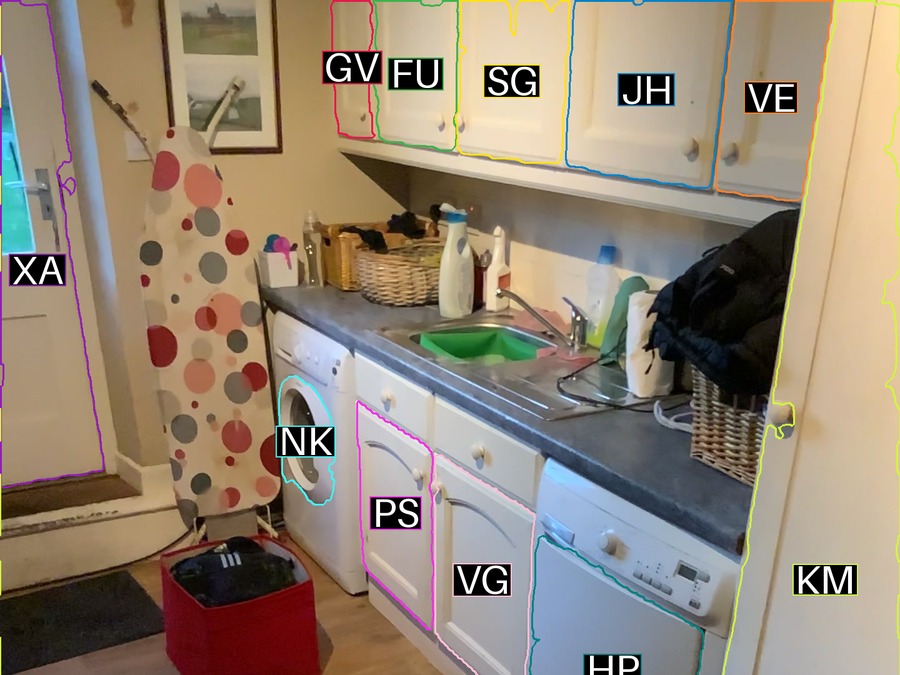}\note{VLM input}
      & \select{third from the left door of the cabinet}
              {open the third from the left cabinet door located directly above the sink}{SG}
      & \includegraphics[width=\linewidth]{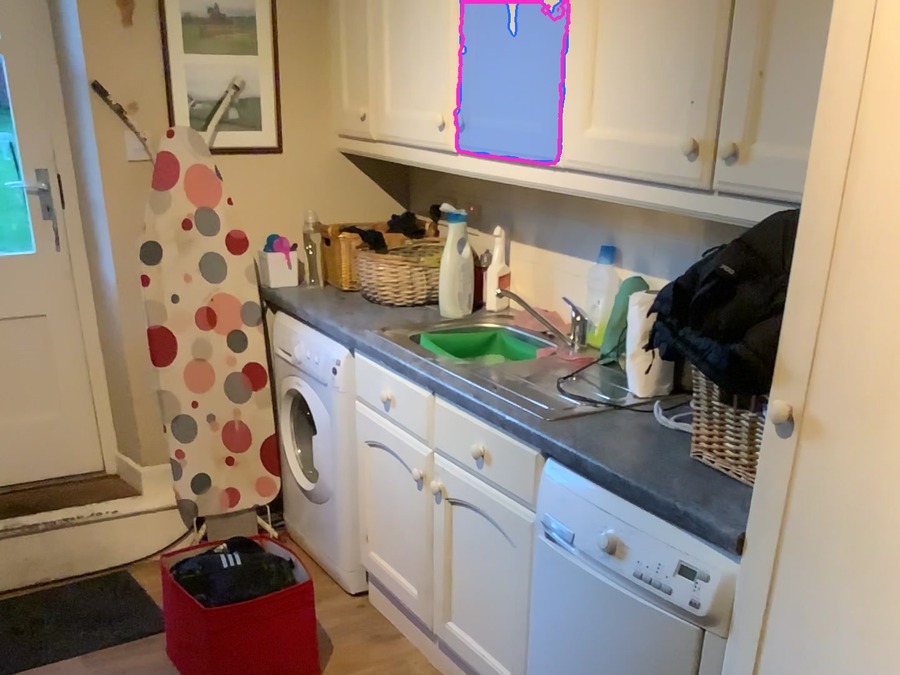}\note{IoU 0.94} \tabularnewline\midrule
    \parse{Open the bottom drawer of the cabinet with the mirror and lamp on top}{drawer}
          {bottom drawer of the cabinet with the mirror and lamp on top}
      & \includegraphics[width=\linewidth]{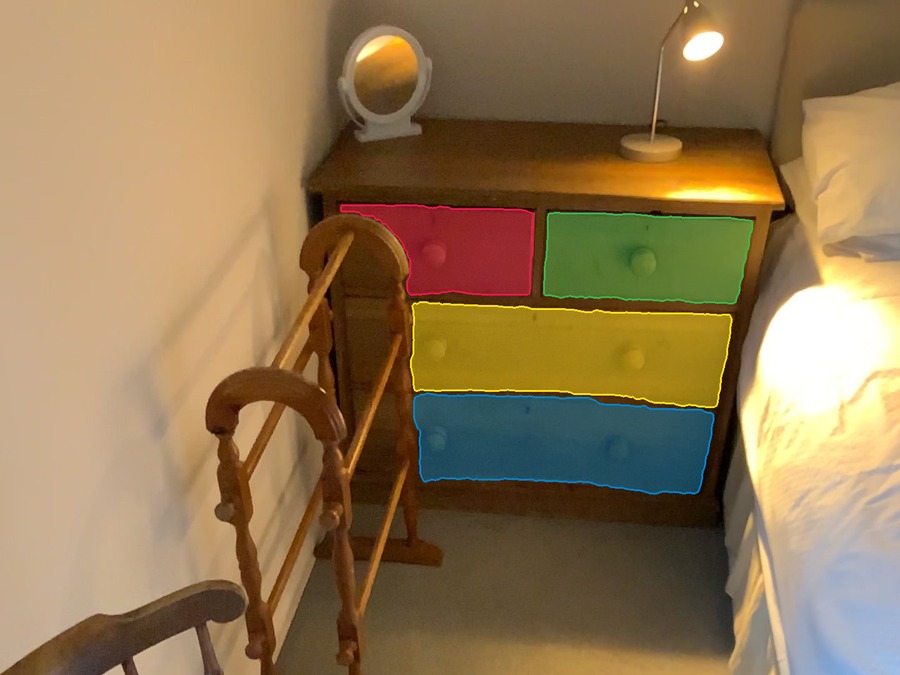}\note{4 candidates} & \includegraphics[width=\linewidth]{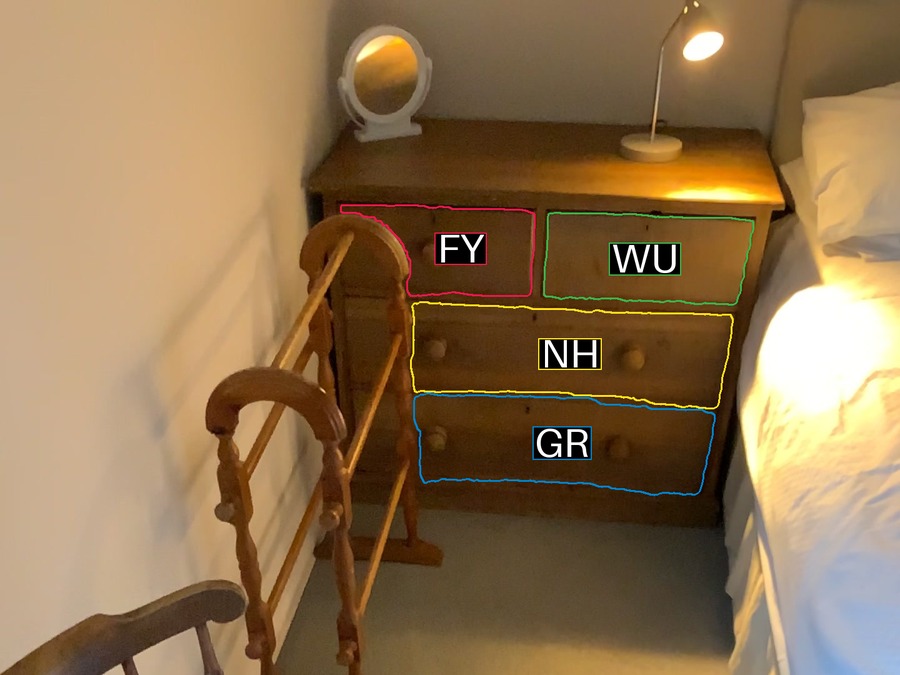}\note{VLM input}
      & \select{bottom drawer of the cabinet with the mirror and lamp on top}
              {open the bottom drawer of the cabinet with the mirror and lamp on top}{GR}
      & \includegraphics[width=\linewidth]{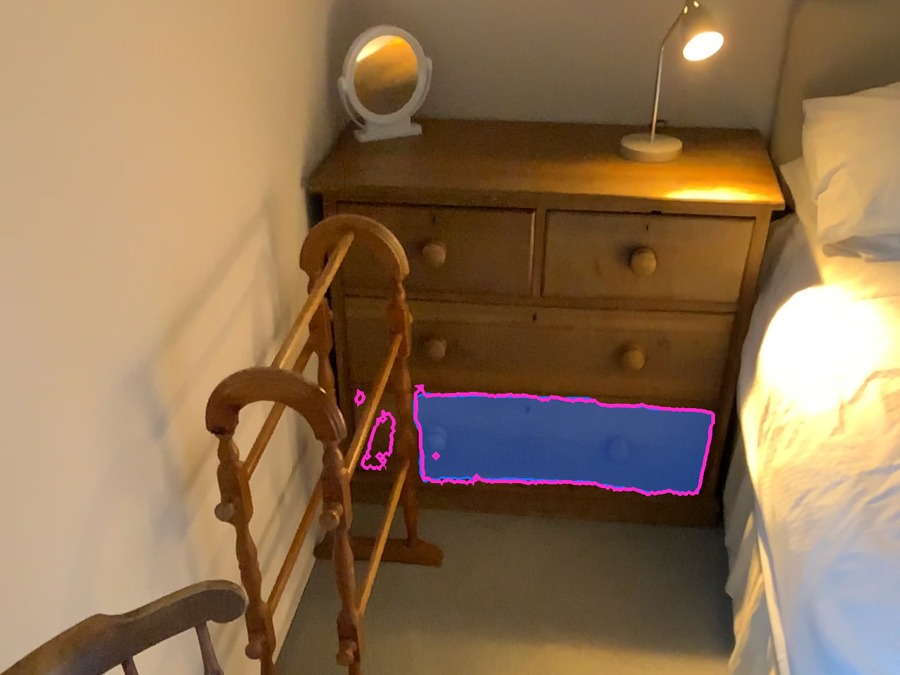}\note{IoU 0.91} \tabularnewline\midrule
    \parse{Open the left cabinet door above the computer}{door}
          {left door of the cabinet}
      & \includegraphics[width=\linewidth]{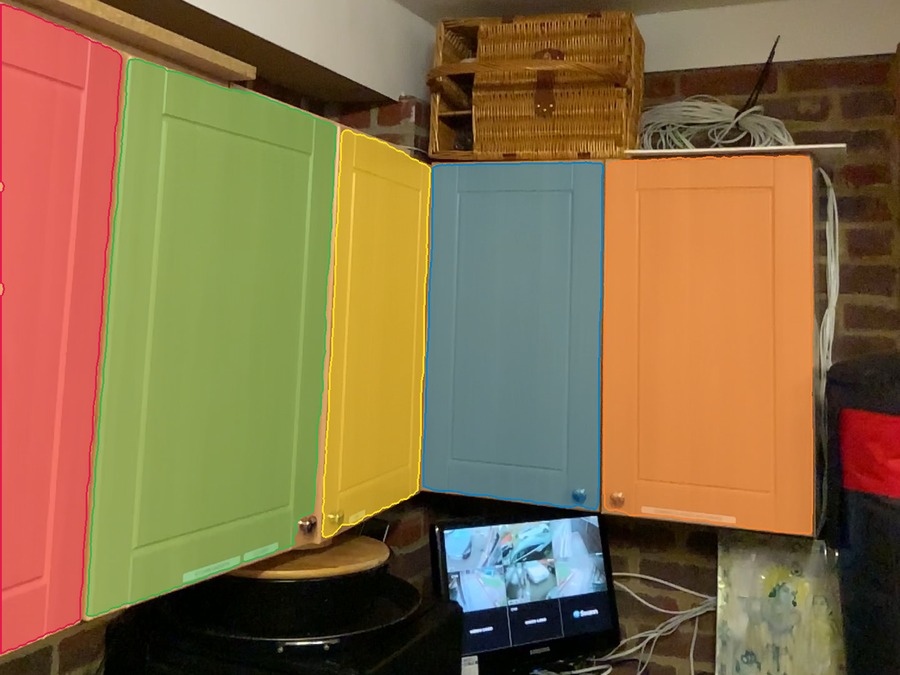}\note{5 candidates} & \includegraphics[width=\linewidth]{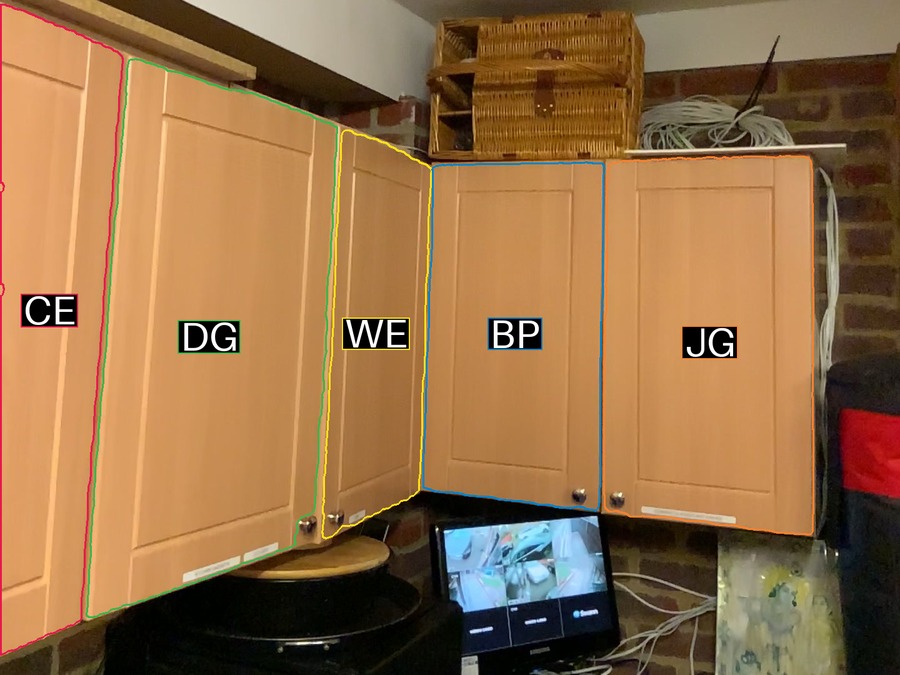}\note{VLM input}
      & \select{left door of the cabinet}{open the left cabinet door above the computer}{BP}
      & \includegraphics[width=\linewidth]{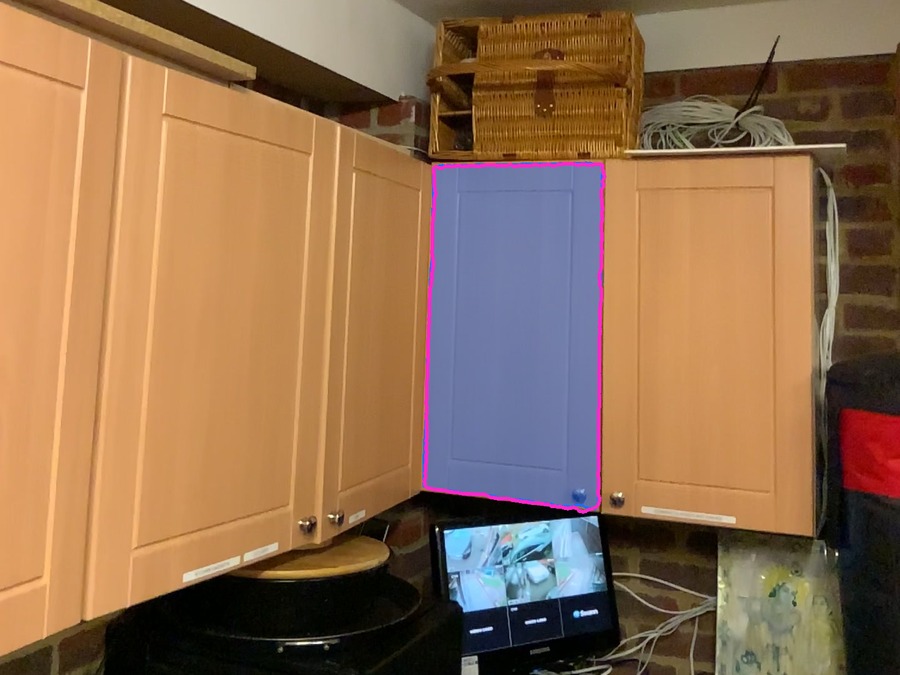}\note{IoU 0.98} \tabularnewline\midrule
    \parse{Open the door to the left of the TV stand}{door}
          {left door}
      & \includegraphics[width=\linewidth]{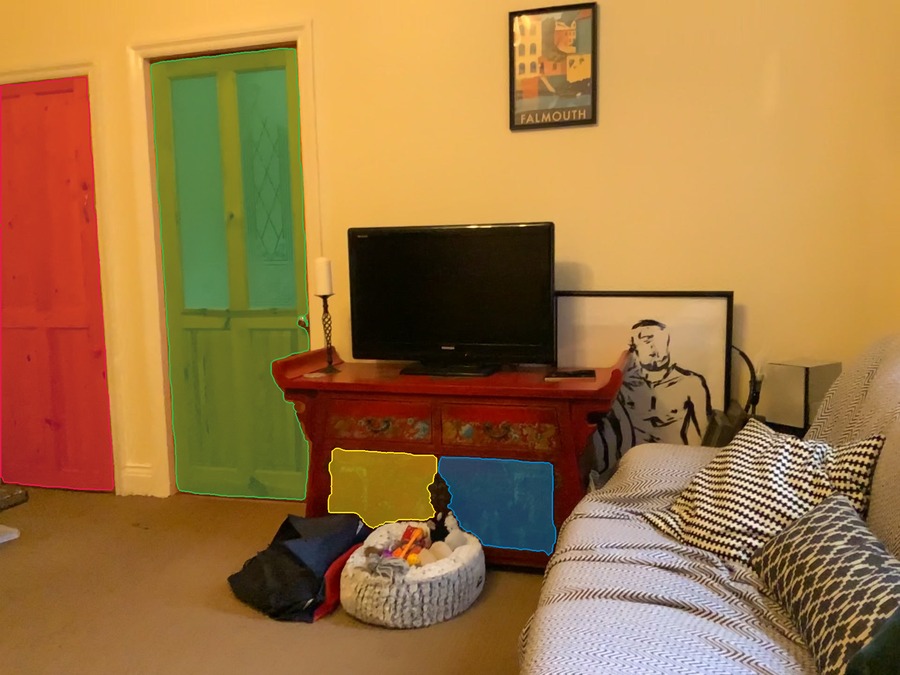}\note{4 candidates} & \includegraphics[width=\linewidth]{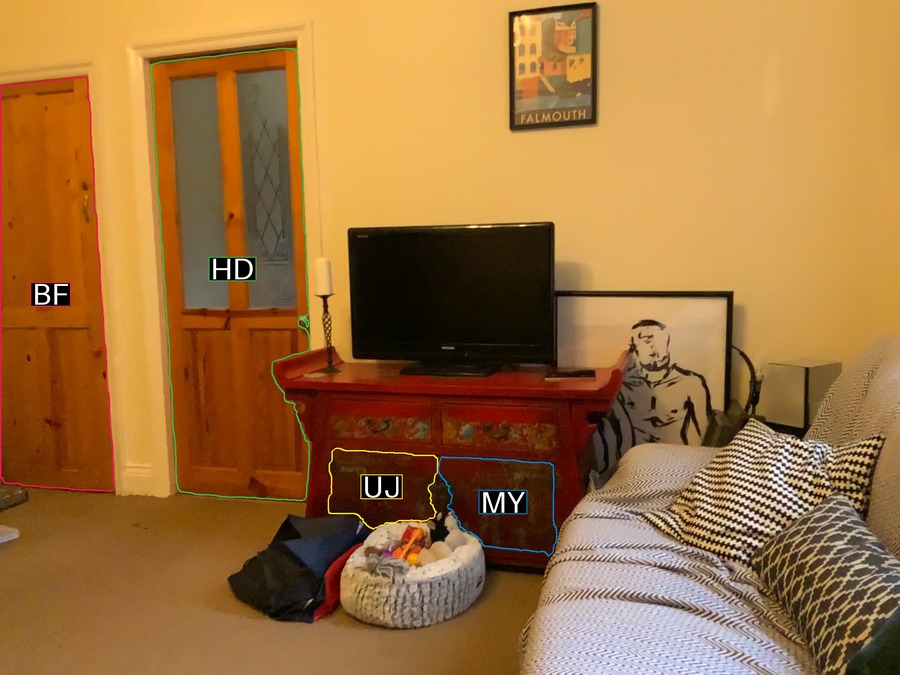}\note{VLM input}
      & \select{left door}{open the door to the left of the TV stand}{BF}
      & \includegraphics[width=\linewidth]{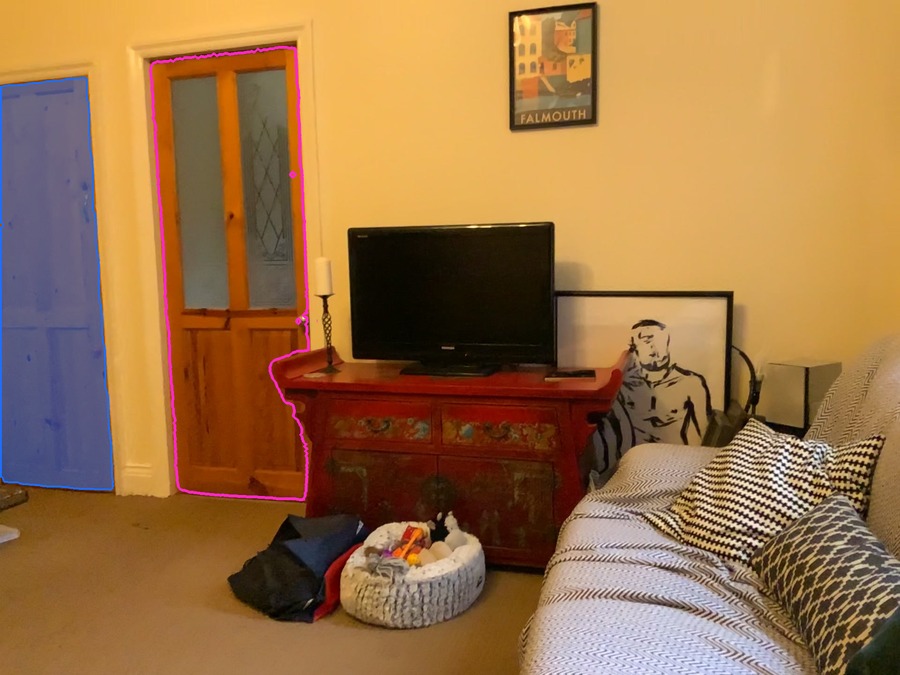}\note{IoU 0.00 (failure)} \tabularnewline
  \end{tabular}

  \vspace{-3mm}
  \caption{
  Mask extraction examples.
  Left to right: the prompt \prompt parsed by the LLM into the
  part category $d$ and target phrase $t$; SAM~3 candidates for $d$ after cleaning and
  de-duplication; the Set-of-Mark image with random two-letter IDs, given to the VLM together with
  the selection question (the fixed preamble describing the marks is omitted); the VLM answer; the
  predicted mask \mask (blue) and the ground-truth mask \gtmask (magenta contour), with mask IoU. The last row is a failure: the correct door (HD) is among the candidates, but
  the VLM picks the leftmost door (BF). Images are
  crops around the target; the VLM sees the full frame.
  }
  \label{fig:mask_extraction}
\end{figure*}

Given an RGB image \image of size $H \times W$ and a prompt \prompt, we predict the binary mask
\mask $\in \{0,1\}^{H \times W}$ of the part to manipulate, zero-shot and without task-specific training.
Prompts identify the part through a category (\eg~drawer, door), its position among identical siblings (ordinals, left/right, top/bottom) and landmarks around the owning object (\ie~the object that contains the target object, such as the dresser of a drawer).
Open-vocabulary segmenters produce accurate masks for a category but do not resolve such references, whereas VLMs can resolve them but do not output pixel-accurate masks.
We therefore split the problem into \emph{what} to segment and \emph{which} instance is meant, using only off-the-shelf models (\cref{fig:mask_extraction}). 
The steps are:
(i) Parsing: an LLM reads \prompt and extracts the part category $d$, used as the segmentation query, and a target phrase $t$, used to query the VLM selector.
(ii) Proposals: SAM~3, queried with $d$, proposes every instance of the part in the image \image. 
The proposals are cleaned, de-duplicated and ordered into a candidate set, keeping overlaps so that nested regions remain selectable. 
(iii) Selection: the candidates are drawn on \image as Set-of-Mark regions with random two-letter IDs, and a VLM answers with the ID of the candidate that has to be manipulated to carry out \prompt. 
The VLM only chooses among masks, so mask quality comes from SAM~3 and
instance disambiguation from the VLM.
We specify the models used and thresholds applied in each step:

\noindent \textbf{Parsing.}
An LLM (Claude Haiku~5.5~\cite{anthropic2026haiku55}), given \prompt, returns $(d, t)$: $d$ is a singular noun for the target object (excluding the owning object and attached sub-parts such as handles), and $t$ a target phrase made of the target object, its position words and its owning object, without relations to other objects.

\noindent \textbf{Proposals.}
SAM~3~\cite{carion2025sam3segmentconcepts} is run on the full frame with $d$ as the text query. We keep instances with confidence score $\ge 0.4$, drop boxes smaller than $4$\,px, apply greedy NMS (IoU $0.7$) in increasing box-area order, so parts are kept over the objects containing them, and retain at most $30$ boxes, each with its SAM~3 mask.
Each mask is reduced to its largest connected component and its holes are filled, but it is discarded if that component covers less than $50\%$ of the mask. 
Masks with IoU $\ge 0.7$ to an earlier candidate are merged into it. 
Candidates may overlap. 

\noindent \textbf{Selection.}
Candidates are drawn Set-of-Mark style~\cite{yang2023som} on the full frame, upscaled to a short side of $\ge 768$\,px: one colored outline per mask, with a label at the mask centroid. 
Labels are random two-letter IDs (distinct letters from a 22-letter alphabet, seeded per sample), which avoids confusing them with ordinals in \prompt. 
A VLM (Claude Haiku~5.5~\cite{anthropic2026haiku55}, high reasoning effort) is asked:
\textit{``Which label marks the $\langle t \rangle$ that has to be manipulated in order to $\langle \mathcal{P} \rangle$? Answer with only the label.''} 
The VLM returns a candidate label, and its mask is used as the predicted mask \mask.

\subsection{Motion calibration score}\label{sec:motion_calib}

\begin{figure}[t]
  \centering
  \captionsetup[subfigure]{font=scriptsize, labelformat=empty, skip=1pt, justification=centering}
  \definecolor{corrmu}{HTML}{0072B2}
  \definecolor{corrR}{HTML}{E69F00}
  \definecolor{corrnu}{HTML}{CC79A7}
  \newcommand{\corrbars}[3]{%
    \makebox[\linewidth]{\begin{tikzpicture}[x=0.2667\linewidth, y=0.45cm]
      \foreach \v/\c/\l [count=\i from 0] in {#1/corrmu/\mu, #2/corrR/r, #3/corrnu/\nu} {
        \fill[\c] (\i+0.15,0) rectangle (\i+0.85,\v);
        \node[font=\tiny, above, inner sep=1pt] at (\i+0.5,\v) {\v};
        \node[font=\tiny, below, inner sep=1pt] at (\i+0.5,0) {$\l$};
      }
      \draw[gray] (0,0) -- (3,0);
    \end{tikzpicture}}}
  \begin{subfigure}[t]{0.245\linewidth}
    \caption[Static]{(a) Static}
    \begin{overpic}[width=\linewidth]{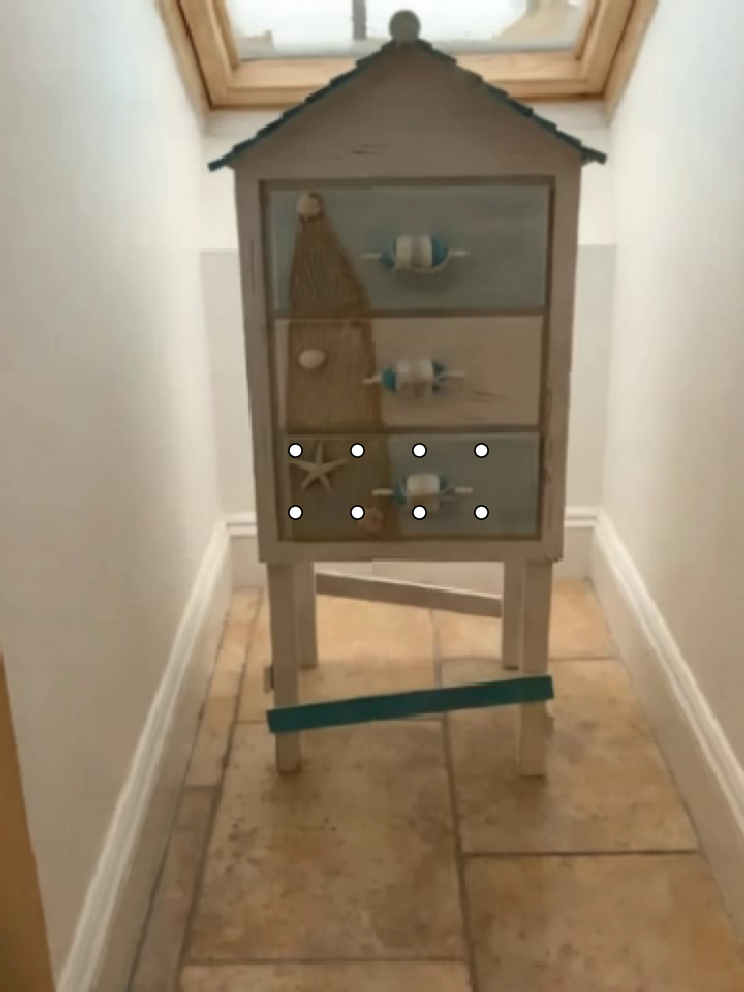}
    \end{overpic}\par\vspace{1pt}
    \corrbars{0.01}{1.00}{1.00}
  \end{subfigure}\hfill
  \begin{subfigure}[t]{0.245\linewidth}
    \caption[Scattered]{(b) Scattered}
    \begin{overpic}[width=\linewidth]{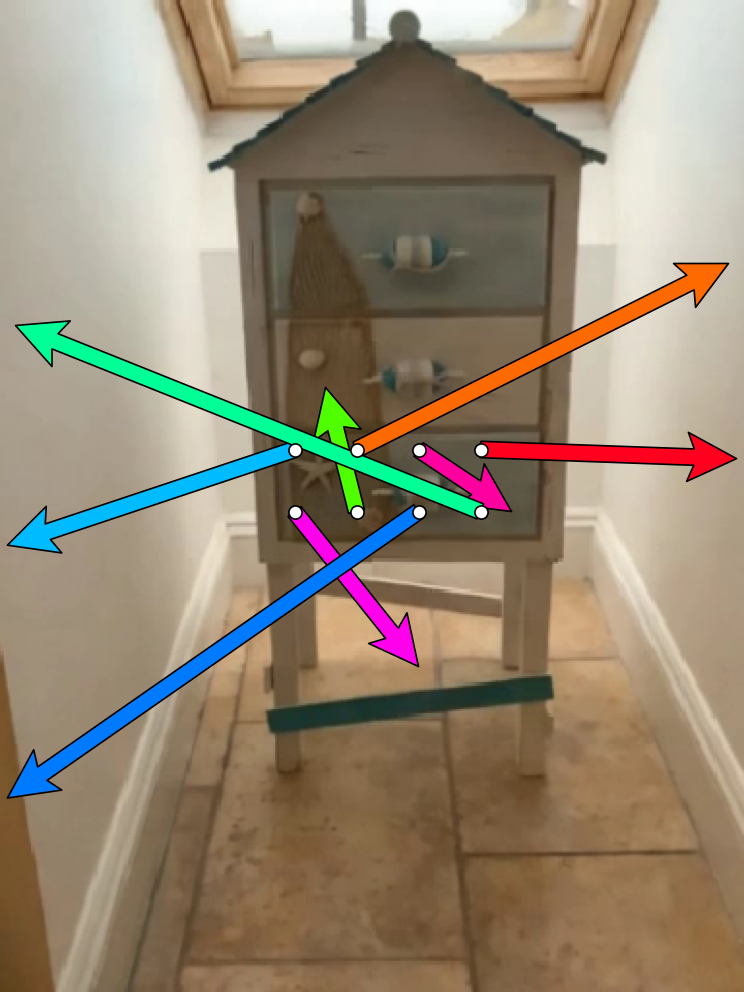}
    \end{overpic}\par\vspace{1pt}
    \corrbars{0.97}{0.06}{0.91}
  \end{subfigure}\hfill
  \begin{subfigure}[t]{0.245\linewidth}
    \caption[Sink]{(c) Sink}
    \begin{overpic}[width=\linewidth]{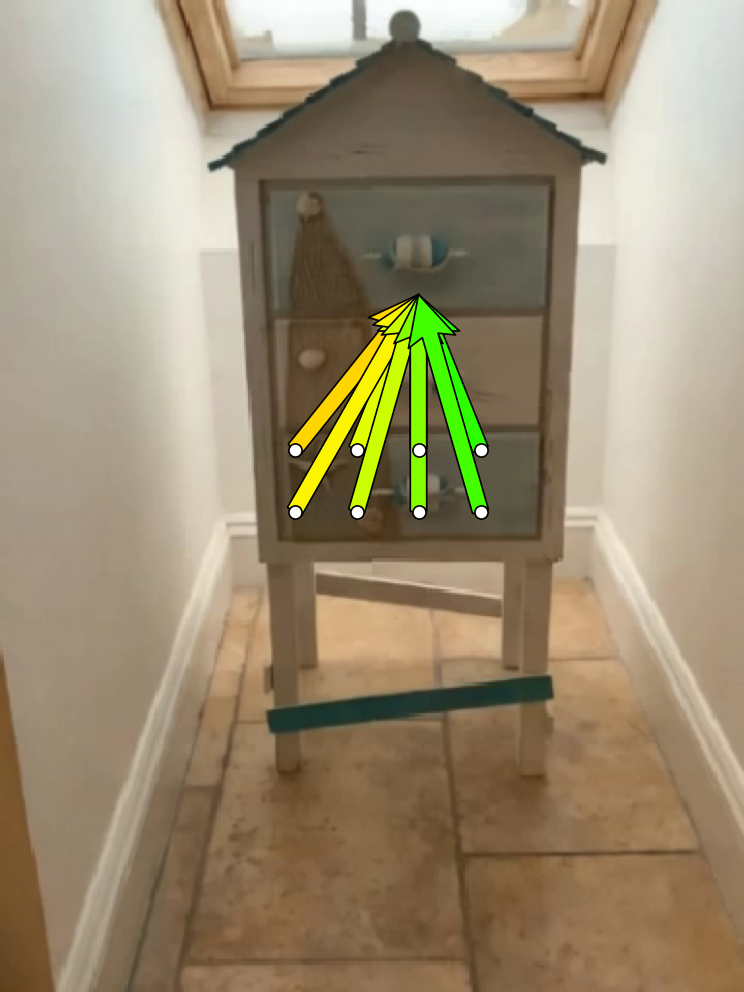}
    \end{overpic}\par\vspace{1pt}
    \corrbars{1.00}{0.95}{0.14}
  \end{subfigure}\hfill
  \begin{subfigure}[t]{0.245\linewidth}
    \caption[Good]{(d) Good}
    \begin{overpic}[width=\linewidth]{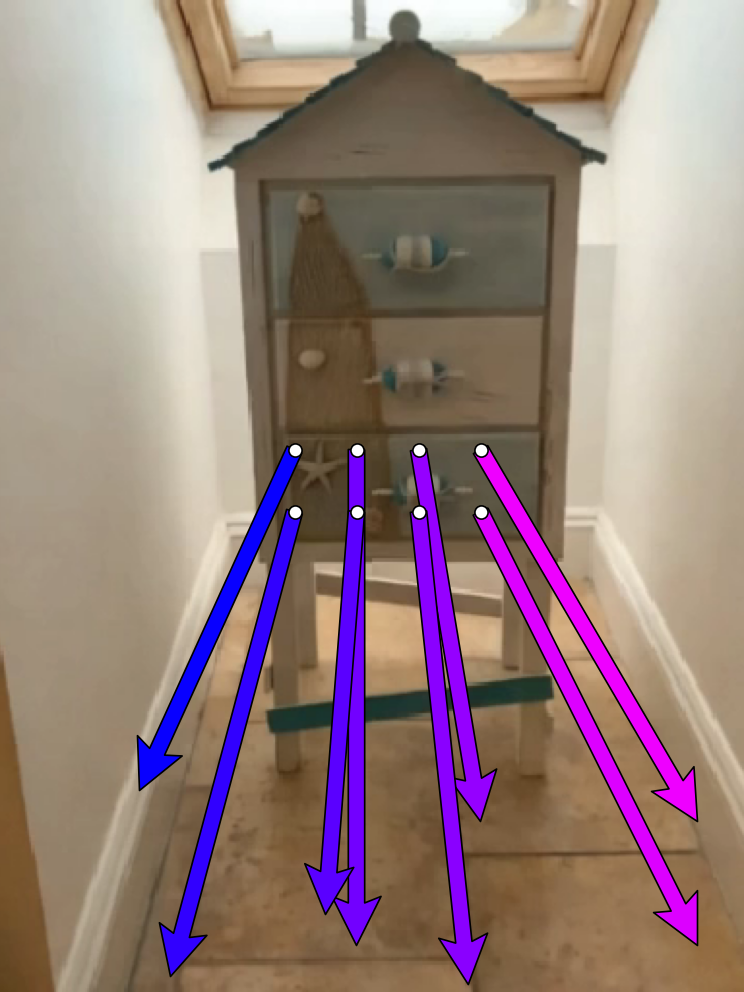}
    \end{overpic}\par\vspace{1pt}
    \corrbars{1.00}{0.94}{0.95}
  \end{subfigure}

  \vspace{2pt}
  \begin{subfigure}[t]{0.245\linewidth}
    \begin{overpic}[width=\linewidth]{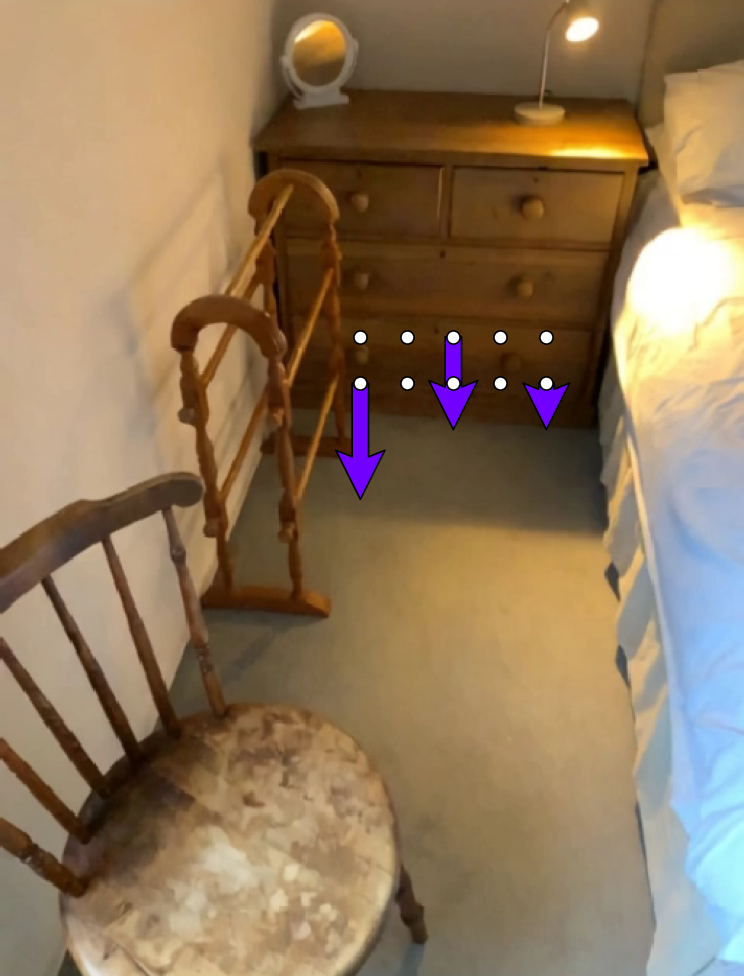}
    \end{overpic}\par\vspace{1pt}
    \corrbars{0.12}{1.00}{0.81}
  \end{subfigure}\hfill
  \begin{subfigure}[t]{0.245\linewidth}
    \begin{overpic}[width=\linewidth]{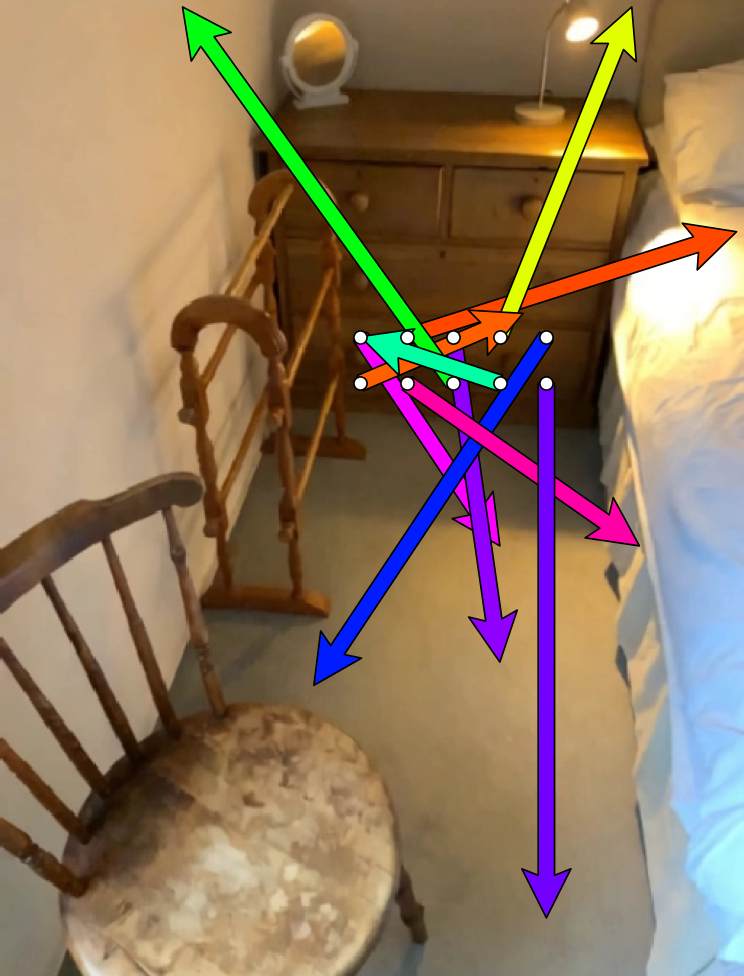}
    \end{overpic}\par\vspace{1pt}
    \corrbars{0.97}{0.05}{0.73}
  \end{subfigure}\hfill
  \begin{subfigure}[t]{0.245\linewidth}
    \begin{overpic}[width=\linewidth]{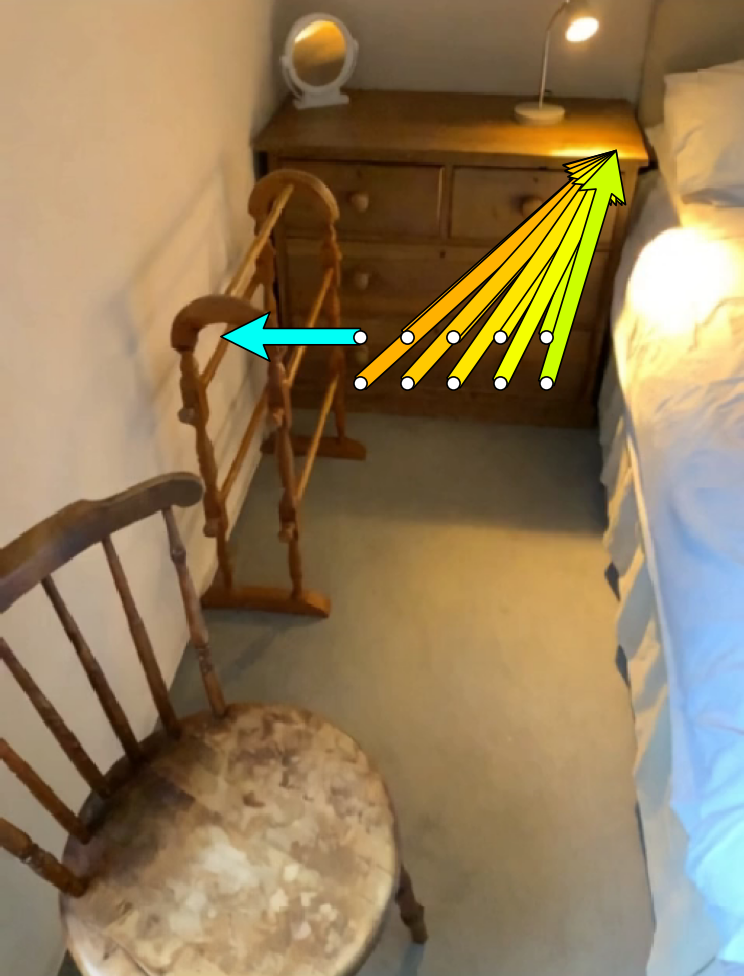}
    \end{overpic}\par\vspace{1pt}
    \corrbars{1.00}{0.96}{0.16}
  \end{subfigure}\hfill
  \begin{subfigure}[t]{0.245\linewidth}
    \begin{overpic}[width=\linewidth]{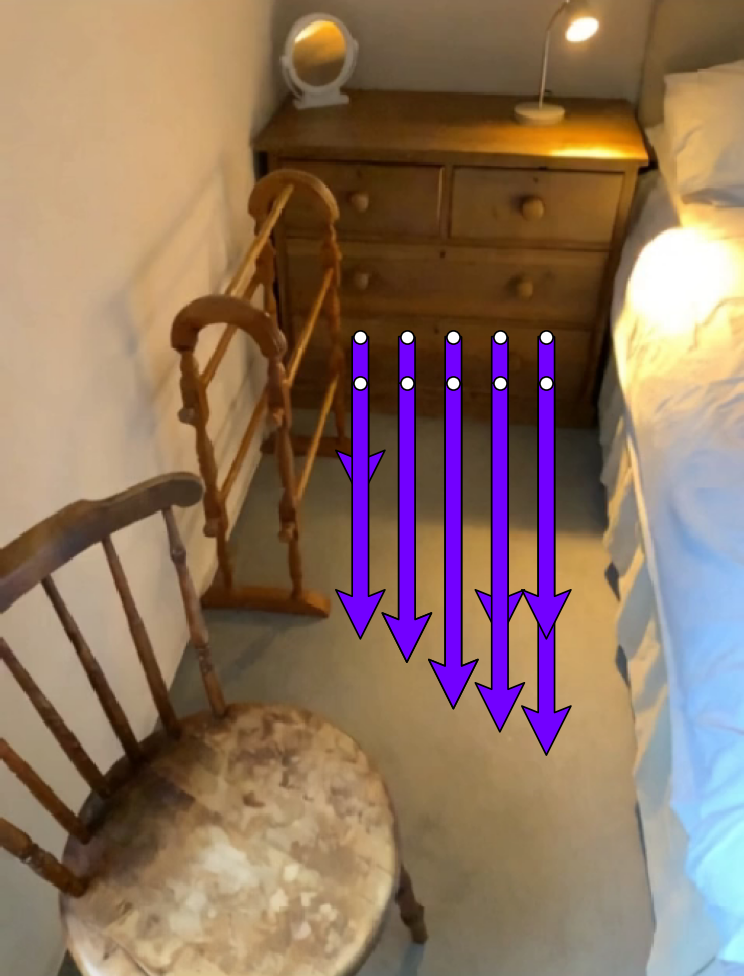}
    \end{overpic}\par\vspace{1pt}
    \corrbars{0.77}{1.00}{0.72}
  \end{subfigure}

  \vspace{2pt}
  \begin{subfigure}[t]{0.245\linewidth}
    \begin{overpic}[width=\linewidth]{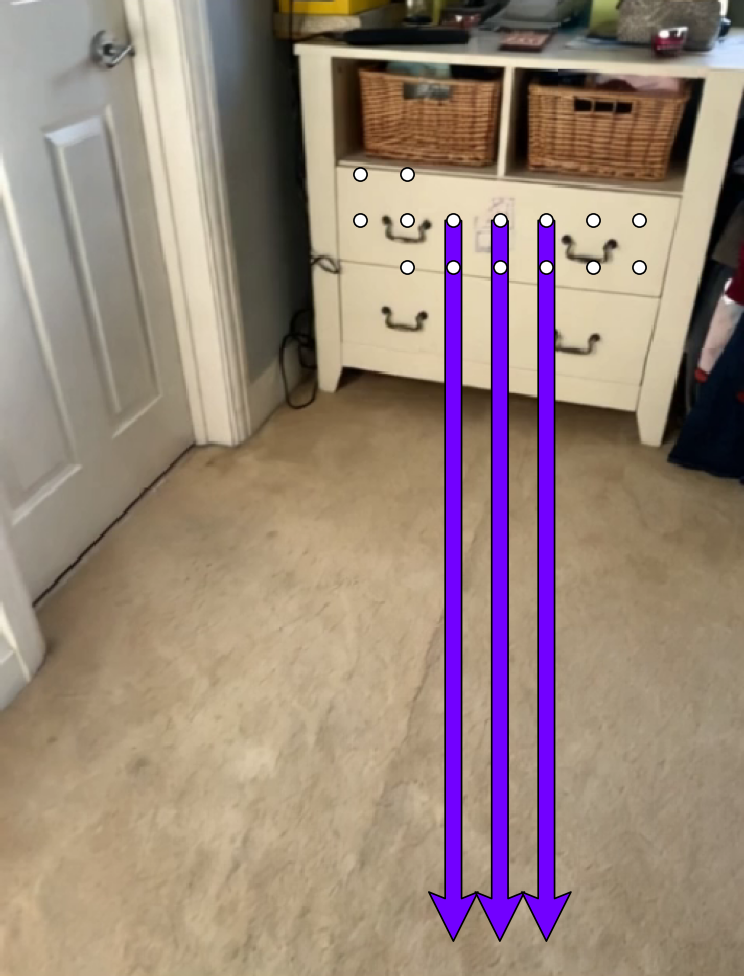}
    \end{overpic}\par\vspace{1pt}
    \corrbars{0.24}{1.00}{0.99}
  \end{subfigure}\hfill
  \begin{subfigure}[t]{0.245\linewidth}
    \begin{overpic}[width=\linewidth]{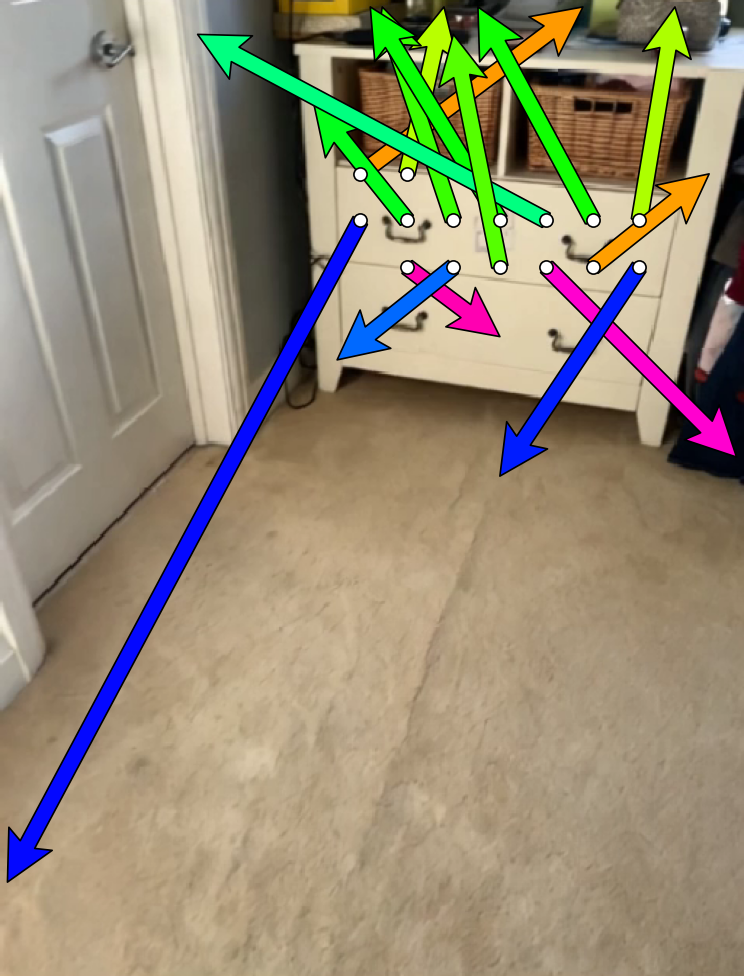}
    \end{overpic}\par\vspace{1pt}
    \corrbars{1.00}{0.06}{0.69}
  \end{subfigure}\hfill
  \begin{subfigure}[t]{0.245\linewidth}
    \begin{overpic}[width=\linewidth]{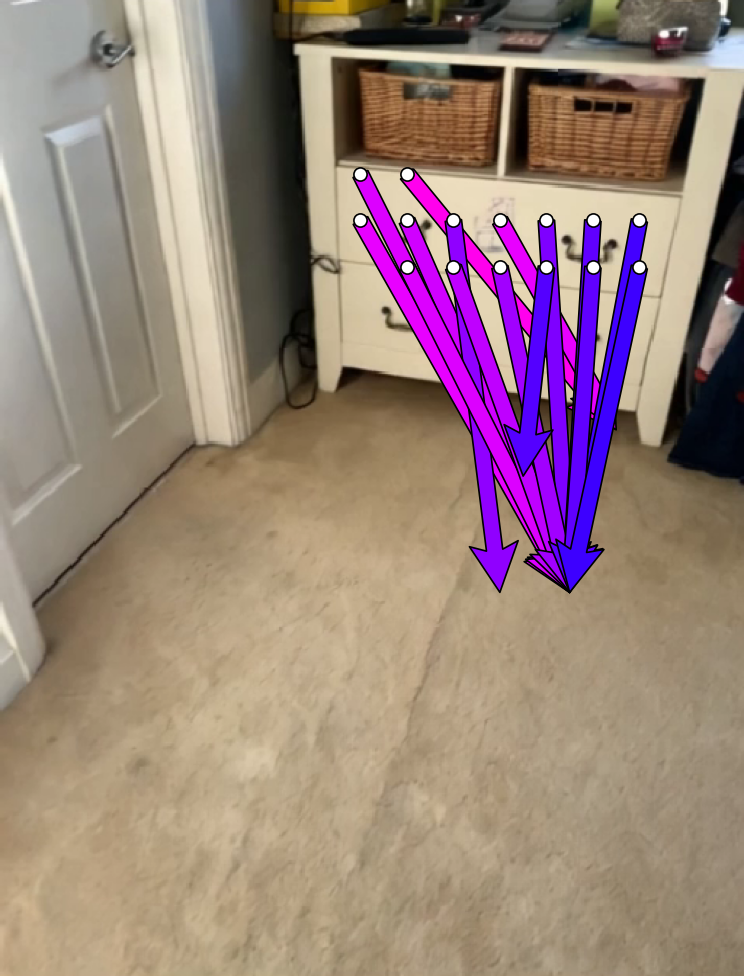}
    \end{overpic}\par\vspace{1pt}
    \corrbars{1.00}{0.91}{0.07}
  \end{subfigure}\hfill
  \begin{subfigure}[t]{0.245\linewidth}
    \begin{overpic}[width=\linewidth]{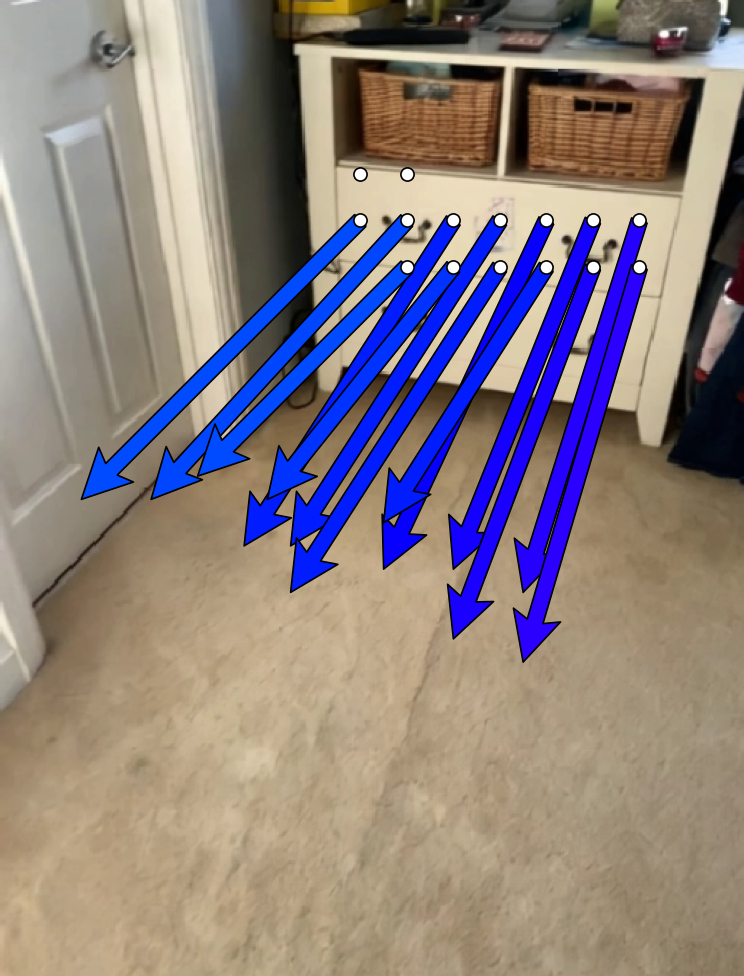}
    \end{overpic}\par\vspace{1pt}
    \corrbars{0.83}{0.90}{0.87}
  \end{subfigure}

  \vspace{-2mm}
  \caption{
  Correspondence maps of different attention heads on three scenes (rows).
  (a) Static head with low moving fraction $\mu$,
  (b) Scattered head with low consistency $r$, and
  (c) Sink head with low unique rate $\nu$.
  (d) High-quality head, with all scores high.
  Arrow color encodes the correspondence direction.
  }
  \label{fig:correspondence_examples}
\end{figure}

\begin{figure}[t]
  \centering
  \newlength{\colw}\setlength{\colw}{0.24\linewidth}
  \setlength{\tabcolsep}{1pt}
  \renewcommand{\arraystretch}{0.5}
  \begin{tabular}{cccc}
    {\scriptsize Layer 44, head 5} & {\scriptsize Layer 48, head 1} & {\scriptsize Layer 30, head 2} & {\scriptsize Layer 41, head 11} \\
    \includegraphics[width=\colw]{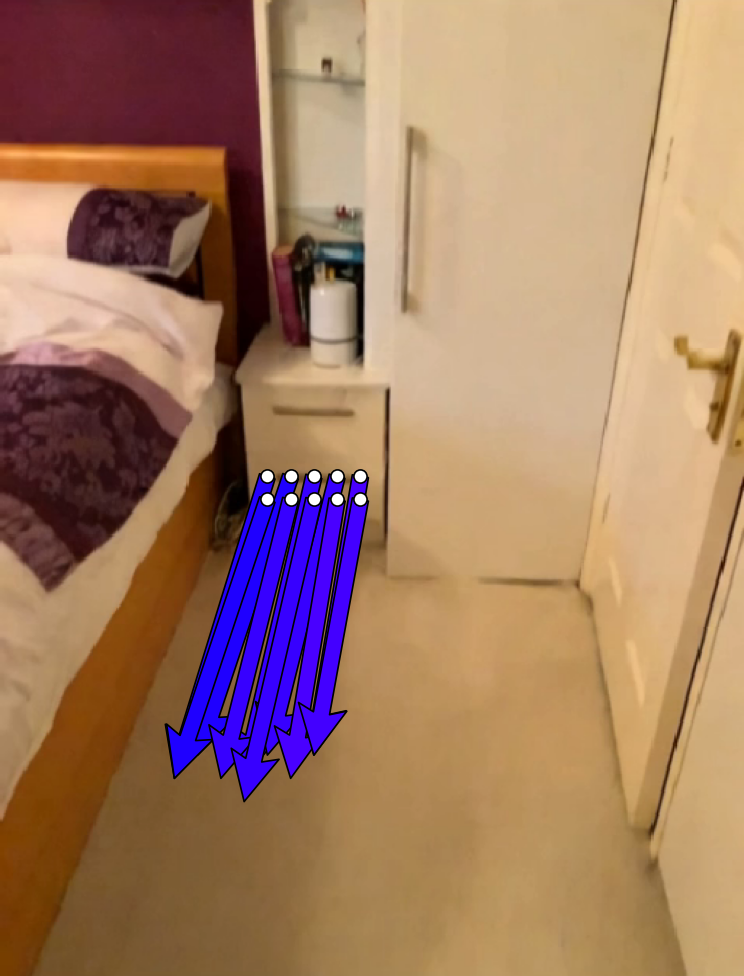} & \includegraphics[width=\colw]{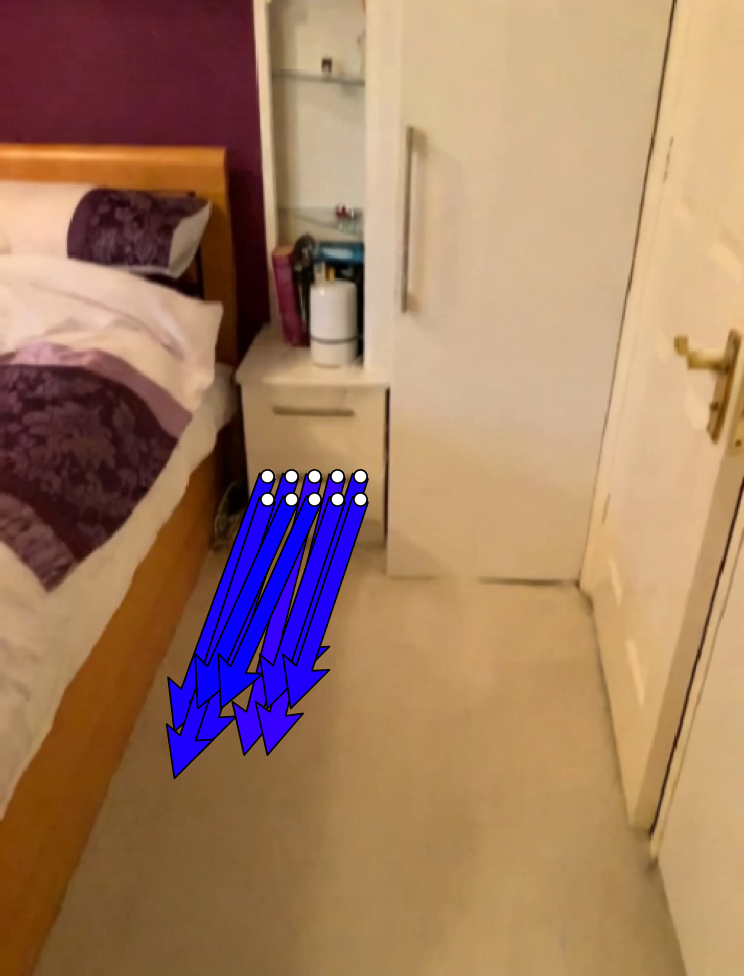} & \includegraphics[width=\colw]{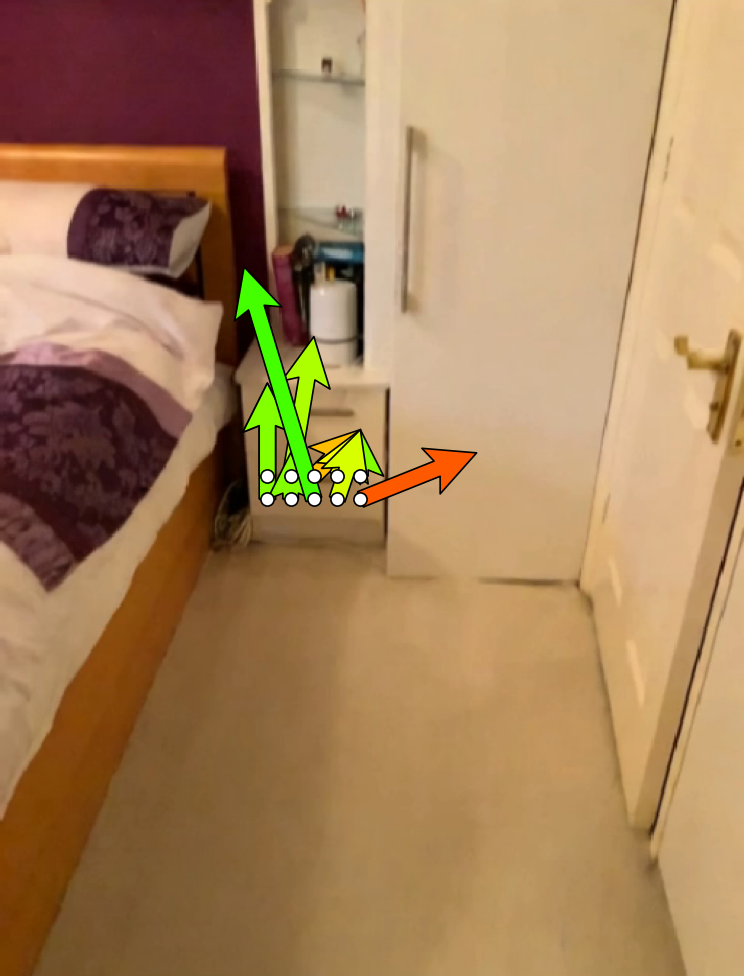} & \includegraphics[width=\colw]{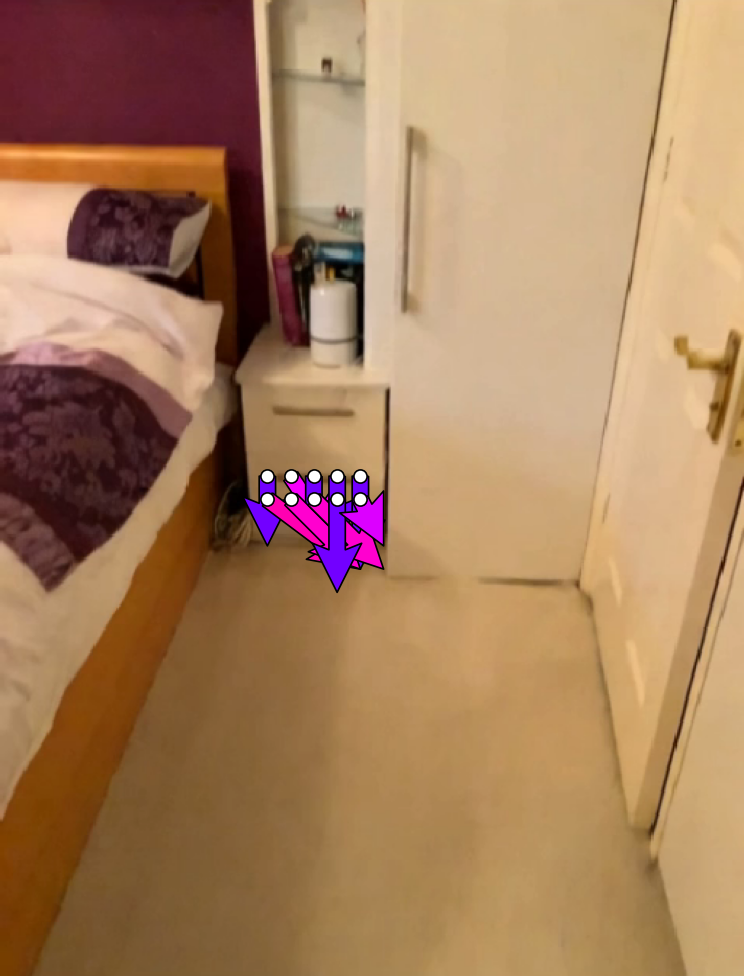} \\
    \includegraphics[width=\colw]{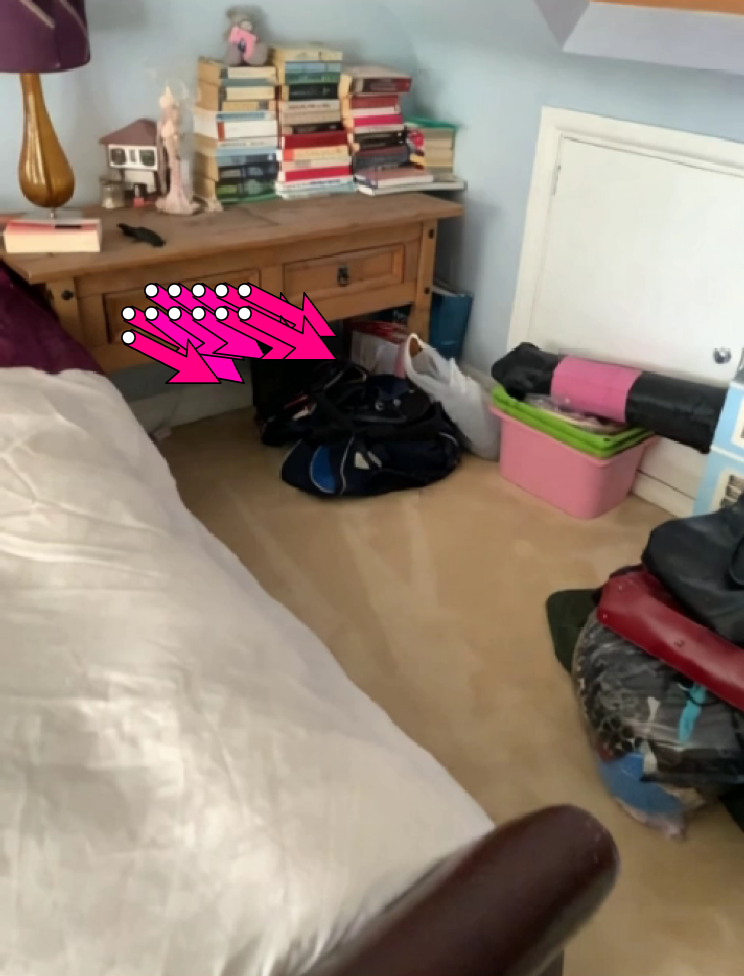} & \includegraphics[width=\colw]{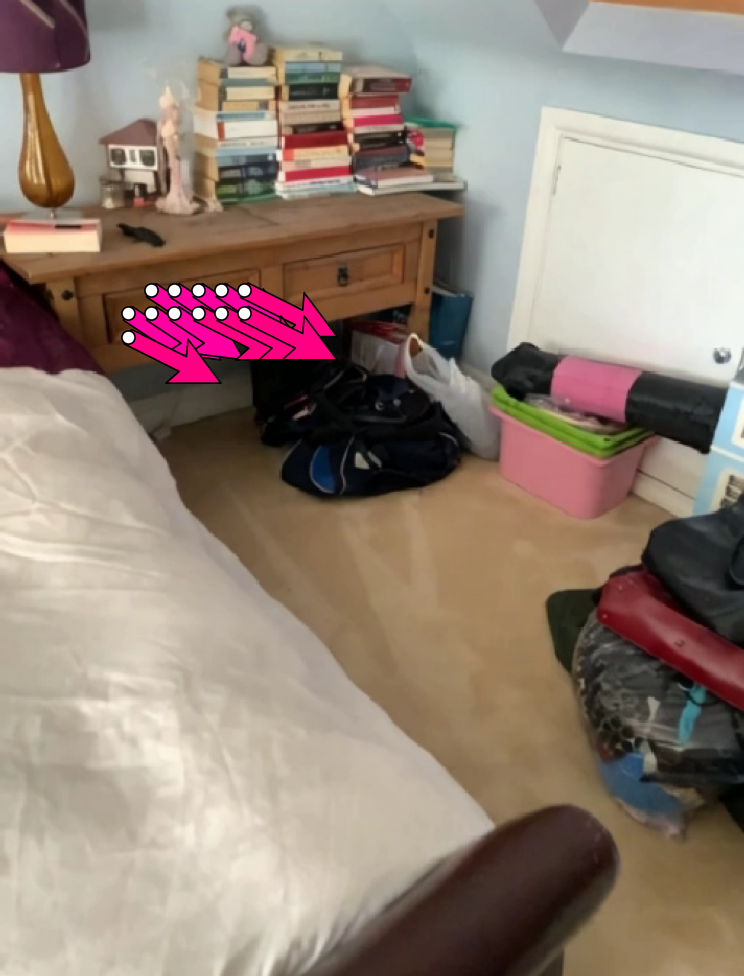} & \includegraphics[width=\colw]{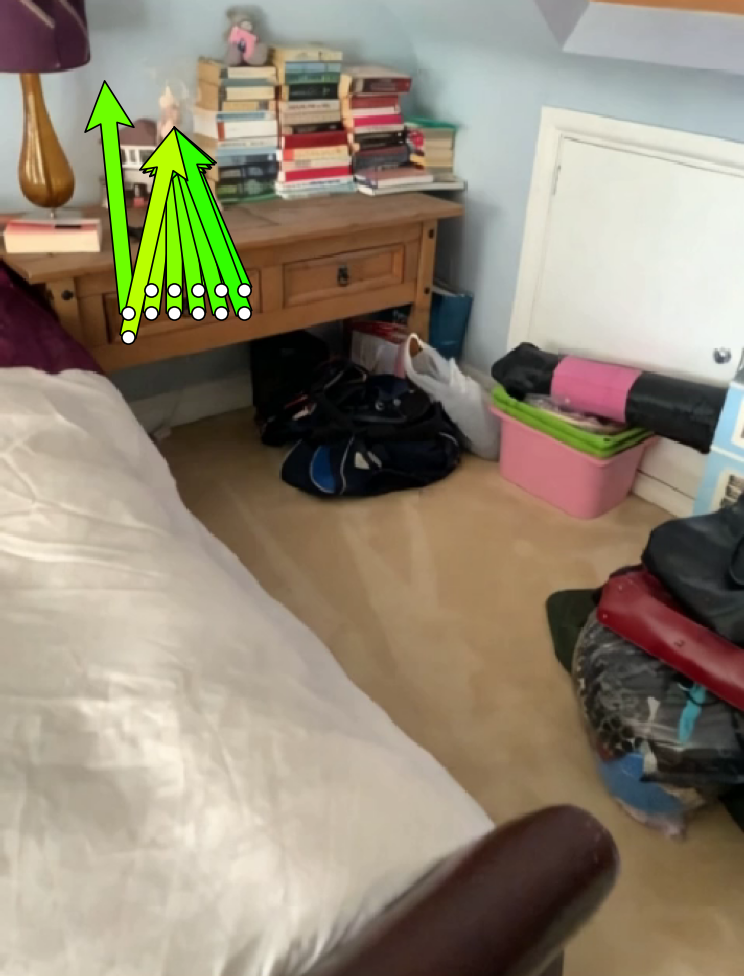} & \includegraphics[width=\colw]{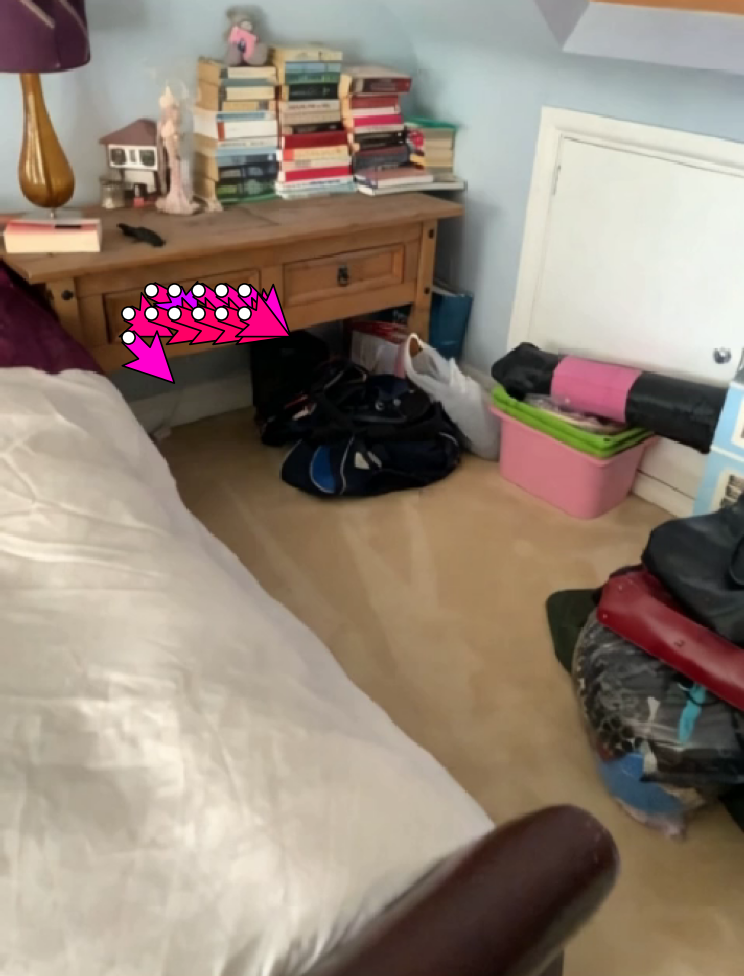} \\
    \includegraphics[width=\colw]{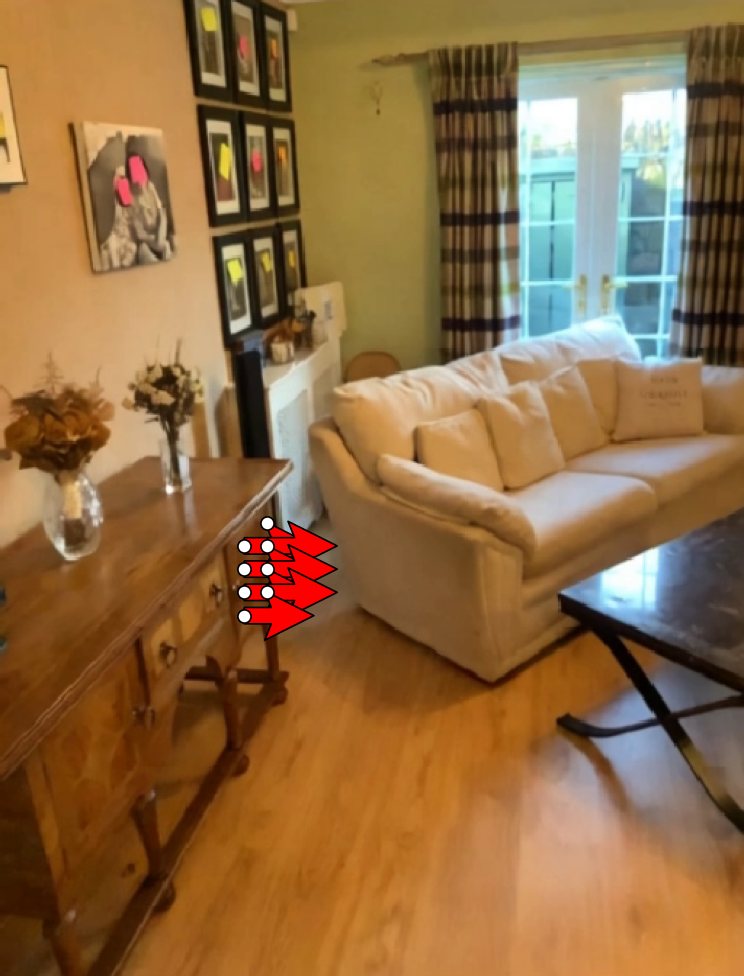} & \includegraphics[width=\colw]{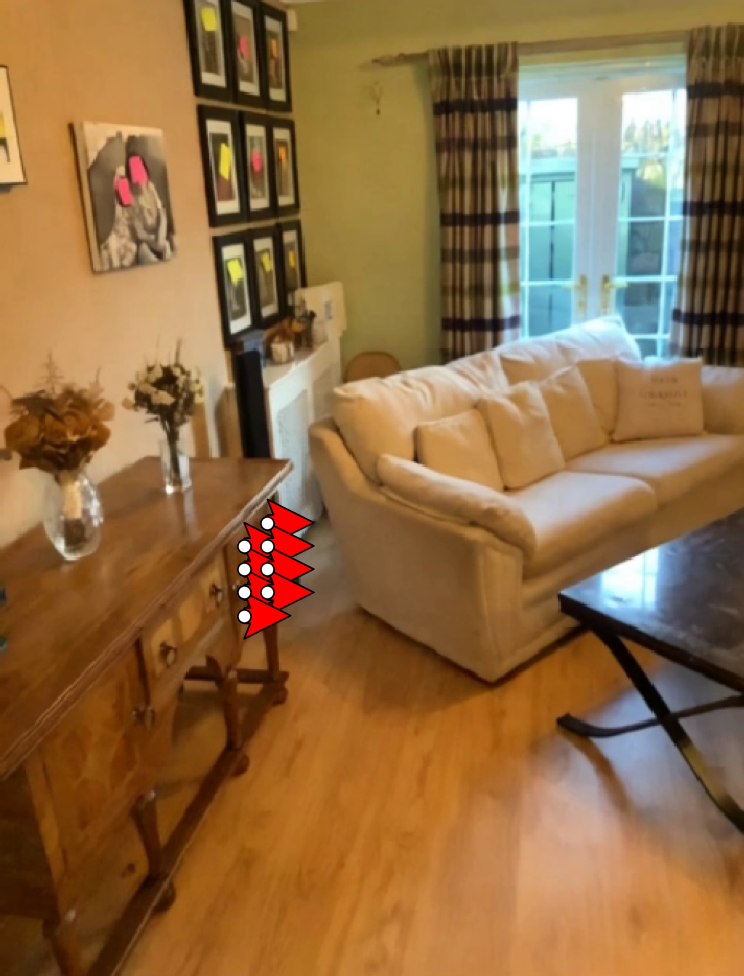} & \includegraphics[width=\colw]{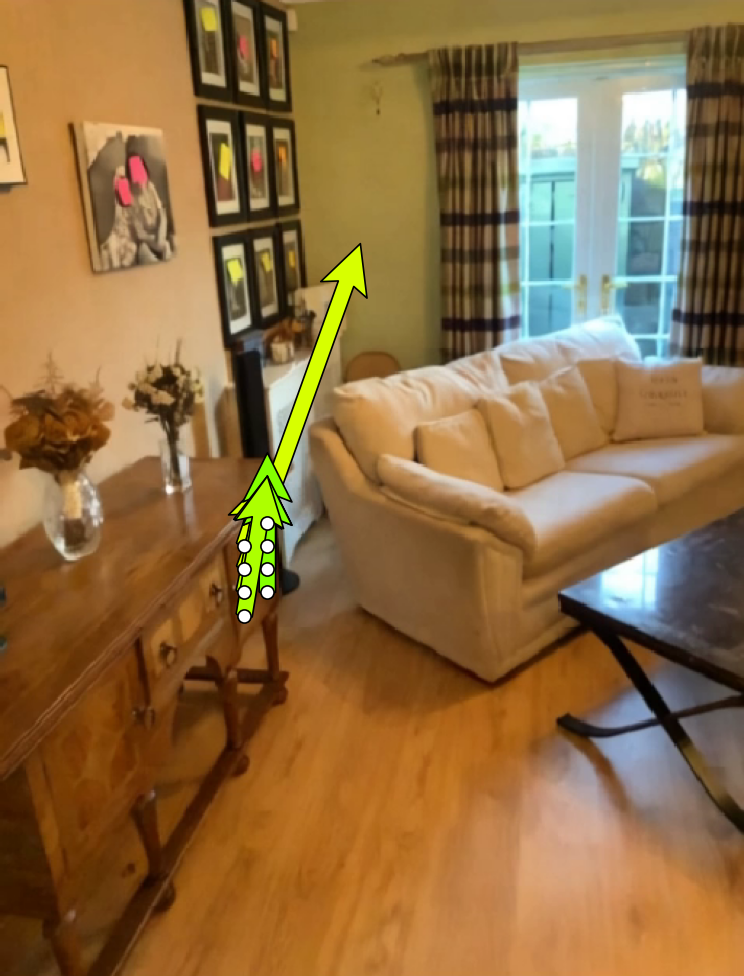} & \includegraphics[width=\colw]{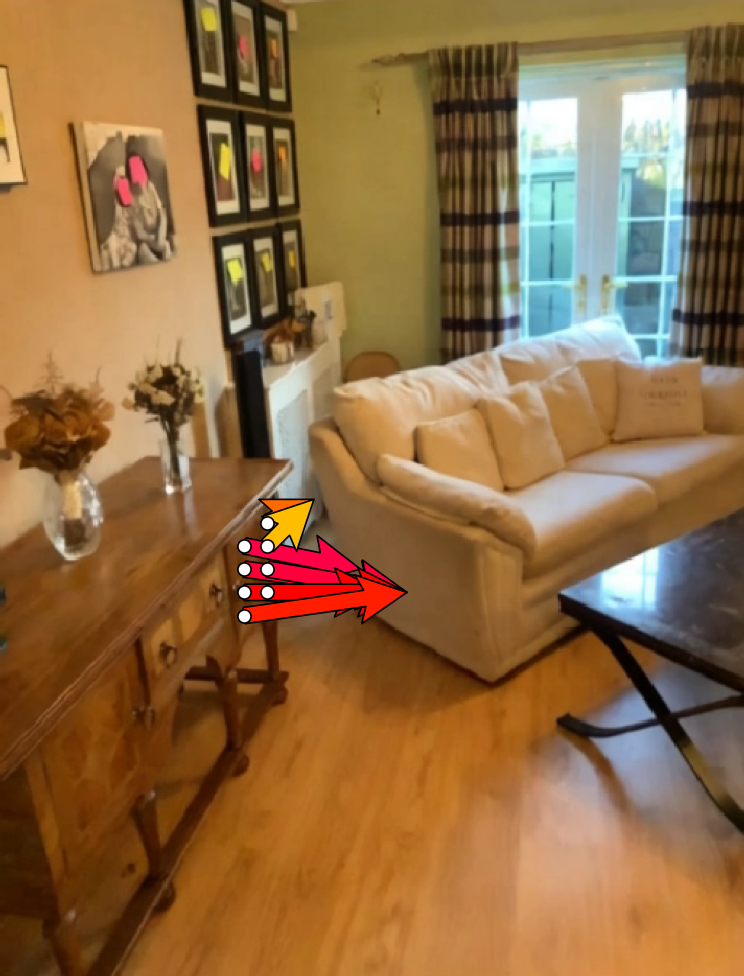} \\
    \includegraphics[width=\colw]{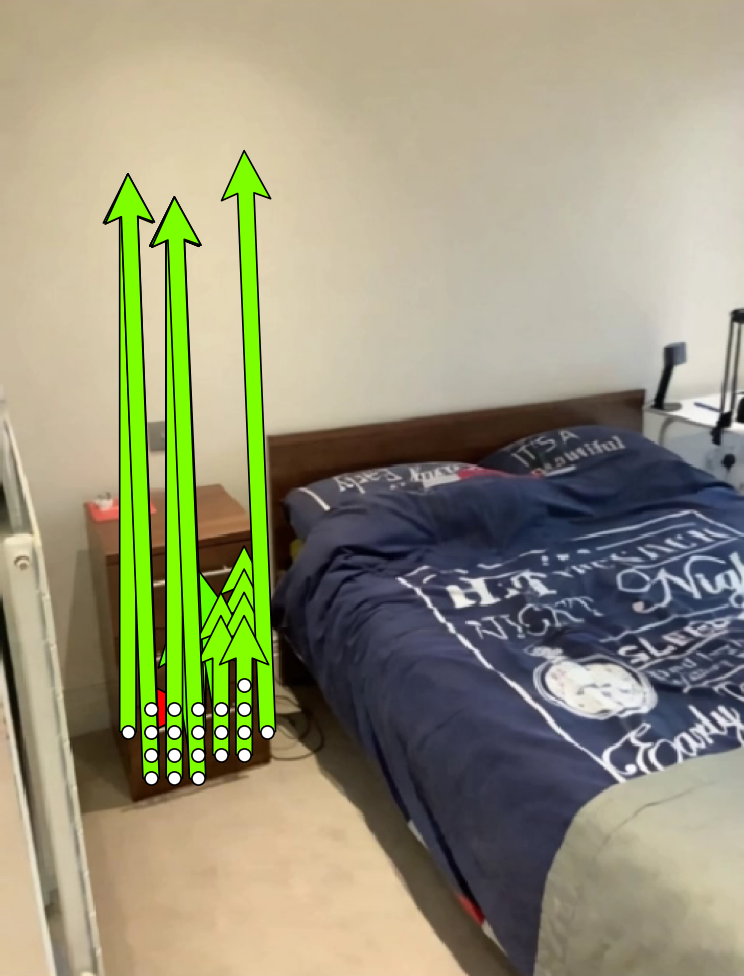} & \includegraphics[width=\colw]{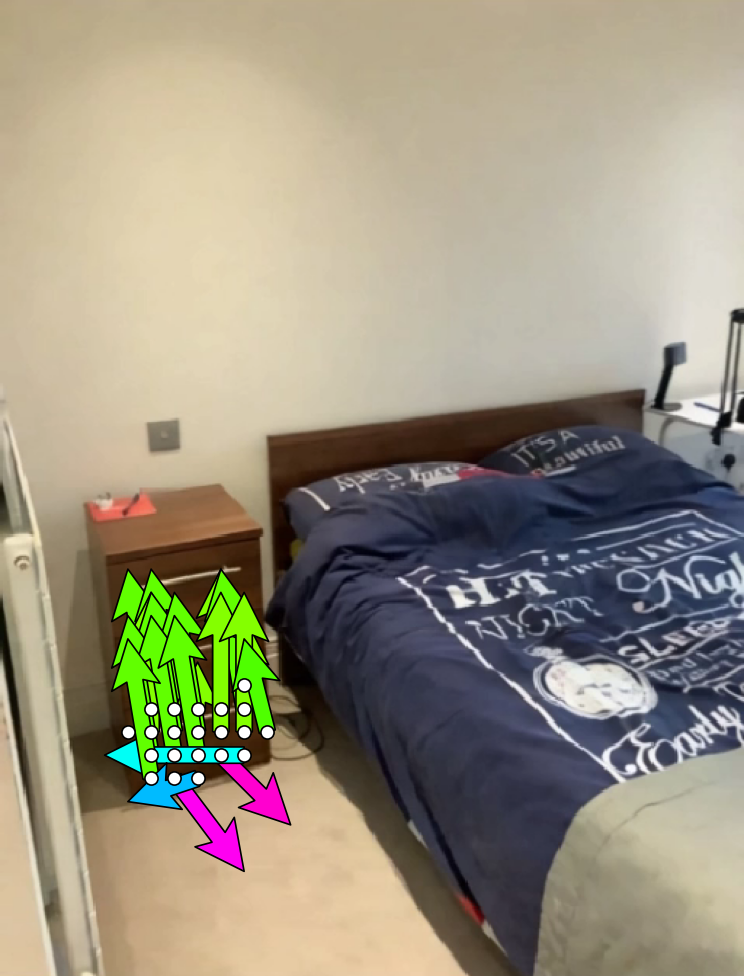} & \includegraphics[width=\colw]{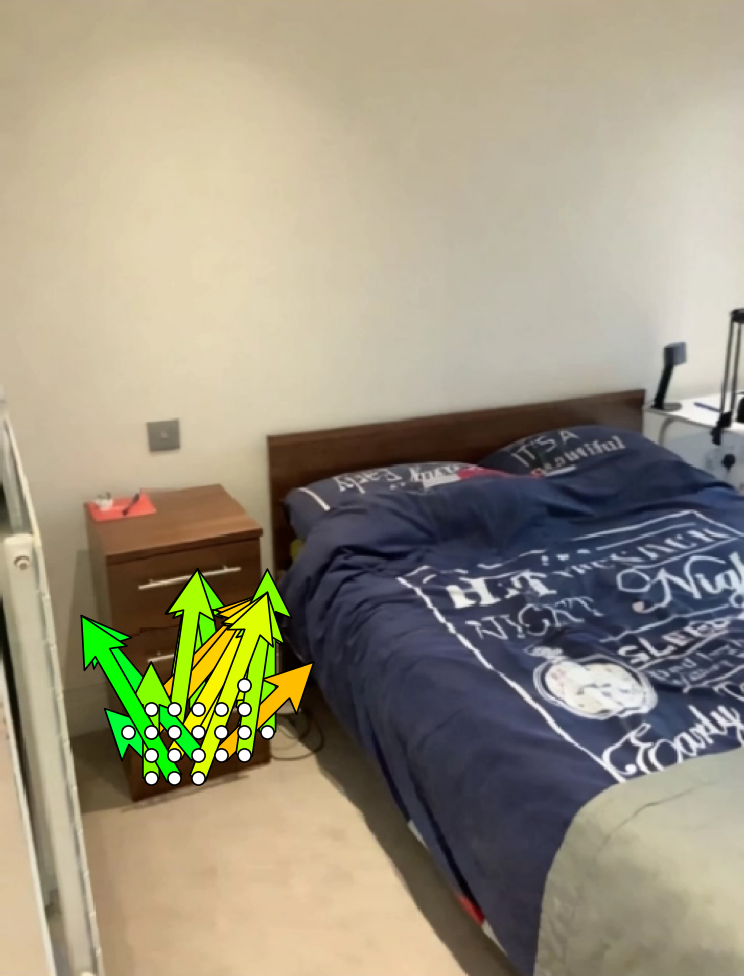} & \includegraphics[width=\colw]{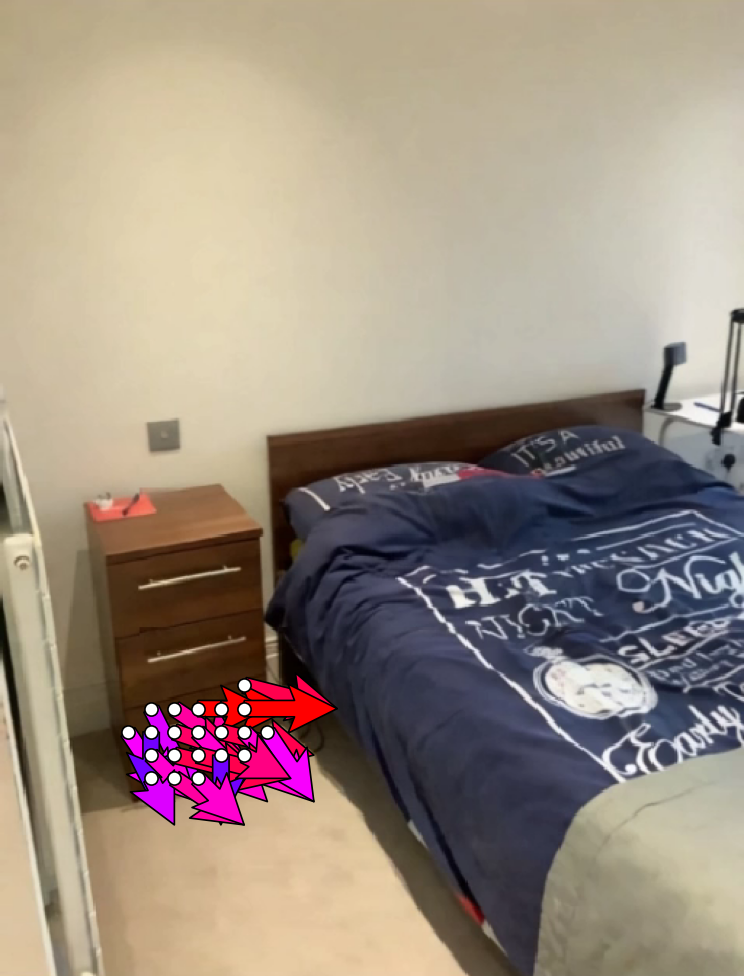} \\
    {\scriptsize $B{=}0.42$} & {\scriptsize $B{=}0.52$} & {\scriptsize $B{=}0.96$} & {\scriptsize $B{=}0.93$} \\
  \end{tabular}

  \vspace{-2mm}
  \caption{
  Example of positional bias of attention heads.
  Each column shows the correspondences of one head on the same four calibration scenes (rows), in which the target moves in different directions.
  Arrow color encodes the correspondence direction.
  }
  \label{fig:positional_bias}
\end{figure}

A motion head should behave like a point tracker on the target object: when the object moves, the patches on it should be matched to displaced cells, all displaced in the same direction, and each patch to its own cell.
We capture these three properties with the moving fraction $\mu$, the consistency $r$, and the unique rate $\nu$ (Sec.~4.1 of the main paper), each of which is low for one of three typical head failures.
A \emph{static} head matches every patch to its frame-0 position, so few correspondences move and $\mu$ is low.
A \emph{scattered} head moves the patches in random directions, so their directions disagree and $r$ is low.
A \emph{sink} head sends most patches to the same few cells, so few cells are distinct and $\nu$ is low.
A good head scores high on all three.
Since the score only looks at the object, a fourth failure remains: a head that always points in the same direction, whatever the scene, because of a positional bias.
We detect such heads after scoring, by comparing their directions across calibration scenes.

\noindent \textbf{Correspondences.}
We score each head $(\ell, h)$ at each denoising step $k$ on each calibration scene $s$ independently.
In the remainder of this section, we drop these indices when they are fixed.
Following Eq.~(2) of the main paper, every patch $p \in \Omega$ has a correspondence $m_t(p)$ and a displacement $d_t(p) = m_t(p) - m_0(p)$ in each latent frame $t$, in latent cells.
On the calibration set, $\Omega$ contains the latent cells covered for at least half of their area by $\mathcal{M}_{\mathrm{GT}}$.
We pool all patches and all latent frames after the first, $\Pi = \Omega \times \{1, \dots, T'-1\}$, and define a cell-timestep pair $(p, t) \in \Pi$ as moving when its displacement exceeds $\delta_{\mathrm{h}} = 2$ cells:
\begin{equation}
  \Pi_{\mathrm{m}} = \left\{ (p, t) \in \Pi \;:\; \lVert d_t(p) \rVert_2 \ge \delta_{\mathrm{h}} \right\}.
\end{equation}
The threshold ignores small displacements caused by the bias of many heads towards the cells close to the query's own position.
The moving fraction $\mu$ is defined as the share of pairs that move:
\begin{equation}
  \mu = \frac{\lvert \Pi_{\mathrm{m}} \rvert}{\lvert \Pi \rvert}.
\end{equation}
The consistency $r$ is the length of the mean of the directions, which is $1$ when all moving pairs point the same way and close to $0$ when their directions are uniformly spread:
\begin{equation}
  r = \left\lVert \frac{1}{\lvert \Pi_{\mathrm{m}} \rvert} \sum_{(p, t) \in \Pi_{\mathrm{m}}} \frac{d_t(p)}{\lVert d_t(p) \rVert_2} \right\rVert_2,
\end{equation}
with $r = 0$ when $\Pi_{\mathrm{m}} = \emptyset$.
Each moving pair counts once, whatever its magnitude, so a few large displacements cannot dominate.
The unique rate $\nu$ is the number of distinct cells the patches are matched to, relative to the total number of patches, and averaged over frames:
\begin{equation}
  \nu = \frac{1}{T'-1} \sum_{t=1}^{T'-1} \frac{\bigl\lvert \{ m_t(p) : p \in \Omega \} \bigr\rvert}{\lvert \Omega \rvert}.
\end{equation}
It is $1$ when every patch has its own correspondence and $1 / \lvert \Omega \rvert$ when all collapse onto a single sink cell.

\noindent \textbf{Head score.}
The score of head $(\ell, h)$ at step $k$ in scene $s$ is $S = \mu\, r\, \nu$.
Let $\mathcal{C}^{+} \subseteq \mathcal{C}$ be the calibration scenes whose generated video moves the target correctly, labeled manually, and $\mathcal{W}_{\mathrm{c}}$ the first 15\% of the denoising steps in HunyuanVideo.
The score of a head $\bar{S}(\ell, h)$ averages over steps and then over the scenes:
\begin{equation}
  \bar{S}(\ell, h) = \frac{1}{\lvert \mathcal{C}^{+} \rvert} \sum_{s \in \mathcal{C}^{+}} \frac{1}{\lvert \mathcal{W}_{\mathrm{c}} \rvert} \sum_{k \in \mathcal{W}_{\mathrm{c}}} S^k_s(\ell, h).
\end{equation}
In \cref{fig:correspondence_examples} we report an extended example of how different correspondence distributions influence the components of the scores.
The leftmost column shows samples with a low moving fraction $\mu$, in which cross-frame correspondences have very little or no motion. 
In the bottom case, the motion is more significant, but it covers only a few patches, thereby resulting in a score of $\mu=0.24$.
The second column from the left shows a common case of scattered correspondences, so that the consistency score $r$ is close to $0$, while in the third column most correspondences point to few patches, singling out sink heads, which have low unique rate $\nu$.
The rightmost column instead shows examples of high-quality heads, which have a high score in all three components.

\noindent \textbf{Positional bias.}
To understand if a head has a positional bias, it is necessary to look at its correspondences across several scenes.
In a single scene such a head will likely have a high score $S(\ell, h)$, as its correspondences will coherently point to one direction.
To remove such heads, we devise a metric to measure how much a head agrees in direction across different scenes.
For each scene $s$, we take the mean direction of the moving pairs at each step, weight it by the step's coherence $\mu\, r$, and normalize the sum over steps:
\begin{equation}
\begin{aligned}
  e_s(\ell, h) &= \frac{\mathbf{w}_s(\ell, h)}{\lVert \mathbf{w}_s(\ell, h) \rVert_2},
  \\
  \mathbf{w}_s(\ell, h) &= \sum_{k \in \mathcal{W}_{\mathrm{c}}} \mu^k_s\, r^k_s\, \frac{\mathbf{r}^k_s}{\lVert \mathbf{r}^k_s \rVert_2},
\end{aligned}
\end{equation}
where $\mathbf{r}^k_s$ is the resultant $\sum_{(p, t) \in \Pi_{\mathrm{m}}} d_t(p) / \lVert d_t(p) \rVert_2$ of step $k$.
Intuitively, $\mathbf{r}^k_s$ is the majority vote direction at a scene $s$ and denoising step $k$.
Weighted by the moving fraction $\mu^k_s$ and consistency $r^k_s$, it gives the scene-level direction $e_s$ for a scene $s$.
Therefore, steps where the head does not move the object coherently barely affect its direction.
The bias of the head is the agreement of these directions across scenes:
\begin{equation}
  B(\ell, h) = \left\lVert \frac{1}{\lvert \mathcal{C}^{+} \rvert} \sum_{s \in \mathcal{C}^{+}} e_s(\ell, h) \right\rVert_2 \in [0, 1].
\end{equation}
Target objects in different scenes move in different directions, so a head that follows them has a low $B$, while a head pointing the same way in every scene has $B$ close to $1$.
We discard the heads with $B(\ell, h) \ge 0.9$ and take the 10 remaining heads with the highest $\bar{S}(\ell, h)$ as the motion heads $\mathcal{H}$.
In \cref{fig:positional_bias} we report examples of four heads across the same four scenes, and show how the bias $B$ can reveal heads which have a coherent motion within a scene, but actually encode a positional bias.
The two heads on the left exhibit coherent motion correspondences within each scene, but the directions differ across scenes, so that the bias score is low (0.42, 0.52).
Instead, the two heads on the right point in the same direction across the four scenes: this reveals that they have a high directional bias, and are therefore discarded.

\subsection{Motion probing gates}\label{sec:motion_probing}

The probe estimates the target motion from the motion heads after 15\% of the denoising steps in HunyuanVideo, and the gates decide whether this estimate can be trusted (Sec.~4.3 of the main paper).
Each motion head at each probing step casts a vote for the displacement of every patch of the predicted mask $\mathcal{M}$.
The gates accept the estimate only if three conditions hold.
\emph{Coverage}: the heads agree on the displacement of most patches, whether they move or not, so the estimate is reliable.
\emph{Magnitude}: enough of the object moves by a significant amount, but not further than the scene allows.
\emph{Coherence}: the motion is consistent, \ie the moving patches share one direction, and the early and late probing steps predict the same motion.
If any check fails, the VDM has not committed to a coherent motion of the target in this attempt, and we restart from new noise.

\noindent \textbf{Votes and consensus.}
The probe runs the first $\lceil 0.15 J \rceil$ denoising steps without intervention, \ie 8 of the $J = 50$ steps on HunyuanVideo.
We discard the first half, where the correspondences are still noisy, and keep the last 4 steps.
Each of these steps and each motion head in $\mathcal{H}$ gives a vote, \ie a displacement of each patch $p \in \Omega$ in each latent frame $t$, computed as in Eq.~(2) of the main paper with $\Omega$ the patches of $\mathcal{M}$, for 40 votes in total.
For each patch and frame, the consensus $\tilde{d}_t(p)$ is the per-coordinate median of the votes, and the spread $\rho_t(p)$ is the median distance of the votes from it.
The heads \emph{agree} on $(p, t)$ when $\rho_t(p) \le 1.5$ cells.
At the last latent frame, $\Omega_{\mathrm{a}} \subseteq \Omega$ are the agreed patches, and $\Omega_{\mathrm{m}} \subseteq \Omega_{\mathrm{a}}$ those that also move by at least one cell.

\noindent \textbf{Velocity.}
In each latent frame, $\bar{d}_t$ is the per-coordinate median of $\tilde{d}_t(p)$ over the patches of $\Omega_{\mathrm{m}}$ agreed in that frame, \ie the moving patches of Sec.~4.3 of the main paper.
The velocity $\mathbf{v}$ is fitted as in Eq.~(3) of the main paper, over the frames where such patches exist, and converted from cells to pixels of \image, with 16 pixels per cell on HunyuanVideo.

\noindent \textbf{Coverage gate.}
The coverage $g_{\mathrm{cov}}$ is the share of pairs $(p, t) \in \Pi$ on which the heads agree, with $\Pi = \Omega \times \{1, \dots, T'-1\}$ as in \cref{sec:motion_calib}, and must reach $0.5$.
A patch on which heads agree for no motion is considered reliable.

\noindent \textbf{Magnitude gate.}
The magnitude is the 80th percentile of the last-frame displacement of the agreed patches, and must reach at least one cell.
Formally:
\begin{equation}
  g_{\mathrm{mag}} = P_{80}\bigl\{ \lVert \tilde{d}_{T'-1}(p) \rVert_2 : p \in \Omega_{\mathrm{a}} \bigr\} \;\ge\; 1,
\end{equation}
with $g_{\mathrm{mag}} = 0$ when $\Omega_{\mathrm{a}} = \emptyset$.
It thus requires about a fifth of the agreed patches to move, which lets the gate pass when only part of the mask moves, \eg the door of a cabinet whose frame is also in $\mathcal{M}$.
The travel check bounds the total motion over the video, $\lVert \mathbf{v} \rVert_2 (T - 1)$ in pixels, by half the image diagonal $\sqrt{H^2 + W^2}$, which rejects implausible velocities.

\noindent \textbf{Coherence gate.}
Across the object, the direction coherence is the mean resultant length of the last-frame directions of the moving patches, and must reach $0.5$:
\begin{equation}
  g_{\mathrm{dir}} = \left\lVert \frac{1}{\lvert \Omega_{\mathrm{m}} \rvert} \sum_{p \in \Omega_{\mathrm{m}}} \frac{\tilde{d}_{T'-1}(p)}{\lVert \tilde{d}_{T'-1}(p) \rVert_2} \right\rVert_2 \;\ge\; 0.5,
\end{equation}
with $g_{\mathrm{dir}} = 0$ when fewer than three patches move.
Across steps, we recompute the consensus separately on the votes of the first two and of the last two probing steps, and require their last-frame displacements to differ by at most $1.5$ cells, in median over the patches agreed in both.

An attempt passes when all three gates pass.
If none of the $R$ attempts passes, we generate from $\varepsilon_0$ with the velocity $\mathbf{v}_0$ estimated with every gate disabled.
As in the main paper, we denote by $\mathbf{v}^*$ the velocity that expands the free mask of the latent lock.
It is the velocity $\mathbf{v}$ of the first attempt that passes all the gates, or $\mathbf{v}_0$ if none does.

\noindent \textbf{Gates on point tracks (Dream2Flow$^\dagger$).}
For the Dream2Flow baseline of Tab.~2 of the main paper, we apply the gates to the generated videos instead of the attention.
Dream2Flow generates one video from each of the seeds 42 and 43, and CoTracker provides tracks of its object, expressed in 16-pixel cells, which replace the consensus displacements of the motion heads.
A single tracker on a complete video has no vote agreement and no probing steps, so the coverage gate is dropped and the coherence gate keeps only the direction coherence, without the cross-step check.
The magnitude gate is kept as it is, with both its magnitude and travel checks, and all checks keep their thresholds.
For each check, the margin is the distance of its value from the threshold, relative to the threshold, and is positive when the check passes and negative when it fails.
The score of a video is the margin of its weakest check, so it is positive only when both remaining gates pass.
We keep the video with the higher score, which prefers a video passing every gate to one failing any, and otherwise the one that fails least.
When the two scores tie, \eg when the object moves in neither video, we keep the one generated with seed 42.

\subsection{Threshold calibration}\label{sec:thresholds}
The thresholds were set on the calibration set $\mathcal{C}$, which is disjoint from FunFlow, without new generations.
We replayed the probe offline on the correspondences saved during head calibration, and compared the gate decisions with the manual check of whether each generated video moves the target correctly.
We started from round values, \ie a spread of $1.5$ cells, which tolerates the one-cell jitter of the argmax, and $0.5$ for the shares and resultant lengths, and kept the values for which the gates pass on most videos that move the target correctly and reject most of the others.

\section{Evaluation}\label{sec:eval}
In this section we detail the evaluation procedure, common to all methods, starting with how to obtain the predicted (\cref{sec:flow_eval_pred}) and ground-truth (\cref{sec:flow_eval_gt}) object flow.
In~\cref{sec:flow_eval_metrics} we introduce the metrics, and in~\cref{sec:bootstrap} we detail the process used to assess the statistical significance of our results.

\subsection{Predicted flow}
\label{sec:flow_eval_pred}

\noindent \textbf{Object and complement particles.} What counts as the object is always decided by the ground-truth mask $\mathcal{M}_{\mathrm{GT}}$ of the moving part in \image, for every method and every evaluation mask.
A prediction consists of $N^{\mathrm{o}}$ object particles sampled on a regular pixel grid inside $\mathcal{M}_{\mathrm{GT}}$ in the first frame, at pixels $\mathbf{u}_n \in \mathbb{R}^2$, $n = 1, \dots, N^{\mathrm{o}}$.
Each particle is tracked in 2D through the $T$ frames of the generated video and lifted to world coordinates with the per-frame depth, giving positions $\hat{\mathbf{p}}_{n,t} \in \mathbb{R}^3$ for $t = 0, \dots, T-1$.
The tracker also gives a validity mask $v^{\mathrm{o}}_{n,t} \in \{0, 1\}$, which is $1$ when the 2D tracker considers the particle visible and the depth at its tracked pixel is defined.
To test whether the rest of the scene stays still, we also track $N^{\mathrm{c}}$ complement particles $\mathbf{c}_{j,t}$, $j = 1, \dots, N^{\mathrm{c}}$, with validity $v^{\mathrm{c}}_{j,t}$, in the same video.
They lie on the same pixel grid outside $\mathcal{M}_{\mathrm{GT}}$, and only on pixels with valid depth.
Both sets cover their regions uniformly, so we convert particle counts to image areas.
Let $\Omega^{\mathrm{o}}$ and $\Omega^{\mathrm{c}}$ be the object and complement regions from which the particles are sampled.
Each object particle stands for $\omega^{\mathrm{o}} = \lvert \Omega^{\mathrm{o}} \rvert / N^{\mathrm{o}}$ pixels and each complement particle for $\omega^{\mathrm{c}} = \lvert \Omega^{\mathrm{c}} \rvert / N^{\mathrm{c}}$.

\subsection{Ground-truth flow}
\label{sec:flow_eval_gt}

\noindent \textbf{Initial points and filtering.}
Since every object particle lies inside $\mathcal{M}_{\mathrm{GT}}$, we apply the ground-truth motion to the tracked object particles themselves.
This gives a one-to-one correspondence between predicted and ground-truth points.
Each particle is lifted with the sensor depth \depth of the image \image, the intrinsics $\mathbf{K}$, and the camera-to-world pose $\mathbf{E}$:
\begin{equation}
  \mathbf{x}_n = \pi^{-1}\!\left(\mathbf{u}_n, \mathcal{D}(\mathbf{u}_n); \mathbf{K}, \mathbf{E}\right),
\end{equation}
where $\pi^{-1}$ denotes pinhole back-projection.

We provide the first-frame depth \depth, but Dream2Flow's pipeline estimates the depth of all the frames in the video, including the first frame, so the predicted first-frame depth may differ from \depth.
A particle is kept only if its starting point is reliable both in the ground-truth depth \depth and in the prediction.
Let $\hat{z}_n$ be the camera-frame depth of the predicted first-frame position $\hat{\mathbf{p}}_{n,0}$.
The kept set is
\begin{equation}
\begin{aligned}
  \mathcal{K} = \biggl\{ n \;:\;& \mathcal{D}(\mathbf{u}_n) > 0,\;\; v^{\mathrm{o}}_{n,0} = 1,
  \\
  & \frac{\lvert \mathcal{D}(\mathbf{u}_n) - \hat{z}_n \rvert}{\hat{z}_n} \le 0.25 \biggr\}.
\end{aligned}
\end{equation}
The last condition removes particles whose sensor and predicted depths disagree, for example when the sensor measures through a glass surface.
All object terms below use only particles in $\mathcal{K}$, and the object area is $A^{\mathrm{o}} = \omega^{\mathrm{o}} \lvert \mathcal{K} \rvert$.

SceneFun3D's annotations provide only the motion axis, so we expand it into a one-parameter family of ground-truth configurations.
Let $M(\mathbf{x}; \alpha)$ apply the annotated motion with amount $\alpha \ge 0$, in the opening direction, to a point $\mathbf{x}$.
For a rotation with unit axis $\mathbf{a}$ through the origin $\mathbf{o}$, $M(\mathbf{x}; \alpha) = \mathbf{o} + \mathbf{R}_{\mathbf{a}}(\alpha)(\mathbf{x} - \mathbf{o})$, where $\mathbf{R}_{\mathbf{a}}(\alpha)$ is the rotation by angle $\alpha$ about $\mathbf{a}$.
For a translation along the unit direction $\mathbf{a}$, $M(\mathbf{x}; \alpha) = \mathbf{x} + \alpha\,\mathbf{a}$.
We sample the amount densely,
\begin{equation}
  \mathcal{A} = \{ 0,\, \Delta \alpha,\, 2\Delta \alpha,\, \dots,\, \alpha_{\max} \},
\end{equation}
with $\Delta \alpha = 1\,\mathrm{mm}$ and $\alpha_{\max} = 1\,\mathrm{m}$ for translations, and $\Delta \alpha = 0.25^\circ$ and $\alpha_{\max} = 90^\circ$ for rotations.
Each $\alpha \in \mathcal{A}$ defines a candidate ground-truth configuration $\{ M(\mathbf{x}_n; \alpha) \}_{n \in \mathcal{K}}$ of the target object.
All candidates start from the same points $\mathbf{x}_n$ and differ only in how far the part has moved along the annotated axis.

\noindent \textbf{Frame matching.}
In frame $t$, we score the valid kept particles $\mathcal{S}_t = \{ n \in \mathcal{K} : v^{\mathrm{o}}_{n,t} = 1 \}$.
A frame is scorable when at least $10\%$ of the kept particles are valid in it:
\begin{equation}
  \mathcal{T}^{\mathrm{o}} = \left\{ t \;:\; \lvert \mathcal{S}_t \rvert \ge \lceil 0.1 \lvert \mathcal{K} \rvert \rceil \right\}.
\end{equation}

Each scorable frame $t \in \mathcal{T}^{\mathrm{o}}$ is matched to the candidate configuration that best explains it, \ie the one with the lowest mean end-point error over its valid particles:
\begin{equation}
\begin{aligned}
  \alpha_t &= \arg\min_{\alpha \in \mathcal{A}} \frac{1}{\lvert \mathcal{S}_t \rvert} \sum_{n \in \mathcal{S}_t} \left\lVert \hat{\mathbf{p}}_{n,t} - M(\mathbf{x}_n; \alpha) \right\rVert_2,
  \\
  \mathbf{g}_{n,t} &= M(\mathbf{x}_n; \alpha_t).
\end{aligned}
\end{equation}
The ground-truth motion does not specify when or how fast the object opens, so each predicted frame is matched independently to the amount $\alpha_t$ that best fits it.
If the predicted particles move along the ground-truth axis, some candidate configuration places the ground-truth positions close to them, while if they move in any other direction, all candidate configurations stay far from them.
If the predicted object does not move, or moves in the closing direction, the best-fitting amount is $\alpha_t \approx 0$, \ie the ground-truth positions $\mathbf{g}_{n,t}$ are close to the starting points $\mathbf{x}_n$.
To compute the metrics, we compare each predicted particle with its own ground-truth counterpart, $e_{n,t} = \lVert \hat{\mathbf{p}}_{n,t} - \mathbf{g}_{n,t} \rVert_2$.

\subsection{Metrics}
\label{sec:flow_eval_metrics}

\noindent \textbf{Motion tests.}
Depth jitter across frames can cause a still particle to jitter in 3D.
We therefore treat displacements up to $\delta^{\mathrm{m}} = 3\,\mathrm{cm}$ from the first-frame position as noise, and require a particle to exceed $\delta^{\mathrm{m}}$ in at least $\kappa = 2$ frames, which ignores single-frame tracker or depth spikes.
Both tests below count only the frames $t$ in which the particle is valid both in $t$ and in the first frame, its reference; for object particles the latter always holds, as they belong to $\mathcal{K}$.
Formally, let $\gamma^{\mathrm{o}}_{n,t} = \lVert \hat{\mathbf{p}}_{n,t} - \hat{\mathbf{p}}_{n,0} \rVert_2$ be the displacement of object particle $n$ from its first-frame position. The particle moves if:
\begin{equation}
  \eta^{\mathrm{o}}_{n} = \mathbf{1}\!\left[ \,\Bigl\lvert \{ t \in \mathcal{T}^{\mathrm{o}} : \gamma^{\mathrm{o}}_{n,t} > \delta^{\mathrm{m}} \} \Bigr\rvert \ge \kappa \right].
\end{equation}

The scorable frames of the complement particles, $\mathcal{T}^{\mathrm{c}}$, are the frames after the first in which at least $10\%$ of the complement particles are valid both there and in the first frame.
Let $\gamma^{\mathrm{c}}_{j,t} = \lVert \mathbf{c}_{j,t} - \mathbf{c}_{j,0} \rVert_2$ be the displacement of complement particle $j$ from the first frame; the particle moves if
\begin{equation}
  \eta^{\mathrm{c}}_{j} = \mathbf{1}\!\left[ \,\Bigl\lvert \{ t \in \mathcal{T}^{\mathrm{c}} : \gamma^{\mathrm{c}}_{j,t} > \delta^{\mathrm{m}} \} \Bigr\rvert \ge \kappa \right].
\end{equation}
The moving background area defined by $\eta^{\mathrm{c}}_{j}$ is also used as the false-positive area, $\mathrm{FP} = \omega^{\mathrm{c}} \sum_j \eta^{\mathrm{c}}_{j}$.
Invalid frames never count as moved, so background revealed or hidden by the moving part does not count as motion.

\noindent \textbf{Recall.}
In frame $t \in \mathcal{T}^{\mathrm{o}}$, the accuracy is the fraction of valid particles that move and lie within $\delta^{\mathrm{e}}$ of their ground truth, averaged over the thresholds $\mathcal{E} = \{2, 5, 10, 15, 20\}\,\mathrm{cm}$:
\begin{equation}
  a_t = \frac{1}{\lvert \mathcal{E} \rvert} \sum_{\delta^{\mathrm{e}} \in \mathcal{E}} \frac{1}{\lvert \mathcal{S}_t \rvert} \sum_{n \in \mathcal{S}_t} \mathbf{1}\!\left[ \eta^{\mathrm{o}}_{n} = 1 \;\land\; e_{n,t} < \delta^{\mathrm{e}} \right].
\end{equation}
A single threshold would be all-or-nothing, while averaging over $2$--$20\,\mathrm{cm}$ gives partial success.
Lost particles are left out of the denominator, so tracking loss itself is not penalized.
The recall averages the accuracy over the frames $\mathcal{T}^{\mathrm{m}}$ in which the matched ground truth is over the noise level:
\begin{equation}
\begin{aligned}
  \mathcal{T}^{\mathrm{m}} &= \left\{ t \in \mathcal{T}^{\mathrm{o}} : \frac{1}{\lvert \mathcal{S}_t \rvert} \sum_{n \in \mathcal{S}_t} \lVert \mathbf{g}_{n,t} - \mathbf{x}_n \rVert_2 > \delta^{\mathrm{m}} \right\},
  \\
  \mathrm{R} &= \begin{cases} \dfrac{1}{\lvert \mathcal{T}^{\mathrm{m}} \rvert} \sum_{t \in \mathcal{T}^{\mathrm{m}}} a_t & \mathcal{T}^{\mathrm{m}} \neq \emptyset, \\ 0 & \text{otherwise.} \end{cases}
\end{aligned}
\end{equation}

Note that the motion test $\eta^{\mathrm{o}}_{n}$ is essential to the definition of a recall success: without it, particles that stay still or only jitter would count as correct whenever they lie close to their ground truth, \eg~near the hinge of a door, or when only part of the object moves.

\noindent \textbf{Precision, F1 and SBR.}
The true-positive area is the object area that moves correctly, $\mathrm{TP} = \mathrm{R}\, A^{\mathrm{o}}$, and the false-positive area is the moving background $\mathrm{FP}$.
The precision is the share of the correctly moving object area among all the area counted as moving, and the F1 is the harmonic mean of precision and recall:
\begin{equation}
  \mathrm{Pr} = \frac{\mathrm{TP}}{\mathrm{TP} + \mathrm{FP}},
  \qquad
  \mathrm{F1} = \frac{2\, \mathrm{Pr}\, \mathrm{R}}{\mathrm{Pr} + \mathrm{R}},
\end{equation}
with $\mathrm{Pr} = 0$ when $\mathrm{TP} + \mathrm{FP} = 0$ and $\mathrm{F1} = 0$ when $\mathrm{Pr} + \mathrm{R} = 0$.
The SBR scores whether the background stays still, independently of how the target object moves:
\begin{equation}
  \mathrm{SBR} = \frac{A^{\mathrm{o}}}{A^{\mathrm{o}} + \mathrm{FP}} \in (0, 1].
\end{equation}
The SBR can also be considered as the precision of an object flow whose object moves perfectly ($\mathrm{TP} = A^{\mathrm{o}}$) with the same moving background.
Unlike the precision, the SBR does not drop when the object fails to move, so recall and SBR separate whether the object moves correctly from whether the rest stays still.
The SBR is defined based on the size of the target object, which penalizes movements of distractor objects which are large compared to the target one.

\noindent \textbf{Failure cases.}
A method may produce no flow for a sample, for example because the mask is empty, or the tracker loses the object in every frame.
The first case is common with Dream2Flow's mask extraction method on \ourdataset, which produces many empty masks, while the second case may arise in the case of hallucinations that change the scene from one frame to the next.
This can also happen when the target object is a small part made of a reflective or transparent material (\eg~an oven door).
In these cases, the tracker can find no valid points due to invalid depth.
We count such a sample as a failure rather than dropping it, and set all four metrics, including the SBR, to $0$.

\noindent \textbf{Evaluation regions and aggregation.}
By default the metrics consider the whole image: object particles are sampled in $\mathcal{M}_{\mathrm{GT}}$ and complement particles everywhere else.
To evaluate the flow restricted to a region, we drop the object and complement particles whose first-frame pixel lies outside an evaluation mask before scoring.

\subsection{Statistical significance}
\label{sec:bootstrap}

FunFlow has a finite number of samples, so every value is an estimate that depends on the choice of samples.
We quantify this uncertainty with a non-parametric bootstrap over samples~\cite{efron1979bootstrap}.
It gives a 95\% confidence interval for each result and tests whether the gain of \ourmethod{} over Dream2Flow would hold on a different set of samples.
We draw 10\,000 bootstrap resamples, each made of 178 samples drawn uniformly with replacement, and recompute the mean on each of them.
The 95\% confidence interval is given by the 2.5th and 97.5th percentiles of these means.
The resamples are drawn once, with a fixed random seed, and shared by all methods, metrics and evaluation regions, so each bootstrap resample compares all methods on the same samples.
To test a gain, we bootstrap the per-sample difference between \ourmethod{} and Dream2Flow.
The Improvements rows of \cref{tab:funflow_main_ci} report the mean difference and its interval, computed as above.
We call a gain significant when its interval excludes $0$, which corresponds to a two-sided test at the 5\% level.

\section{Additional results}\label{sec:quant}
In this section we report additional results, starting with a split of our main results across difficulty, movement type, and probe outcome in~\cref{sec:result_split}.
In~\cref{sec:cosmos}, we show that our method can generalize to another VDM, and in~\cref{sec:ablations} we report additional ablations in which we use information from the ground-truth motion as part of the motion probing mechanism.
\cref{sec:mask_extraction_results} reports the quantitative performance of the mask extraction procedure, while in~\cref{sec:confidence} we report the bootstrap confidence intervals of our results in the main paper.

\subsection{Result splits}\label{sec:result_split}

\begin{figure*}[t]
    \centering
    \begin{overpic}[width=0.975\textwidth]{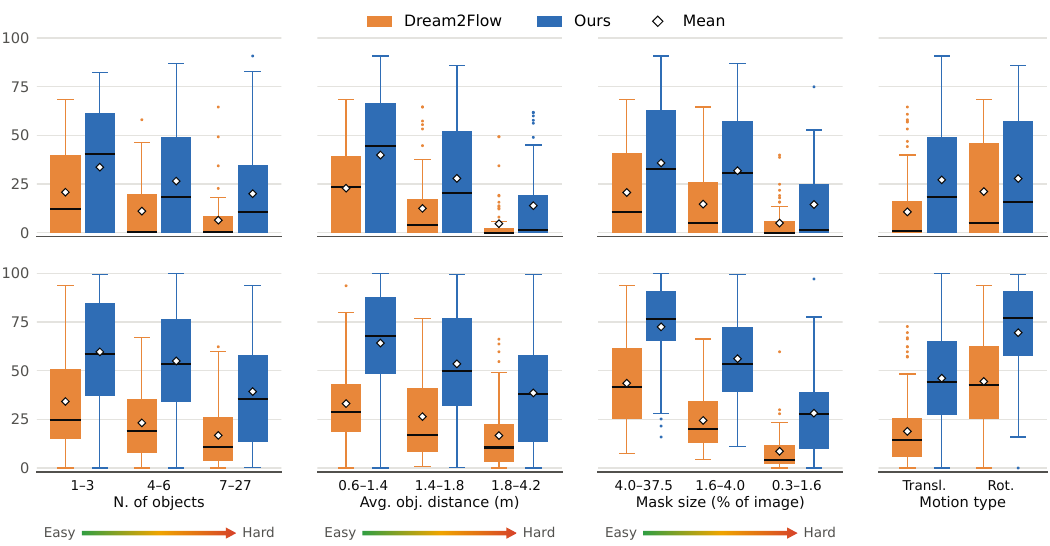}
        \put(-0.6,39.1){\rotatebox{90}{\makebox[0pt]{\footnotesize F1 score}}}
        \put(-0.6,16.7){\rotatebox{90}{\makebox[0pt]{\footnotesize Still Background Rate}}}
    \end{overpic}
    \vspace{-3mm}
    \caption{
    F1 score (top) and SBR (bottom) of Dream2Flow (orange) and \ourmethod{} (blue) with HunyuanVideo, on the whole image.
    Samples are split by, from left to right: the number of objects detected, the camera-to-object distance, the size of the ground-truth object mask, and the motion type of the object. 
    Boxes span the 25th--75th percentile of the samples; line: median; $\diamond$: mean; whiskers: 1.5$\times$ IQR; dots: outliers.
    }
    \label{fig:hardness_split}
\end{figure*}

\begin{table*}[t]
    \centering
    \caption{F1 score and Still Background Rate (SBR) of Dream2Flow (D2F) and \ourmethod{} (Ours) with HunyuanVideo, on the whole image. Samples are split by the number of objects detected, the camera-to-object distance, the size of the ground-truth object mask and the motion type of the object, as in Fig.~\ref{fig:hardness_split}. 
    We report the mean and median over the samples of each group; $\Delta$ is Ours $-$ D2F; Samples (\%) is the share of samples in each group.}
    \label{tab:hardness_split}
    \vspace{-3mm}
    \tabcolsep 4pt
    \definecolor{hardeasy}{HTML}{2F9E44}\definecolor{hardmid}{HTML}{F0A500}\definecolor{hardhard}{HTML}{D6402B}
    \newcommand{\hardarrow}{\makebox[1.6mm]{\tikz[overlay, x=1mm, y=\normalbaselineskip]
        \shade[top color=hardeasy, middle color=hardmid, bottom color=hardhard]
        (-0.3,0.75) -- (0.3,0.75) -- (0.3,-1.9) -- (0.8,-1.9) -- (0,-2.25) -- (-0.8,-1.9) -- (-0.3,-1.9) -- cycle;}}
    \resizebox{0.9\textwidth}{!}{%
    \begin{tabular}{r@{\hspace{3pt}}c@{\hspace{3pt}}lc ccc c ccc c ccc c ccc}
    \toprule
    & & & & \multicolumn{7}{c}{F1 $\uparrow$} & & \multicolumn{7}{c}{SBR $\uparrow$} \\
    \cmidrule{5-11} \cmidrule{13-19}
    & & & & \multicolumn{3}{c}{Mean} & & \multicolumn{3}{c}{Median} & & \multicolumn{3}{c}{Mean} & & \multicolumn{3}{c}{Median} \\
    \cmidrule{5-7} \cmidrule{9-11} \cmidrule{13-15} \cmidrule{17-19}
    \color{gray} \# & & Group & Samples (\%) & D2F & Ours & $\Delta$ & & D2F & Ours & $\Delta$ & & D2F & Ours & $\Delta$ & & D2F & Ours & $\Delta$ \\
    \midrule
    & & \multicolumn{17}{l}{\textit{N. of objects detected}} \\
    \color{gray} 1 & \hardarrow & \quad 1--3 & 37.6 & 20.8 & 33.7 & \color{forestgreen}{+12.9} & & 12.1 & 40.4 & \color{forestgreen}{+28.3} & & 34.2 & 59.7 & \color{forestgreen}{+25.5} & & 24.6 & 58.8 & \color{forestgreen}{+34.2} \\
    \color{gray} 2 & & \quad 4--6 & 33.1 & 11.1 & 26.5 & \color{forestgreen}{+15.4} & & 0.2 & 18.6 & \color{forestgreen}{+18.4} & & 23.2 & 55.0 & \color{forestgreen}{+31.8} & & 18.8 & 53.2 & \color{forestgreen}{+34.4} \\
    \color{gray} 3 & & \quad 7--27 & 29.2 & 6.4 & 20.0 & \color{forestgreen}{+13.6} & & 0.4 & 10.7 & \color{forestgreen}{+10.3} & & 16.8 & 39.3 & \color{forestgreen}{+22.5} & & 10.6 & 35.4 & \color{forestgreen}{+24.8} \\
    \specialrule{0.1pt}{\aboverulesep}{\belowrulesep}
    & & \multicolumn{17}{l}{\textit{Avg. obj. distance (m)}} \\
    \color{gray} 4 & \hardarrow & \quad 0.6--1.4 & 33.7 & 22.9 & 39.9 & \color{forestgreen}{+17.0} & & 23.7 & 44.5 & \color{forestgreen}{+20.8} & & 33.1 & 64.2 & \color{forestgreen}{+31.1} & & 28.9 & 67.7 & \color{forestgreen}{+38.8} \\
    \color{gray} 5 & & \quad 1.4--1.8 & 33.1 & 12.5 & 27.9 & \color{forestgreen}{+15.4} & & 4.0 & 20.3 & \color{forestgreen}{+16.3} & & 26.4 & 53.5 & \color{forestgreen}{+27.1} & & 17.2 & 49.9 & \color{forestgreen}{+32.7} \\
    \color{gray} 6 & & \quad 1.8--4.2 & 33.1 & 4.6 & 13.9 & \color{forestgreen}{+9.3} & & 0.0 & 1.5 & \color{forestgreen}{+1.5} & & 16.7 & 38.6 & \color{forestgreen}{+21.9} & & 10.5 & 38.2 & \color{forestgreen}{+27.7} \\
    \specialrule{0.1pt}{\aboverulesep}{\belowrulesep}
    & & \multicolumn{17}{l}{\textit{Mask size (\% of image)}} \\
    \color{gray} 7 & \hardarrow & \quad 4.0--37.5 & 33.1 & 20.6 & 35.8 & \color{forestgreen}{+15.2} & & 10.7 & 32.5 & \color{forestgreen}{+21.8} & & 43.6 & 72.6 & \color{forestgreen}{+29.0} & & 41.6 & 76.8 & \color{forestgreen}{+35.2} \\
    \color{gray} 8 & & \quad 1.6--4.0 & 33.1 & 14.7 & 31.8 & \color{forestgreen}{+17.1} & & 5.0 & 30.8 & \color{forestgreen}{+25.8} & & 24.5 & 56.2 & \color{forestgreen}{+31.7} & & 20.1 & 53.6 & \color{forestgreen}{+33.5} \\
    \color{gray} 9 & & \quad 0.3--1.6 & 33.7 & 4.9 & 14.5 & \color{forestgreen}{+9.6} & & 0.0 & 1.4 & \color{forestgreen}{+1.4} & & 8.6 & 28.2 & \color{forestgreen}{+19.6} & & 4.1 & 27.6 & \color{forestgreen}{+23.5} \\
    \specialrule{0.1pt}{\aboverulesep}{\belowrulesep}
    & & \multicolumn{17}{l}{\textit{Motion type}} \\
    \color{gray} 10 & & \quad Translation & 74.2 & 10.7 & 27.1 & \color{forestgreen}{+16.4} & & 0.9 & 18.4 & \color{forestgreen}{+17.5} & & 18.8 & 46.1 & \color{forestgreen}{+27.3} & & 14.5 & 44.3 & \color{forestgreen}{+29.8} \\
    \color{gray} 11 & & \quad Rotation & 25.8 & 21.1 & 27.8 & \color{forestgreen}{+6.7} & & 5.2 & 15.6 & \color{forestgreen}{+10.4} & & 44.5 & 69.6 & \color{forestgreen}{+25.1} & & 42.7 & 77.3 & \color{forestgreen}{+34.6} \\
    \bottomrule
    \end{tabular}
    }
\end{table*}

In \cref{fig:hardness_split} we report the results on \ourdataset of \ourmethod and Dream2Flow on the whole image, with results split according to difficulty or movement type.
For difficulty we consider three axes: number of objects detected by our mask extraction pipeline (first column), average object distance, obtained by cropping the depth \depth with the ground-truth mask (second column) and mask size as a percentage of the area of \image (third column).
The rightmost column shows the results split according to the movement type: translation (\eg drawers), or rotation (\eg doors and windows).
For each method and split we report F1 and Still Background Rate (SBR) scores.
In \cref{tab:hardness_split} we report the values of mean and median for each column in \cref{fig:hardness_split}, and our increments.

Dream2Flow performance gets worse in both F1 and SBR as the difficulty of the sample increases (\eg~with the number of objects detected, F1 falls from 20.8 in the easy cases to 6.4 in the hard ones).
\ourmethod trends are the same, as our method struggles with objects that are small or far away (\eg~with the object distance, our method scores an F1 of 39.9 on close objects, and of 13.9 on far-away objects), but we consistently beat Dream2Flow on all groups.

In all settings the mean and median SBR increment given by \ourmethod are quite similar, showing that the increment is large and consistent. 
On the contrary, the F1 score increment is less stable.
The most difficult samples for our method (and Dream2Flow as well) are the ones in which the objects are quite far from the camera and/or appear very small (rows 6 and 9). 
In row 6, our method has a mean F1 increment of 9.3 points, and a median of 1.5.
Similarly, in row 9 the mean and median are respectively 9.6 and 1.4.
These samples are particularly difficult as they greatly lower the number of patches that represent the object of interest.
This makes it difficult for the VDM to perceive the object, and gives our method fewer votes when probing the motion $\mathbf{v}^*$.
The F1 increment provided by \ourmethod when the number of objects detected is high is quite large: row 3 shows a mean increment of 13.6, and a median of 10.3.

Finally, in rows 10--11 the results are split by motion type (rotation and translation).
Our method improves SBR by a similar amount in both rotation and translation samples (by 25.1 and 27.3 in mean, 34.6 and 29.8 in median).
On the other hand, the F1 improvement is larger in translation than in rotation, scoring respectively +16.4 and +6.7 in mean.
Our motion probing mechanism can easily model translation, which is also easier to approximate in the 2D space, but still the increment on rotation samples, which cover one fourth of the dataset, is large.

\begin{figure}[t]
    \centering
    \includegraphics[width=0.8\columnwidth]{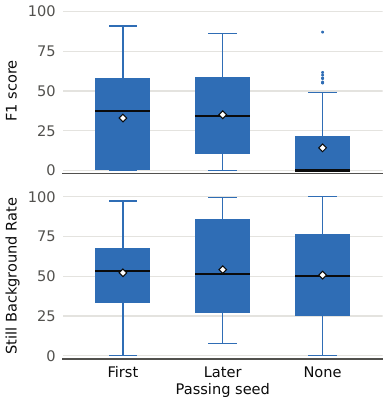}
    
    \vspace{-3mm}
    \caption{
    F1 score (top) and Still Background Rate (bottom) of \ourmethod{} with HunyuanVideo on FunFlow, over the whole image, with samples split by the first probe that passed the motion gates.
    Boxes span the 25th--75th percentile of the samples; line: median; $\diamond$: mean; whiskers: 1.5$\times$ IQR; dots: outliers.
    }
    \label{fig:probe_split}
\end{figure}

In \cref{fig:probe_split}, we report the results of \ourmethod using HunyuanVideo on \ourdataset, with whole image evaluation.
We split the samples according to the first probe that passed the motion gates: the first (leftmost column, 40\% of samples), one between the second and the fifth (central column, 26\% of samples), or none (last column, 33\% of samples).
In the top plot, we observe how the F1 score varies across probe outcomes.
The median F1 score stays high when a probe passes (37.2 and 34.5 for first and later probe, respectively), and drops to 0 in the samples where no probe passed.
The samples that never pass are particularly hard: 46\% of the samples that never passed show a still video, against 15\% in the other two groups.
On the other hand, the SBR is rather stable: the first, second and third group have a median of 53.1, 51.6 and 49.9 respectively.
This is consistent with the behavior of the latent lock, which forces a frozen scene outside the target object area.

\subsection{VDM generalization}\label{sec:cosmos}
In \cref{tab:cosmos}, we show that \ourmethod can generalize to another attention-based VDM, Cosmos3-Nano (Cosmos).
We follow the same procedure as for HunyuanVideo, \ie we calibrate the motion gate thresholds and probing parameters on the calibration set after manual selection and motion head classification.
Cosmos generally performs worse than HunyuanVideo.
We found that most of its errors are due to (1) the target object not moving, and (2) strong hallucinations that change the scenes.
Applied to Cosmos, \ourmethod significantly improves F1 score, Precision, and SBR in the whole image by 7.4, 10.5, and 31.8 points respectively.
Within our predicted masks increments are smaller, and not significant according to our bootstrap evaluation.

During the calibration phase, we observed that in Cosmos the correspondences in the latent space occur later in the denoising process, so that with probe 50\% of the denoising steps instead of 15\% as in HunyuanVideo.
Cosmos performs on average 2.26 attempts, against 2.9 for HunyuanVideo, but due to the higher number of probing steps, the average generation cost is higher than with HunyuanVideo (2.16 vs 1.46, where 1 is a complete generation).
We attribute this to the uncommon architecture of Cosmos: instead of using classic Diffusion Transformer (DiT) layers as HunyuanVideo, each of Cosmos' layers has one set of weights for the video tokens and a separate set for the language tokens.
Therefore, this result shows that the improvements provided by \ourmethod are not limited to DiT-based architectures.

\begin{table}[t]
\caption{Comparison between Dream2Flow and \ourmethod{} with Cosmos3-Nano as the video generation model on FunFlow, over the whole image and inside our mask.
Our results are highlighted.
Differences are in \textcolor{forestgreen}{green} (\textcolor{red}{red}) when significantly positive (negative), and in \textcolor{gray}{gray} otherwise.}
\label{tab:cosmos}
\vspace{-3mm}
\tabcolsep 5pt
\resizebox{\linewidth}{!}{%
\begin{tabular}{rl cccc ccc}
\toprule
 & & \multicolumn{4}{c}{Whole image} & \multicolumn{3}{c}{Our mask} \\
\cmidrule(lr){3-6} \cmidrule(lr){7-9}
\color{gray} \# & Method & F1 $\uparrow$ & R $\uparrow$ & Pr $\uparrow$ & SBR $\uparrow$ & F1 $\uparrow$ & R $\uparrow$ & Pr $\uparrow$ \\
\midrule
\color{gray} 1 & Dream2Flow~\cite{dharmarajan2025dream2flow} & 9.6 & 16.5 & 8.6 & 39.1 & 18.4 & 15.2 & 27.2 \\
\rowcolor{myazure}
\color{gray} 2 & \ourmethod{} & \textbf{16.9} & \textbf{17.3} & \textbf{19.1} & \textbf{70.8} & \textbf{21.0} & \textbf{17.2} & \textbf{31.5} \\
\midrule
\multicolumn{9}{l}{\textit{Improvements}} \\
\color{gray} 3 & $\Delta$ (2 $-$ 1) & \color{forestgreen}{+7.4} & \color{gray}{+0.8} & \color{forestgreen}{+10.5} & \color{forestgreen}{+31.8} & \color{gray}{+2.7} & \color{gray}{+1.9} & \color{gray}{+4.3} \\
\bottomrule
\end{tabular}
}
\end{table}

\subsection{Additional ablations}\label{sec:ablations}

In \cref{tab:ablation_motion} we show an additional ablation on \ourmethod, in which we use a motion derived from the ground-truth as $\mathbf{v}^*$, instead of estimating it via motion probing.
In this ablation we use our predicted mask for steering, and we still perform the motion probing to perform MASS (motion-aware seed selection), but the resulting $\mathbf{v}^*$ is replaced by the projection of the ground-truth motion (linearized for objects with rotational motion).
In row 2, we replace the probed motion with the one derived from the ground-truth, setting a translation bound of 1\,m and a rotation bound of $90^\circ$.
In row 3, we perform the same experiment but set the translation bound to 0.5\,m, while the rotation one is unchanged.
In both cases, the only significant difference is a loss in SBR.
We speculate that such values of ground-truth velocity could expand the free mask before the video diffusion model initiates the motion of the target object.
This creates a free area in which the model can hallucinate, thereby causing a drop in SBR.
To counter this, in row 4 we still use the ground-truth motion, but only to specify the direction of the latent lock, while the intensity of the probed motion is used to regulate the expansion of the free mask.
This result has no significant difference from our standard setting, and shows the importance of expanding the free mask with a velocity probed by the model.

\begin{table}[t]
\caption{Ablations on the motion probing mechanism with HunyuanVideo, on the whole image and inside our mask.
The latent lock is expanded along the motion $\mathbf{v}^*$ probed from the heads (probe), along the ground-truth 2D motion of the part (GT), along the ground-truth motion at half the translation range (GT$_{0.5}$), or along the direction of GT$_{0.5}$ only (GT dir.).
Our standard setting is highlighted.
Differences are in \textcolor{forestgreen}{green} (\textcolor{red}{red}) when significantly positive (negative), and in \textcolor{gray}{gray} otherwise.}

\label{tab:ablation_motion}
\vspace{-3mm}
\tabcolsep 6pt
\resizebox{\linewidth}{!}{%
\begin{tabular}{r l cc ccc}
\toprule
 & & \multicolumn{2}{c}{Whole image} & \multicolumn{3}{c}{Our mask} \\
\cmidrule(lr){3-4} \cmidrule(lr){5-7}
\color{gray} \# & Latent lock & F1 $\uparrow$ & SBR $\uparrow$ & F1 $\uparrow$ & R $\uparrow$ & Pr $\uparrow$\\
\midrule
\rowcolor{myazure}
\color{gray} 1 & probe & \textbf{27.3} & \textbf{52.2} & 42.3 & 35.1 & \textbf{60.5} \\
\color{gray} 2 & GT    & 25.1 & 46.4 & \textbf{44.5} & \textbf{37.6} & 59.2 \\
\color{gray} 3 & GT$_{0.5}$ & 25.9 & 49.5 & 42.2 & 35.0 & 58.0 \\
\color{gray} 4 & GT dir. & 26.2 & 51.8 & 41.1 & 34.1 & 59.3 \\
\midrule
\multicolumn{7}{l}{\textit{Improvements}} \\
\color{gray} 5 & $\Delta$ (2 $-$ 1) \scriptsize(GT vs probe) & \color{gray}{-2.2} & \color{red}{-5.8} & \color{gray}{+2.3} & \color{gray}{+2.5} & \color{gray}{-1.2} \\
\color{gray} 6 & $\Delta$ (3 $-$ 1) \scriptsize(GT$_{0.5}$ vs probe) & \color{gray}{-1.4} & \color{red}{-2.7} & \color{gray}{-0.1} & \color{gray}{-0.1} & \color{gray}{-2.5} \\
\color{gray} 7 & $\Delta$ (4 $-$ 1) \scriptsize(GT dir. vs probe) & \color{gray}{-1.1} & \color{gray}{-0.4} & \color{gray}{-1.1} & \color{gray}{-1.0} & \color{gray}{-1.2} \\
\bottomrule
\end{tabular}
}
\end{table}

\subsection{Mask extraction}
\label{sec:mask_extraction_results}

In \cref{tab:funflow_segmentation} we compare the first-frame mask $\mathcal{M}$ of \ourmethod{}, Dream2Flow~\cite{dharmarajan2025dream2flow} and Fun3DU~\cite{corsetti2025fun3du} against the ground-truth part mask $\mathcal{M}_{\mathrm{GT}}$ on FunFlow.
Dream2Flow requires an object name, which it uses to prompt GroundingDINO~\cite{liu2024groundingdino}, and segments the top box with SAM~2~\cite{ravi2024sam2}.
We extract this name from the instruction with Claude Haiku~5.5, removing the action and keeping the text that contextualizes the object, \eg \texttt{second drawer of the dresser} for \texttt{open the second drawer of the dresser}.
Fun3DU was developed on SceneFun3D~\cite{delitzas2024scenefun3d}, from which \ourdataset derives, but with a different objective: instead of segmenting the object part (\eg the top drawer), it segments the functional element (\eg the top drawer handle). 
It first uses Molmo~\cite{deitke2025molmo} (\texttt{allenai/Molmo-7B-D-0924}) to point to the object of interest, and then feeds the resulting points to SAM~\cite{kirillov2023segmentany} (\texttt{facebook/sam-vit-huge}) to obtain the masks.
We change the prompt, by using the same name provided to the Dream2Flow baseline, and adding \texttt{Point to the} as preamble (\eg, \texttt{Point to the second drawer of the dresser}).

\begin{table}[t]
    \centering
    \caption{
    First-frame segmentation of Dream2Flow, Fun3DU and of \ourmethod{} on FunFlow, against the ground-truth part mask.
    We report mIoU and the accuracy Acc@$t$, \ie the percentage of samples whose mask reaches an IoU of at least $t$ (\%, higher is better), and the number of samples with an empty mask out of 178 (lower is better).
    Our results are highlighted.
    }
    \label{tab:funflow_segmentation}

    \vspace{-3mm}
    \tabcolsep 6pt
    \resizebox{\linewidth}{!}{%
    \begin{tabular}{rl cccc}
    \toprule
    \color{gray} \# & Method & mIoU $\uparrow$ & Acc@0.5 $\uparrow$ & Acc@0.75 $\uparrow$ & \# Empty $\downarrow$ \\
    \midrule
    \color{gray} 1 & Dream2Flow~\cite{dharmarajan2025dream2flow} & 22.6 & 19.1 & 18.0 & 73 \\
    \color{gray} 2 & Fun3DU~\cite{corsetti2025fun3du} & 52.5 & 50.0 & 44.4 & \textbf{0} \\
    \rowcolor{myazure}\color{gray} 3 & \ourmethod{} & \textbf{82.7} & \textbf{87.6} & \textbf{84.3} & \textbf{0} \\
    \bottomrule
    \end{tabular}
    }
\end{table}

\ourmethod{} largely beats both methods, reaching 82.7 mIoU against 22.6 of Dream2Flow and 52.5 of Fun3DU, and likewise improves Acc@0.5 (87.6 against 19.1 and 50.0) and Acc@0.75 (84.3 against 18.0 and 44.4).
Neither Fun3DU nor our method ever returns an empty mask, while Dream2Flow does on 73 samples (41\%).
In general, we observe that GroundingDINO does not resolve the complex referring phrases of \ourdataset, and tends to ground the whole object.
Additionally, generic furniture names often yield no detection above threshold (cabinets, nightstands and closets account for 60 of the 73 empty masks), which explains the low accuracy of Dream2Flow.
Due to our evaluation method, these samples result in an automatic failure in all metrics.
Our pipeline (\cref{sec:mask_extraction}) avoids these failures by splitting \emph{what} to segment from \emph{which} instance is meant: SAM~3 is prompted with the part category (\textit{``drawer''}), proposes every instance at the granularity of the part, and the VLM picks one using the full instruction.
By including a VLM (Molmo), Fun3DU is more successful than Dream2Flow's method, while still falling behind ours.

\subsection{Confidence intervals}\label{sec:confidence}

\providecommand{\cirow}[2]{{\scriptsize\color{gray}[#1,\,#2]}}
\begin{table*}[t]
    \centering
    \caption{Comparison between \ourmethod{} and Dream2Flow on FunFlow.
    We report F1, recall (R), precision (Pr) and Still Background Rate (SBR) of the 3D object flow over the whole image, within our predicted mask, within the ground-truth mask, which also steers \ourmethod{}, or within each method's own mask (whole pipeline).
    Row 1 reports Dream2Flow with the proprietary Omni-Flash~\cite{deepmind2026geminiomniflash}.
    \textbf{Bold} denotes the best result for open VDMs. 
    Our results are highlighted.
    {\color{gray}[Brackets]} give the 95\% confidence interval of each gain from 10\,000 paired bootstrap resamples. 
    Differences are in \textcolor{forestgreen}{green} (\textcolor{red}{red}) when significantly positive (negative), and in \textcolor{gray}{gray} otherwise.}
    \label{tab:funflow_main_ci}
    \vspace{-3mm}
    \tabcolsep 5pt
    \resizebox{\textwidth}{!}{%
    \begin{tabular}{rl cccc c ccc c ccc c cccc}
    \toprule
    & & \multicolumn{4}{c}{Whole image} & & \multicolumn{3}{c}{Mask filter (our mask)} & & \multicolumn{3}{c}{Mask filter (GT mask)} & & \multicolumn{4}{c}{Whole pipeline} \\
    \cmidrule{3-6} \cmidrule{8-10} \cmidrule{12-14} \cmidrule{16-19}
    \color{gray} \# & Method & F1 $\uparrow$ & R $\uparrow$ & Pr $\uparrow$ & SBR $\uparrow$ & & F1 $\uparrow$ & R $\uparrow$ & Pr $\uparrow$ & & F1 $\uparrow$ & R $\uparrow$ & Pr $\uparrow$ & & F1 $\uparrow$ & R $\uparrow$ & Pr $\uparrow$ & SBR $\uparrow$ \\
    \midrule

    \multicolumn{19}{l}{\textit{Gemini Omni-Flash~\cite{deepmind2026geminiomniflash} \color{gray}(API)}} \\
    \color{gray} 1 & Dream2Flow~\cite{dharmarajan2025dream2flow} & 21.7 & 40.1 & 18.6 & 32.9 & & 44.9 & 38.0 & 60.3 & & 48.1 & 40.1 & 65.7 & & 19.5 & 21.8 & 23.1 & 32.5 \\
    \specialrule{0.1pt}{\aboverulesep}{\belowrulesep}
    \multicolumn{19}{l}{\textit{HunyuanVideo~\cite{kong2024hunyuanvideo}}} \\
    \color{gray} 2 & Dream2Flow~\cite{dharmarajan2025dream2flow} & 13.4 & 30.6 & 10.5 & 25.4 & & 34.2 & 28.8 & 49.2 & & 37.0 & 30.6 & 54.5 & & 13.2 & 17.7 & 15.8 & 29.5 \\
    \rowcolor{myazure}\color{gray} 3 & \ourmethod{} & \textbf{27.3} & \textbf{35.9} & \textbf{27.0} & \textbf{52.2} & & \textbf{42.3} & \textbf{35.1} & \textbf{60.5} & & \textbf{44.8} & \textbf{36.9} & \textbf{64.6} & & \textbf{42.3} & \textbf{35.1} & \textbf{60.5} & \textbf{88.1} \\
    \midrule
    \multicolumn{19}{l}{\textit{Improvements}} \\
    \color{gray} 4 & $\Delta$ (3 $-$ 2) & \color{forestgreen}{+13.9} & \color{forestgreen}{+5.3} & \color{forestgreen}{+16.5} & \color{forestgreen}{+26.8} & & \color{forestgreen}{+8.1} & \color{forestgreen}{+6.3} & \color{forestgreen}{+11.3} & & \color{forestgreen}{+7.8} & \color{forestgreen}{+6.3} & \color{forestgreen}{+10.1} & & \color{forestgreen}{+29.1} & \color{forestgreen}{+17.4} & \color{forestgreen}{+44.7} & \color{forestgreen}{+58.6} \\
     & & \cirow{+10.5}{+17.5} & \cirow{+0.8}{+9.9} & \cirow{+12.6}{+20.5} & \cirow{+23.5}{+30.1} & & \cirow{+3.2}{+13.0} & \cirow{+1.9}{+10.7} & \cirow{+5.4}{+17.4} & & \cirow{+2.8}{+13.0} & \cirow{+1.8}{+10.9} & \cirow{+3.4}{+16.9} & & \cirow{+23.3}{+34.8} & \cirow{+11.9}{+23.0} & \cirow{+37.1}{+52.2} & \cirow{+52.0}{+65.2} \\
    \color{gray} 5 & $\Delta$ (3 $-$ 1) & \color{forestgreen}{+5.6} & \color{gray}{-4.2} & \color{forestgreen}{+8.4} & \color{forestgreen}{+19.3} &  & \color{gray}{-2.6} & \color{gray}{-2.9} & \color{gray}{+0.2} &  & \color{gray}{-3.3} & \color{gray}{-3.2} & \color{gray}{-1.1} &  & \color{forestgreen}{+22.8} & \color{forestgreen}{+13.3} & \color{forestgreen}{+37.4} & \color{forestgreen}{+55.6} \\
     & & \cirow{+0.7}{+10.5} & \cirow{-10.1}{+1.9} & \cirow{+3.4}{+13.5} & \cirow{+15.2}{+23.5} & & \cirow{-9.1}{+3.8} & \cirow{-8.7}{+2.9} & \cirow{-8.0}{+8.0} & & \cirow{-10.0}{+3.6} & \cirow{-9.2}{+3.0} & \cirow{-10.1}{+7.3} & & \cirow{+15.9}{+29.7} & \cirow{+6.9}{+19.6} & \cirow{+28.7}{+45.8} & \cirow{+48.6}{+62.3} \\
    \bottomrule
    \end{tabular}
    }
\end{table*}

\providecommand{\no}{\ding{55}}
\providecommand{\cirow}[2]{{\scriptsize\color{gray}[#1,\,#2]}}

\begin{table}[t]
\caption{\ourmethod{} ablation on HunyuanVideo with our predicted mask and the whole image.
\emph{L.~lock} is the latent lock, expanded along the motion $\mathbf{v}^*$ probed from the heads (probe), or isotropically.
$\mathcal{H}$ is the motion head set used for motion probing: our top-10 heads, their whole layers, or all heads.
Seeds is the number of seeds available to MASS, which is disabled with one.
A dash marks a disabled module. 
Our standard setting is highlighted.
The bottom rows give the paired difference between two rows, with {\color{gray}[brackets]} below it giving the 95\% confidence interval from 10\,000 paired bootstrap resamples of the samples. Differences are in \textcolor{forestgreen}{green} (\textcolor{red}{red}) when significantly positive (negative), and in \textcolor{gray}{gray} otherwise.}
\label{tab:ablation_supp}
\vspace{-3mm}
\tabcolsep 5pt
\resizebox{\linewidth}{!}{%
\begin{tabular}{r c c c cc ccc}
\toprule
 & & & & \multicolumn{2}{c}{Whole image} & \multicolumn{3}{c}{Our mask} \\
\cmidrule(lr){5-6} \cmidrule(lr){7-9}
\color{gray} \# & L. lock & $\mathcal{H}$ & Seeds & F1 $\uparrow$ & SBR $\uparrow$ & F1 $\uparrow$ & R $\uparrow$ & Pr $\uparrow$\\
\midrule
\color{gray} 1 &  -- & --    & 1 & 13.4 & 25.4 & 34.2 & 28.8 & 49.2 \\
\color{gray} 2 & probe & heads & 1 & 25.4 & \textbf{55.5} & 38.1 & 31.4 & 55.4 \\
\color{gray} 3 & --  & heads & 5 & 14.5 & 24.4 & 38.7 & 32.9 & 52.0 \\
\rowcolor{myazure}
\color{gray} 4 & probe & heads & 5 & \textbf{27.3} & 52.2 & \textbf{42.3} & \textbf{35.1} & \textbf{60.5} \\
\midrule
\multicolumn{9}{l}{\textit{Variants of row 2}} \\
\color{gray} 5 & isotropic & --     & 1 & 21.0 & 51.7 & 32.4 & 26.0 & 51.9 \\
\color{gray} 6 & probe     & layers & 1 & 21.0 & 52.9 & 31.8 & 25.5 & 49.1 \\
\color{gray} 7 & probe     & all    & 1 & 21.4 & 52.4 & 32.7 & 26.3 & 49.7 \\
\midrule
\multicolumn{9}{l}{\textit{Improvements}} \\
\color{gray} 8 & \multicolumn{3}{l}{$\Delta$ (2 $-$ 1) \scriptsize(L.~lock)} & \color{forestgreen}{+12.0} & \color{forestgreen}{+30.1} & \color{gray}{+3.9} & \color{gray}{+2.5} & \color{gray}{+6.3} \\
\multicolumn{4}{l}{} & \cirow{+8.8}{+15.4} & \cirow{+26.9}{+33.5} & \cirow{-1.0}{+8.7} & \cirow{-1.8}{+6.8} & \cirow{-0.4}{+12.9} \\
\color{gray} 9 & \multicolumn{3}{l}{$\Delta$ (3 $-$ 1) \scriptsize(MASS)} & \color{gray}{+1.2} & \color{red}{-1.1} & \color{forestgreen}{+4.5} & \color{forestgreen}{+4.0} & \color{gray}{+2.8} \\
\multicolumn{4}{l}{} & \cirow{-0.1}{+2.6} & \cirow{-2.0}{-0.2} & \cirow{+1.1}{+8.1} & \cirow{+1.1}{+7.3} & \cirow{-1.1}{+6.8} \\
\color{gray} 10 & \multicolumn{3}{l}{$\Delta$ (4 $-$ 3) \scriptsize(L.~lock on MASS)} & \color{forestgreen}{+12.8} & \color{forestgreen}{+27.8} & \color{gray}{+3.6} & \color{gray}{+2.2} & \color{forestgreen}{+8.5} \\
\multicolumn{4}{l}{} & \cirow{+9.7}{+16.0} & \cirow{+24.5}{+31.2} & \cirow{-0.8}{+8.2} & \cirow{-1.8}{+6.3} & \cirow{+3.3}{+14.0} \\
\color{gray} 11 & \multicolumn{3}{l}{$\Delta$ (4 $-$ 2) \scriptsize(MASS on L.~lock)} & \color{gray}{+1.9} & \color{red}{-3.3} & \color{forestgreen}{+4.2} & \color{forestgreen}{+3.7} & \color{forestgreen}{+5.0} \\
\multicolumn{4}{l}{} & \cirow{-0.2}{+4.3} & \cirow{-5.3}{-1.6} & \cirow{+1.0}{+7.6} & \cirow{+0.9}{+6.7} & \cirow{+0.5}{+9.6} \\
\color{gray} 12 & \multicolumn{3}{l}{$\Delta$ (5 $-$ 2) \scriptsize(isotropic vs probe)} & \color{red}{-4.4} & \color{red}{-3.8} & \color{red}{-5.7} & \color{red}{-5.3} & \color{gray}{-3.5} \\
\multicolumn{4}{l}{} & \cirow{-7.4}{-1.4} & \cirow{-6.0}{-1.6} & \cirow{-9.9}{-1.6} & \cirow{-9.0}{-1.8} & \cirow{-9.0}{+1.7} \\
\color{gray} 13 & \multicolumn{3}{l}{$\Delta$ (6 $-$ 2) \scriptsize(layers vs heads)} & \color{red}{-4.4} & \color{red}{-2.6} & \color{red}{-6.3} & \color{red}{-5.8} & \color{red}{-6.3} \\
\multicolumn{4}{l}{} & \cirow{-7.1}{-1.7} & \cirow{-4.9}{-0.2} & \cirow{-10.0}{-2.5} & \cirow{-9.2}{-2.5} & \cirow{-11.8}{-0.9} \\
\color{gray} 14 & \multicolumn{3}{l}{$\Delta$ (7 $-$ 2) \scriptsize(all vs heads)} & \color{red}{-3.9} & \color{red}{-3.1} & \color{red}{-5.4} & \color{red}{-5.1} & \color{red}{-5.7} \\
\multicolumn{4}{l}{} & \cirow{-6.7}{-1.2} & \cirow{-5.5}{-0.7} & \cirow{-9.3}{-1.4} & \cirow{-8.6}{-1.6} & \cirow{-11.4}{-0.2} \\
\bottomrule
\end{tabular}
}
\end{table}

In \cref{tab:funflow_main_ci} we report the bootstrap confidence interval of our main table, as described in \cref{sec:bootstrap}, on the improvement of our method over Dream2Flow using HunyuanVideo (row 4) and Dream2Flow using Omni-Flash (row 5).
All our increments on HunyuanVideo (row 4) are statistically significant.
When comparing to Omni-Flash, our increments in the whole image are significant (except for recall), while under the same mask filtering (either ground-truth or predicted) the differences are not significant.

In \cref{tab:ablation_supp} we report the original ablation table of the main paper, and add all the row comparisons we discussed, along with confidence intervals for each value.
Considering rows 8--11, all our additions provide significant improvements in at least two metrics, with large gains in the whole image setting when adding the latent lock (rows 8 and 10).
The two exceptions are SBR when comparing rows 3-1 (-1.1) and when comparing rows 4-2 (-3.3), which show small, but significant losses.
Both ablations add the MASS retries.
This loss could be due to the bias of MASS to favor samples in which the object moves, which can also encourage spurious motions.
Nonetheless, both losses are compensated by gains in F1 score when using our mask to crop (4.5 and 4.2).
In rows 5, 6 and 7 we observe that the losses due to using an isotropic latent lock (row 5) or choosing a different head set (rows 6 and 7) are significant in nearly all metrics.

\section{Implementation details}\label{sec:implementation}

FunFlow images keep the resolution of SceneFun3D, \ie $1440 \times 1920$ pixels in landscape or portrait orientation.
Each VDM generates at its own resolution, so we resize the start image \image to a size close to the one the VDM expects, and rescale its depth map and camera intrinsics accordingly.
We call $\hat{\mathcal{I}}$ the resized start image.
Every resize uses a single scale factor for both axes, so images are never distorted.
When the aspect ratio does not match, we pad the image with black bands (except for Cosmos, which we crop), set the depth of the bands to zero, \ie invalid, and shift the intrinsics by the padding offset.
The bands are therefore never part of the object flow: particles are only sampled on pixels with valid depth, and the bands are also excluded from the background on which SBR is computed.
Every method outputs $T = 21$ frames aligned to $\hat{\mathcal{I}}$, on which the flow is computed: when a VDM generates at a size different from $\hat{\mathcal{I}}$, we crop out any padding added for it and resize the frames back to $\hat{\mathcal{I}}$.
\cref{tab:vdm_settings} summarizes the settings of the three VDMs.
All the open models were run on a single NVIDIA GB10.

\begin{table*}[t!]
  \tabcolsep 16pt
  \caption{Settings of the three VDMs. Resolutions are for landscape images (height $\times$ width); portrait images use the transposed size.
  FPS is given only for Cosmos, which takes it as an input parameter for the generation. 
  Cosmos and HunyuanVideo have a dedicated input for the negative prompt, while for Omni-Flash we append the negative prompt to the standard prompt.
  Parameters and denoising step numbers of Omni-Flash are unknown, as it is a proprietary model.
  }
  \label{tab:vdm_settings}

  \vspace{-2mm}
  \resizebox{\linewidth}{!}{%
  \begin{tabular}{lccc}
    \toprule
    & HunyuanVideo & Omni-Flash & Cosmos3-Nano \\
    \midrule
    Access & open weights (Hugging Face) & Gemini API & open weights (Hugging Face) \\
    Parameters & 17.3B (8.3B DiT) & ? & 17.0B (15.2B Transformer) \\
    Generation resolution & $512 \times 768$ & $720 \times 1280$ & $832 \times 1088$ \\
    Generated frames & 21 & 96 (4\,s) & 41 (4\,s at 10\,FPS) \\
    Denoising steps & 50 & ? & 35 \\
    Negative prompt & native & in the prompt & native \\
    \bottomrule
  \end{tabular}
  }
\end{table*}

\noindent \textbf{HunyuanVideo.}
We use HunyuanVideo-1.5~\cite{kong2024hunyuanvideo} in its 480p image-to-video version, from the Diffusers checkpoint \texttt{HunyuanVideo-1.5-Diffusers-480p\_i2v} from \texttt{hunyuanvideo-community} on Hugging Face, run in bfloat16 with the Diffusers image-to-video pipeline.
We keep the original sampling parameters, \ie 50 denoising steps and a guidance scale of 6.0, and reduce only the number of frames from 121 to 21 for efficiency, which gives $T' = 6$ latent frames.
We downscale \image to $480 \times 640$ and pad its long side with 40 black pixels at each end, which gives $\hat{\mathcal{I}}$ at $480 \times 720$.
The pipeline resizes $\hat{\mathcal{I}}$ to $512 \times 768$, and the generated frames are downscaled back to $480 \times 720$.

\noindent \textbf{Omni-Flash.}
Omni-Flash~\cite{deepmind2026geminiomniflash} is only available through the Gemini API.
We use the model \texttt{gemini-omni-1.1-flash}.
The API exposes no negative prompt, number of frames or duration, so we append the negative prompt to the prompt and ask for a 4-second video.
$\hat{\mathcal{I}}$ is the same $480 \times 720$ padded image as for HunyuanVideo.
The API returns 720p videos in 16:9 or 9:16, so we send $\hat{\mathcal{I}}$ already fitted to the 720p frame that matches its orientation: we upscale $\hat{\mathcal{I}}$ to $720 \times 1080$ and pad its long side with black pixels to 1280.
From the returned video we sample 21 frames uniformly, crop each one to the area that $\hat{\mathcal{I}}$ filled, and downscale it back to $480 \times 720$.

\noindent \textbf{Cosmos.}
We use Cosmos3-Nano~\cite{agarwal2026cosmos}, from the Diffusers checkpoint \texttt{Cosmos3-Nano} from \texttt{nvidia} on Hugging Face, run in bfloat16 with the Diffusers \texttt{Cosmos3OmniPipeline}.
We use 35 denoising steps and a guidance scale of 6.0, and generate 41 frames at 10 frames per second, which gives $T' = 11$ latent frames.
We generate 41 frames because we observed better results than with 21, and keep every second frame before computing the flow, so that Cosmos is also evaluated on $T = 21$ frames.
Cosmos only generates at a fixed set of resolutions, grouped in tiers named after the short side of their 16:9 resolution.
We use the 704 tier ($704 \times 1280$ for 16:9), whose resolution closest to the 3:4 FunFlow images is $832 \times 1088$.
We therefore resize \image to $832 \times 1109$, and center-crop its long side to 1088 pixels, which gives $\hat{\mathcal{I}}$.
Since Cosmos generates directly at the size of $\hat{\mathcal{I}}$, no padding or resizing is necessary.

\noindent \textbf{Prompts.}
All VDMs receive the task description rephrased by Claude Haiku~5.5.
HunyuanVideo and Omni-Flash receive it followed by a request for a static camera (\texttt{Static camera, locked shot, no zoom, no pan, no tilt, no dolly.}), and the negative prompt \texttt{fast motion, camera motion}.
Cosmos receives the task description inside its official prompt, whose fields request a static camera, with its official negative prompt.

{
    \small
    \bibliographystyle{supp/ieeenat_fullname}
    \bibliography{supp}
}